\documentclass[10pt]{article} 
\usepackage[accepted]{tmlr}

\usepackage{amsmath,amsfonts,bm}

\def\eqref#1{equation~\ref{#1}}

\def\1{\bm{1}}

\DeclareMathAlphabet{\mathsfit}{\encodingdefault}{\sfdefault}{m}{sl}
\SetMathAlphabet{\mathsfit}{bold}{\encodingdefault}{\sfdefault}{bx}{n}

\newcommand{\R}{\mathbb{R}}

\usepackage{hyperref}
\usepackage{url}
\usepackage{bbm}

\title{From Tables to Time:\\Extending \tabpfn to time-series forecasting}

\author{\name Shi Bin Hoo \email hoos@tf.uni-freiburg.de \\
      \addr University of Freiburg\\
      Prior Labs
      \AND
      \name Samuel Müller\thanks{Work conducted while at the University of Freiburg.} \email{sammuller@meta.com}\\
      \addr University of Freiburg\\
      Meta
      \AND
      \name David Salinas \email{david.salinas@tue.ellis.eu}\\
      \addr ELLIS Institute Tübingen\\
      University of Freiburg
      \AND
      \name Frank Hutter \email{frank@priorlabs.ai}\\
      \addr Prior Labs\\
      ELLIS Institute Tübingen\\
      University of Freiburg
}

\usepackage{subcaption}
\usepackage{graphicx}
\usepackage{xspace}
\usepackage{algorithm, algorithmicx, algpseudocode}
\usepackage{wrapfig}
\usepackage{booktabs}
\usepackage{multirow}
\usepackage{multicol}
\usepackage{todonotes}
\usepackage{chngcntr}
\counterwithout{figure}{section}

\def\tabpfnts{TabPFN-TS\xspace}
\def\tabpfn{TabPFN-v2\xspace}

\def\month{06}  
\def\year{2026} 
\def\openreview{\url{https://openreview.net/forum?id=KIkQj8VOUY}} 

\begin{document}

\maketitle

\begin{abstract}
Recent progress in foundation models has enabled strong zero-shot performance for time-series forecasting.
In this work, we show that these capabilities also emerge from tabular foundation models.
We introduce \tabpfnts, a simple method that treats forecasting as a tabular regression problem by combining lightweight temporal featurization with the pretrained \tabpfn.
This formulation requires no time-series–specific pretraining and naturally supports both univariate and covariate-informed forecasting.
Despite its compact size ($11$M parameters), \tabpfnts achieves state-of-the-art performance on covariate-informed forecasting and competitive accuracy on univariate forecasting across the GIFT-Eval and \texttt{fev-bench} benchmarks.
We further provide controlled analyses examining how the model interprets temporal structure, how featurization choices affect accuracy, and how forecasts change under alternative tabular backbones.
Together, our results demonstrate that tabular foundation models—when paired with suitable temporal features—offer an efficient and versatile alternative for forecasting, bridging tabular and time-series learning within a unified framework.
\end{abstract}

\section{Introduction}

Recent advances in foundation models have reshaped time-series forecasting by enabling zero-shot prediction across a wide variety of domains.
Large pretrained architectures can now generalize to unseen series and horizons without task-specific fine-tuning, setting a new paradigm for forecasting research.
Comprehensive benchmarks~\citep{gift-eval, shchur2025fevbenchrealisticbenchmarktime} have further standardized evaluation, enabling fair and reproducible comparison.

While most foundation models focus on univariate forecasting, real-world systems are often influenced by additional factors.
Incorporating such covariates---e.g., weather, control inputs, or economic indicators---can make forecasts far more informative and actionable. 
Yet leveraging covariates remains challenging for current time-series foundation models, which are typically designed around purely sequential inputs.
A unified and data-efficient framework capable of handling both univariate and covariate-informed forecasting would therefore be highly desirable.

In parallel, foundation models for \emph{tabular learning} have made remarkable progress.
Recent tabular models exhibit strong zero-shot predictive performance across diverse supervised tasks~\citep{erickson2025tabarenalivingbenchmarkmachine}.
Among them, \tabpfn~\citep{hollmann2025accurate} is notable as the first tabular foundation model, supporting many tasks, including classification, regression, outlier detection, density estimation, synthetic data generation, embeddings that are useful for downstream tasks, and fine-tunability. 
This raises a natural question:
\emph{Can a tabular foundation model also perform probabilistic time-series forecasting if the temporal structure is expressed through features?}
If so, tabular models would offer a unified representation for both tabular and time-series forecasting, sidestepping the need for specialized sequence architectures.

In this work, we propose to treat time-series forecasting as a \textit{tabular regression problem}.
We introduce \tabpfnts, a simple method that combines lightweight temporal featurization with the pretrained \tabpfn, enabling zero-shot forecasting without any time-series–specific pretraining or fine-tuning.
Each time step is represented as a row of temporal features (and optional covariates) paired with its observed value.
Forecasting then reduces to predicting future rows---whose temporal features are known in advance---using \tabpfn as a tabular regressor.
This formulation allows all future points within the prediction horizon to be predicted in a single forward pass.

Despite its simplicity and compact size ($11$M parameters), \tabpfnts achieves state-of-the-art performance on covariate-informed forecasting and competitive accuracy on univariate benchmark.
Our analyses further show how the model interprets temporal structure, how featurization affects accuracy, and how forecasts change under different tabular backbones.

Our main contributions are:
\begin{enumerate}
    \item \textbf{Forecasting as Tabular Regression.} We introduce a formulation that enables a pretrained tabular foundation model (\tabpfn) to perform both univariate and covariate-informed forecasting in a zero-shot manner.
	\item \textbf{Lightweight Temporal Featurization.} We design a compact feature scheme that encodes time progression, multi-scale seasonality, and covariates, allowing tabular models to act on temporal data effectively.
	\item \textbf{Strong Empirical Performance.} \tabpfnts achieves state-of-the-art performance on covariate-aware forecasting and competitive accuracy on univariate forecasting across GIFT-Eval and \texttt{fev-bench}.
	\item \textbf{Mechanistic and Ablation Studies.} We provide analyses explaining how \tabpfnts exploits temporal structure, how featurization choices affect performance, and how forecasting quality varies across tabular foundation model backbones.
\end{enumerate}


Taken together, these results demonstrate that tabular foundation models, paired with suitable temporal features, form a simple, efficient, and extensible alternative for general-purpose forecasting, bridging tabular and time-series learning within a unified framework.



\section{Background and Related Work}

\paragraph{time-series forecasting.} The goal is to predict future observations of a sequence based on its historical context and, optionally, a set of covariates---external variables that provide additional information about the system's dynamics (e.g. holidays, known prices, or control inputs).
Incorporating such signals can improve forecast accuracy by capturing influences beyond the intrinsic temporal patterns of the target series.
Formally, the task can be expressed as modeling the conditional distribution 
\[P(y_{C+1:H} \mid y_{1:C}, Z_{1:C+H}),\]
where $y_{1:C} = [y_1, \dots, y_C]$ denotes the observed history of the target series, $y_{C+1:H} = [y_{C+1},\dots,y_H]$ are the future values to be predicted, and $Z_{1:C+H} = [\mathbf{z}_1,\dots,\mathbf{z}_{C+H}]$ represents the associated covariates.
When no covariates are provided, the task reduces to \textit{univariate forecasting}; otherwise, it corresponds to \textit{covariate-informed forecasting}.

\paragraph{Time Series Foundation Models.}
Time series foundation models (TSFMs) aim to provide zero-shot forecasting across domains by pretraining (potentially) large sequence models on broad collections of real-world time series.  
Early models such as Chronos~\citep{ansari2024chronos} and TimesFM~\citep{das2024timesfm} adopt transformer architectures trained at scale for univariate forecasting.  
Chronos-Bolt improves efficiency beyond Chronos through patch-based encoding, while Toto~\citep{cohen2024tototimeseriesoptimized} targets multivariate forecasting with proportional factorized attention and a Student-\(t\) mixture output layer.  
TiRex~\citep{auer2025tirexzeroshotforecastinglong} leverages an xLSTM backbone with strong in-context learning, and Sundial~\citep{liu2025sundialfamilyhighlycapable} introduces a flow-matching objective for flexible probabilistic forecasting.

Several TSFMs incorporate covariates.  
ChronosX~\citep{arango2025chronosx} injects exogenous variables through modular adapters;  
and COSMIC~\citep{auer2025cosmic} augments forecasting with in-context covariate conditioning.\footnote{ChronosX and COSMIC are not publicly available at the time of writing and are therefore not included in our comparison.}

\paragraph{\tabpfn.}
TabPFN~\citep{hollmanntabpfn} is a foundation model pretrained to perform tabular prediction directly from examples. Instead of being optimized on a single dataset, it is pretrained on millions of synthetic regression and classification tasks. Each task consists of input–output pairs $(X, y)$ as the context and query points $X’$ for which the model must predict corresponding targets $y'$. Through this large-scale pretraining, TabPFN learns general patterns of how features relate to targets that transfer across unseen problems.

At inference, the model receives a table of feature-target pairs $(X, y)$ as context and outputs the posterior predictive distribution $p(y’ \mid X, y, x’)$ for new inputs $X'$. TabPFN represents this distribution as a \emph{Riemann distribution}---a discretized probability density over possible target values---and is trained directly on this probabilistic representation. 
Having the model output a distribution allows to handle downstream cases which requires a distribution, for instance to account for the model uncertainty.

\tabpfn~\citep{hollmann2025accurate} extends its predecessor TabPFN with architectural improvements, a richer synthetic data generator, while adding native support for handling missing values, outliers, and uninformative feature.
Together, these properties make \tabpfn a powerful probabilistic tabular regressor.

Our approach to time-series forecasting is complementary to standard TSFMs: rather than pretraining a new sequence model, we use \tabpfn and reformulate forecasting as tabular regression via temporal featurization.   Unlike most TSFMs—which are designed around autoregressive or \texttt{seq2seq} architectures---\tabpfnts predicts the full horizon in a single non-autoregressive forward pass, offering a strong alternative to sequence-based foundation models.

\section{Methodology} \label{sec:methodology}

\def\xtrain{\mathbf{X}_{\text{train}}}
\def\xtest{\mathbf{X}_{\text{test}}}
\def\ytrain{\mathbf{y}_{\text{train}}}

\begin{figure}
    \centering
    \includegraphics[width=0.9\linewidth]{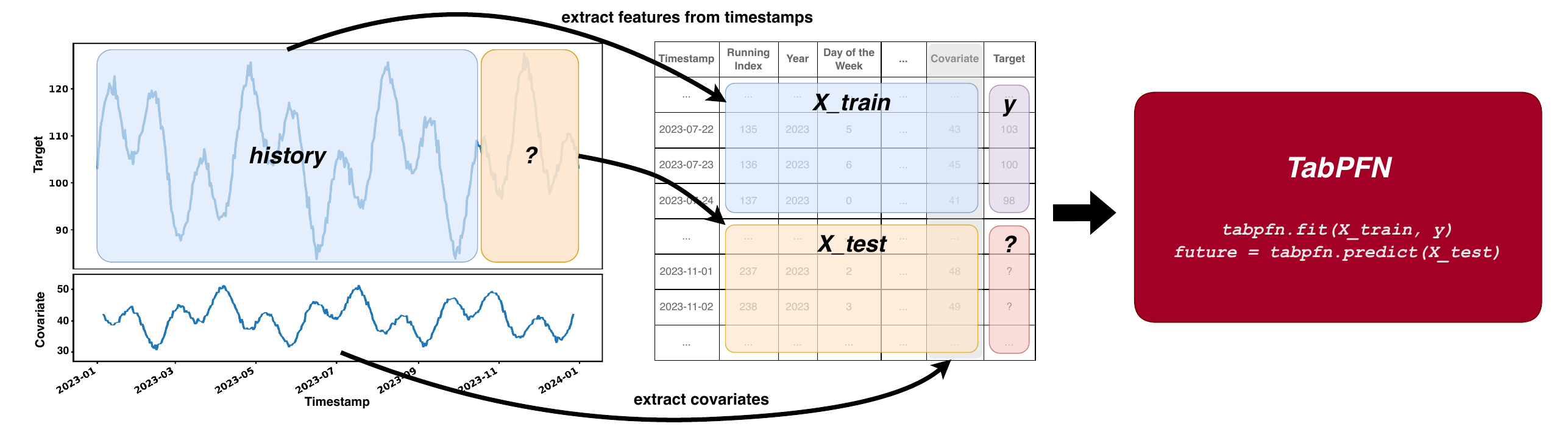}
    \caption{Overview of \tabpfnts.
    From a given time series, we construct feature matrices $\xtrain$ and $\xtest$, optionally appending exogenous covariates when available. The two matrices represent the historical and future windows, respectively.
    The observed historical targets form $\ytrain$.
    \tabpfn then conditions on the context $(\xtrain, \ytrain)$ and predicts the corresponding targets at $\xtest$.
    }
    \label{fig:method-overview}
\end{figure}

In this section, we present \tabpfnts, a novel approach to using \tabpfn for multi-step, univariate time-series forecasting.
We recast time-series forecasting as a tabular regression problem, where a time series sequence is treated as a table, as shown in Figure \ref{fig:method-overview}.

\subsection{From Time Series to Tabular Data} \label{sec:from-ts-to-tabular}

Given a time series $\mathbf{y}_{1:C+H} = [y_1, \dots, y_{C+H}]$, the first $C$ observations define the historical context, and the remaining $H$ correspond to the forecast horizon. 
As illustrated in Figure~\ref{fig:method-overview}, we construct feature matrices $\xtrain \in \mathbb{R}^{C\times D}$ and $\xtest \in \mathbb{R}^{H \times D}$, where $D$ denotes the number of features.
The corresponding historical targets form $\ytrain \in \mathbb{R}^{C}$.
This representation, consisting of $(\xtrain, \ytrain)$ and $\xtest$, allows the use of any standard supervised tabular models, including \tabpfn.

\paragraph{Running Index.}
To introduce a temporal reference within the timeline, we include the index of each time step as a feature (e.g., 0 for the first time step in the time series, 4 for the fifth):
\begin{align*}
\Phi_\text{index}(t) = t \in \R^{1}.
\end{align*}
This provides a straightforward and effective way to track the progression of time across the observations.

\paragraph{Calendar Features.}
From each timestamp, we encode eight cyclic calendar components: second of minute, minute of hour, hour of day, day of week, day of month, day of year, week of year, and month of year.
We additionally include the calendar year as a separate (non-cyclic) feature.
Let $\{P_i\}_{i=1}^8$ be the periods associated with these cyclic components.
The corresponding features for each timestamp $t$ are encoded as
\begin{align*}
\Phi_\text{cal}(t) = (\cos\Bigl(\frac{2\pi t}{P_1}\Bigr), \sin\Bigl(\frac{2\pi t}{P_1}\Bigr), \dots, \cos\Bigl(\frac{2\pi t}{P_8}\Bigr), \sin\Bigl(\frac{2\pi t}{P_8}\Bigr), \text{year}(t)) \in \R^{17}
\end{align*}
For full implementation details, see Appendix~\ref{append-calendar-feat}.

\paragraph{Automatic Seasonal Features.}

Beyond the standard calendar periodicities, time series often have domain-specific cycles that calendar-based encodings fail to capture, e.g., depending on non-Gregorian calendars (Chinese birthdays) or the moon cycle (tides). 
To address this, we apply an automatic extraction process to identify the top-$k$ periodicities and encode them as features, thereby enriching the seasonality inputs to the model.

\begin{algorithm}[ht]
    \caption{Automatically extract top-$k$ Seasonalities, see Appendix \ref{append-auto-seasonal-features} for a detailed algorithm}
    \label{alg-auto-seasonal-feature}
  

\begin{algorithmic}
\Statex \textbf{Require:} univariate series \(\mathrm{series}\), the number of periods to obtain \(k\), the smoothing window size $L$
\Statex
\Statex \textbf{Preprocess $\mathrm{series}$}
\State Detrend linearly: $\mathrm{series}[t] = \mathrm{series}[t] - (\alpha t + \beta)$, where $\alpha$ and $\beta$ are found using least squares
\State Apply Hann window: $\mathrm{series} = \text{conv}(\mathrm{series}, w_\text{Hann}(L))$
\State Double length by symm.\ zero-padding: $\mathrm{series} = [0,\dots,0, \mathrm{series}[0], \dots, \mathrm{series}[N],0,\dots,0]$
\Statex
\Statex \textbf{Fourier Transform}
\State Compute FFT magnitudes $\mathrm{mags}$ and frequencies $\mathrm{freqs}$ based on the preprocessed $\mathrm{series}$
\State Set $\mathrm{mags}[0] = 0$ (remove the zero-frequency component)
\Statex
\Statex \textbf{Select Peaks}
\State Find all peak indices $\mathrm{peaks}$ in $\mathrm{mags}$ (all (groups of) points larger than their neighbors)
\Statex
\Statex \textbf{Convert \& Clean}
\State Invert frequencies to periods and round $\mathrm{periods} = \lfloor1/\mathrm{freqs}\rceil$
\State Remove duplicate and zero periods from the peak indices $\mathrm{peaks}$
\State Keep only top $k$ $\mathrm{peaks}$: $\mathrm{peaks} = [i \text{ for } i \text{ in } \mathrm{peaks} \text{ if } i \text{ in } \text{topk}(\mathrm{mags[peaks]})]$
\Statex
\Statex
\Return $\mathrm{periods[peaks]}$
\end{algorithmic}
\end{algorithm}

We first detrend each series via a simple linear least-squares regression.
To reduce spectral leakage and improve frequency resolution, we apply a Hann window \citep{hann-window} and zero-pad the windowed signal by a factor of two \citep{discrete-signal-processing}.
We then compute the real-valued discrete Fourier transform and remove the zero-frequency component of the spectrum, which corresponds to the mean of the time series and does not represent a seasonal oscillation.
Finally, we select the $k$ largest spectral peaks by magnitude.
Algorithm \ref{alg-auto-seasonal-feature} provides high-level pseudo-code for this extraction process.

Given the detected frequencies $f_1, \dots, f_k$, we then build the following features:
\begin{align*}
\Phi_\text{auto}(t) = (\cos(2\pi f_1 t),~ \sin(2\pi f_1 t),~\dots,~\cos(2\pi f_k t),~\sin(2\pi f_k t)) \in \mathbb{R}^{2k}.
\end{align*}

\paragraph{Time-varying Covariates (when available).}

\tabpfnts also supports time-varying covariates (although only those that are known for future points in time, such as holidays, not unknown ones like weather). We denote them by
$\mathbf{z}_{1:C+H} = [\mathbf{z}_1, \dots, \mathbf{z}_{C+H}]$,
where each $\mathbf{z}_t \in \mathbb{R}^{D_z}$ represents $D_z$ external variables observed (or known) at time $t$.

For each timestamp, the full feature vector becomes:
\[
\mathbf{x_t} =  \Phi_\text{index}(t) \oplus \Phi_\text{cal}(t) \oplus \Phi_\text{auto}(t)  \oplus \mathbf{z}_t
\in \mathbb{R}^{(18 + 2k + D_z)}.
\]

\paragraph{Combining all Features.}
Stacking these feature vectors across all timestamps yields the full feature matrix that describes the time series
\[
\mathbf{X} = [\,\mathbf{x_1}; \dots; \mathbf{x_{C+H}}\,] \in \mathbb{R}^{(C+H) \times D},
\]
where $D=(18 + 2k + D_z)$ is the total number of features.
Note that we do not rely on lagged or auto-regressive features (e.g., moving averages and lag terms), since these require past predictions and conflict with non-auto-regressive, multi-step forecasting making the inference much slower. 

As described in Section~\ref{sec:from-ts-to-tabular}, we split $\mathbf{X}$ into
$\xtrain = X_{1:C,:}$ and $\xtest = X_{C+1:C+H}$,
corresponding to the historical and forecast windows.


\subsection{Point and Probabilistic Forecasting with \tabpfn}
We treat $(\xtrain, \ytrain)$ as a classical regression dataset and feed it into \tabpfn.
For each test input $x$ in $\xtest$, \tabpfn outputs an approximate posterior predictive distribution $p(y \mid \xtrain,\,\ytrain,\,x)\,$.

In \tabpfn, this distribution is represented on a fixed numerical grid over the target space by assigning probabilities to bins (a discretized density). From this representation, we compute point predictions and probability summaries in the following way: the mean for squared-error evaluations, the median for absolute-error evaluations, and arbitrary quantiles (e.g. $5\%$, $50\%$, $95\%$) for probabilistic metrics and uncertainty bands.


\section{Experiments} \label{sec:experiment}

We present quantitative comparisons of \tabpfnts against state-of-the-art forecasting models on both univariate (Section~\ref{sec:univariate-forecasting}) and covariate-aware forecasting (Section~\ref{sec:covariate-forecasting}) tasks.
For all evaluations, we use a fixed configuration of \tabpfn described in Appendix~\ref{sec:append-tabpfn-ts-config}.

\subsection{Univariate Forecasting} \label{sec:univariate-forecasting}
We evaluate \tabpfnts on GIFT-Eval \citep{gift-eval}, a comprehensive benchmark developed to evaluate general time-series forecasting models.

\paragraph{Datasets.} \label{sec:dataset}
GIFT-Eval comprises $23$ datasets with diverse characteristics, encompassing over $144,000$ time series and $177$ million data points across seven application domains and ten different sampling frequencies.
It covers both univariate and multivariate forecasting settings, as well as a wide range of prediction horizons, from short- to long-term forecasts.
Considering all valid combinations of datasets, sampling frequencies, and prediction horizons, GIFT-Eval contains a total of 97 distinct benchmarking tasks.
An overview of the datasets and their corresponding statistics is provided in Appendix~\ref{sec:append-dataset}.

\paragraph{Baselines.}


We benchmark \tabpfnts against a comprehensive set of baselines spanning statistical methods, deep learning models, and time series foundation models\footnote{All baselines included were publicly available before September 2025, our literature cutoff date.}.
For classical statistical methods, we include Seasonal Naive, AutoETS, AutoARIMA, and AutoTheta~\citep{garza2022statsforecast}.
Among deep learning approaches, we evaluate DeepAR~\citep{salinas2020deepar} and Temporal Fusion Transformer (TFT)~\citep{lim2021temporal}.
For foundation models baselines, we include TiRex~\citep{auer2025tirexzeroshotforecastinglong}, Toto-1.0~\citep{cohen2024tototimeseriesoptimized}, TimesFM-2.0~\citep{das2024timesfm}, Chronos-Bolt~\citep{ansari2024chronos}, and Moirai-2~\citep{woo2024unifiedtraininguniversaltime}.





\paragraph{Evaluation Metrics.} \label{metric}

Following prior work~\citep{ansari2024chronos, agtimeseries, gift-eval}, we assess point forecast accuracy using the Mean Absolute Scaled Error (MASE) and probabilistic forecast accuracy using the Weighted Quantile Loss (WQL).
MASE normalizes the absolute forecast error by the historical seasonal error of each series, yielding a scale-invariant measure comparable across datasets.
WQL evaluates the discrepancy between the predictive distribution and the observed value across quantile levels, providing a proxy assessment of probabilistic calibration.
Consistent with GIFT-Eval, we compute WQL at uniformly spaced quantiles ${\{0.1, 0.2, \dots, 0.9\}}$. Following \citet{ansari2024chronos}, we aggregate relative scores across datasets using the geometric mean and additionally report the mean rank of WQL.

\def\timesfm{TimesFM-2.0-500m\xspace}

\paragraph{Results.} \label{main-result}

\begin{figure}
    \centering
    \includegraphics[width=0.95\textwidth]{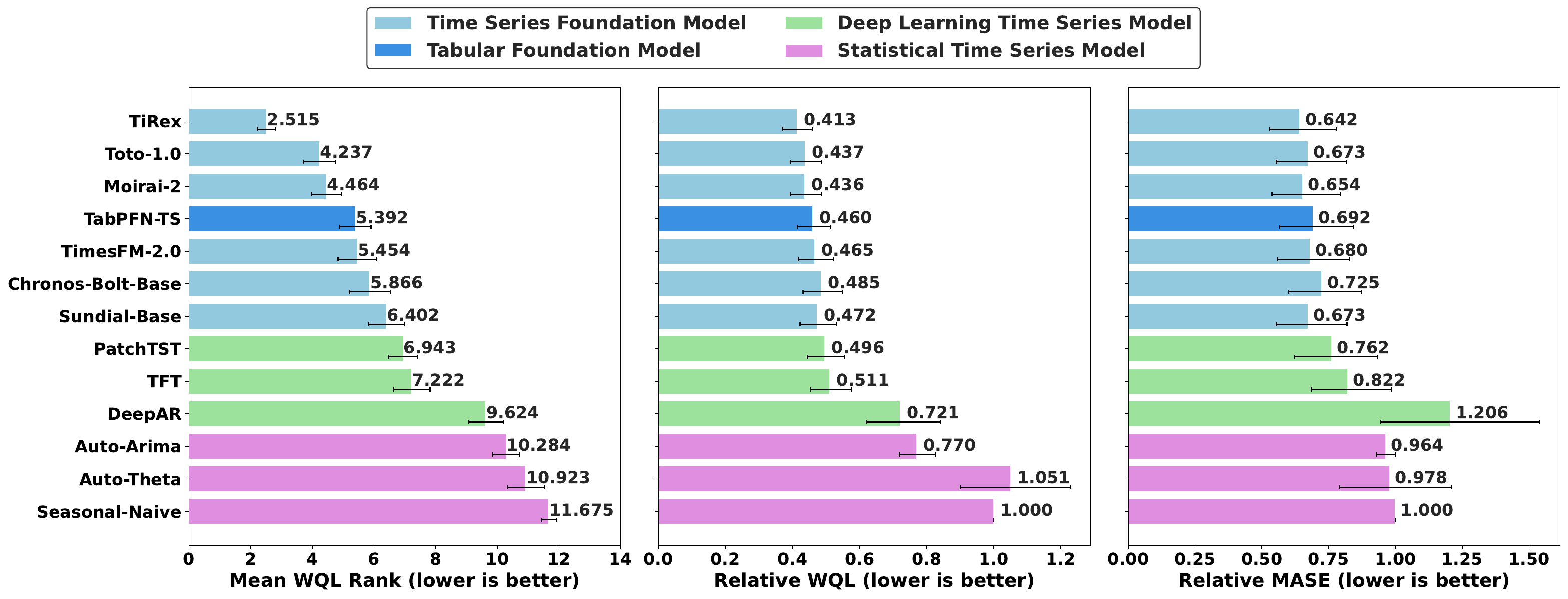}
    \caption{Univariate forecasting performance of \tabpfnts and baseline models on GIFT-Eval benchmark. Although \tabpfnts is not the top-ranked model, it achieves competitive performance on par with other time series foundation models (e.g. TimesFM2.0, Chronos-Bolt) while being a much smaller model ($11$M) and pretrained solely on tabular data. Scores are normalized by Seasonal Naive and aggregated across datasets; error bars show $95$\% confidence intervals.
    }
    \label{fig:main-result}
\end{figure}

Figure~\ref{fig:main-result} summarizes the forecasting performance of \tabpfnts on the GIFT-Eval benchmark.
In both probabilistic forecasting (WQL) and point forecasting (MASE), \tabpfnts ranks closely behind the leading foundation models TiRex, Toto-1.0, and Moirai-2, and matches substantially larger models such as TimesFM-2.0 and Chronos-Bolt, despite being over two orders of magnitude smaller and trained without any time-series–specific pretraining.
Qualitative visualizations are provided in Appendix~\ref{append-additional-res} for reference.

These results demonstrate that \tabpfnts produces accurate forecasts even though its backbone, \tabpfn, is pretrained solely on synthetic tabular data and relies on a simple temporal featurization.
We additionally find that \tabpfnts's predictive intervals are
comparably calibrated to most time-series foundation model baselines (see
Appendix~\ref{app:calibration}).
This finding suggests that (i) the knowledge learned from generic tabular data can transfer effectively to temporal prediction, and (ii) further improvements are likely if tabular foundation models are exposed to real-world time-series data during pretraining.
Together, these results position \tabpfnts as a compact yet competitive alternative to large time-series foundation models.

\subsection{Covariate-Informed Forecasting} \label{sec:covariate-forecasting}

\begin{figure}
    \centering
    \includegraphics[width=0.9\textwidth]{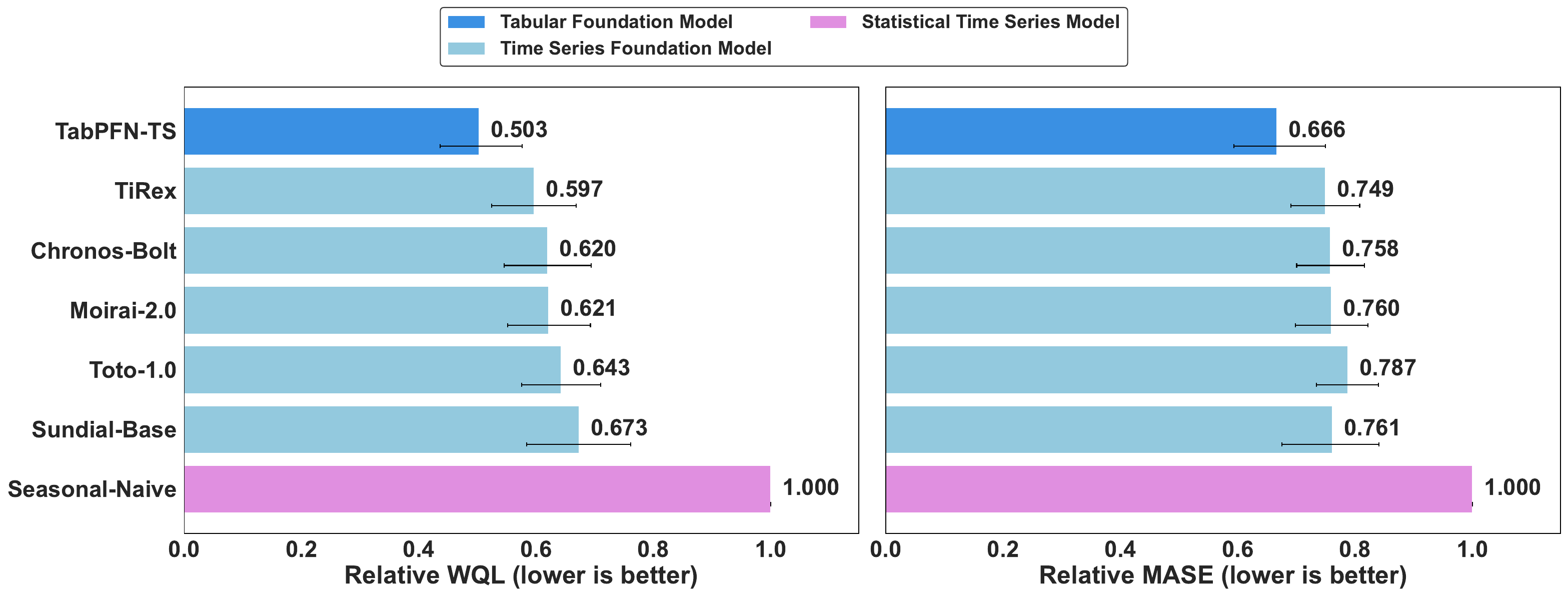}
    \caption{Covariate-informed forecasting performance of \tabpfnts and baseline models on $28$ \texttt{fev-bench} tasks with known dynamic covariates. As the only model that can directly incorporate covariate inputs, \tabpfnts achieves the strongest results, outperforming all other models, including multivariate approaches such as Toto-1.0.
    Scores are normalized by Seasonal Naive and aggregated across datasets; error bars show $95$\% confidence intervals.
    }
    \label{fig:covariate-main-result}
\end{figure}

For this task, we evaluate \tabpfnts on \texttt{fev-bench}~\citep{shchur2025fevbenchrealisticbenchmarktime}, a general time-series forecasting benchmark that includes forecasting tasks where covariates are available.

\paragraph{Datasets.}
\texttt{fev-bench} comprises $100$ forecasting tasks across multiple domains, of which $42$ include covariate information.
We focus on the $30$ tasks where covariates are available in both the historical and future horizons.
This configuration is required by \tabpfnts, which models the dependency between the target series and known covariates and relies on having those covariate values available at prediction time.
When future covariates are missing, the model cannot properly condition its forecasts on relevant contextual signals.
Two of these tasks are excluded because \tabpfnts exceeded the $6$-hour evaluation limit, leaving $28$ tasks for the final evaluation.
Appendix~\ref{sec:append-fev-bench} provides an overview of the selected tasks.

\paragraph{Baselines.}
We follow the standardized evaluation protocol of \texttt{fev-bench}, which reports results only on pretrained time series foundation models.
Deep learning baselines are not included here as \texttt{fev-bench} does not include dataset-specific supervised deep learning baselines.
Therefore, like Section~\ref{sec:univariate-forecasting}, \tabpfnts compares against the following pretrained models: TiRex~\citep{auer2025tirexzeroshotforecastinglong}, Chronos-Bolt (Base)~\citep{ansari2024chronos}, Toto-1.0~\citep{cohen2024tototimeseriesoptimized}, Moirai-2.0~\citep{woo2024unifiedtraininguniversaltime}, and Sundial~\citep{liu2025sundialfamilyhighlycapable}.

\paragraph{Evaluation Metrics.}
Consistently with Section~\ref{sec:univariate-forecasting}, we assess point forecasting accuracy using the Mean Absolute Scaled Error (MASE) and probabilistic forecast accuracy using the Weighted Quantile Loss (WQL).

\paragraph{Results.}
Across the 28 \texttt{fev-bench} tasks with known dynamic covariates, \tabpfnts achieves the strongest overall performance. While strong foundation models, such as TiRex and Chronos-Bolt, remain highly competitive in the univariate setting (see Section~\ref{sec:univariate-forecasting} and \citet{shchur2025fevbenchrealisticbenchmarktime}), \tabpfnts outperforms them when covariates are available. Notably, Toto-1.0, which is designed for the multivariate setting, does not fully recover the benefit obtained when covariates are treated as first-class predictive features like in \tabpfnts.

These results provide strong evidence for the forecasting-as-tabular-regression formulation (Section~\ref{sec:methodology}): it allows \tabpfnts to integrate covariates directly and without architectural changes or fine-tuning.
Together, this positions \tabpfnts as a practical and competitive alternative to current time series foundation models, especially in settings where contextual or external drivers are known and informative.

\section{Ablations} \label{sec:ablations}

We conduct ablation studies to understand how \tabpfnts performs forecasting and to assess the contribution of its components.
Section~\ref{sec:ablation-qualitative-analysis} provides qualitative examples illustrating the model’s forecasting behavior.
Section~\ref{sec:ablation-model-analysis} examines how the model interprets temporal structure within the tabular regression framework to perform forecasting.
Section~\ref{sec:ablation-covariates} studies the integration of covariate information under controlled synthetic setups.
Section~\ref{sec:ablation-tabular-regressor} evaluates alternative tabular regressors to assess the role of the pretrained backbone.
Section~\ref{sec:ablation-featurization} analyzes the impact of the temporal featurization design on overall performance.

\subsection{Qualitative Analysis} \label{sec:ablation-qualitative-analysis}

\begin{figure}
    \centering
    \begin{subfigure}{0.45\textwidth}
        \centering
        \includegraphics[width=0.92\textwidth]{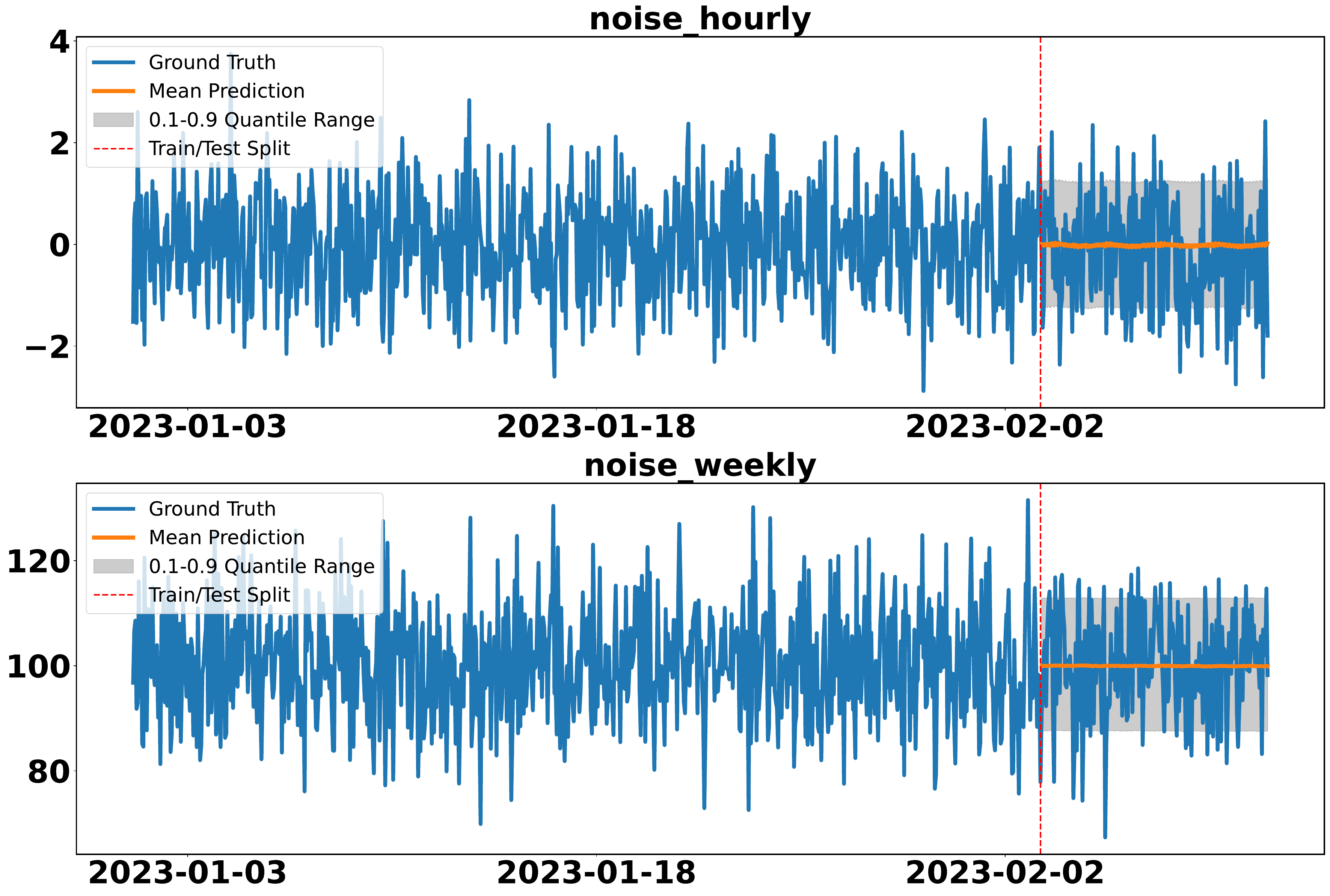}
        \caption{On i.i.d noise}
    \end{subfigure}
    \hspace{0.25cm}
    \begin{subfigure}{0.45\textwidth}
        \centering
        \includegraphics[width=0.92\textwidth]{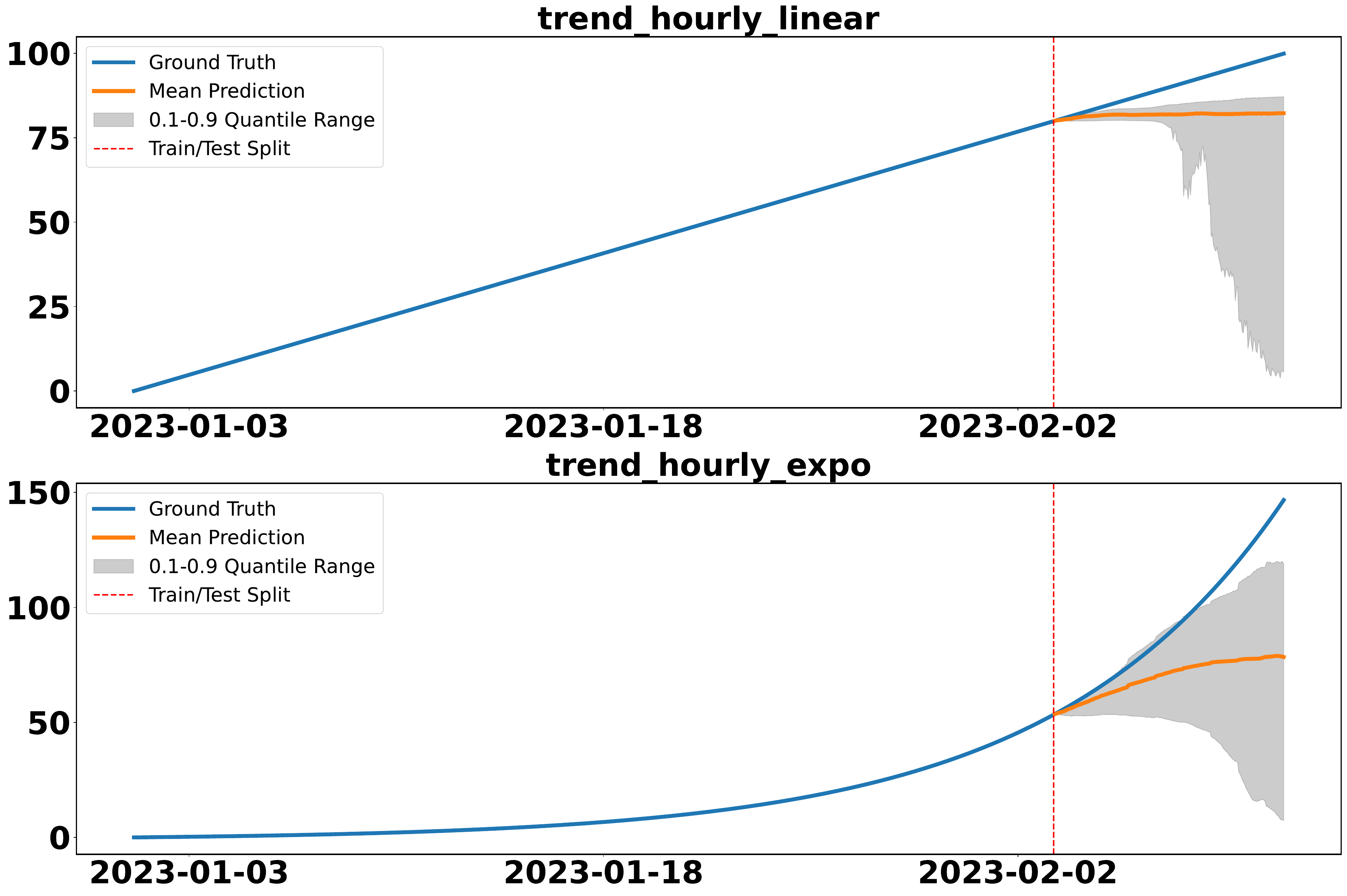}
        \caption{On linear and exponential trends}
    \end{subfigure}
    
    \vspace{0.10cm}
    
    \begin{subfigure}{0.45\textwidth}
        \centering
        \includegraphics[width=0.92\textwidth]{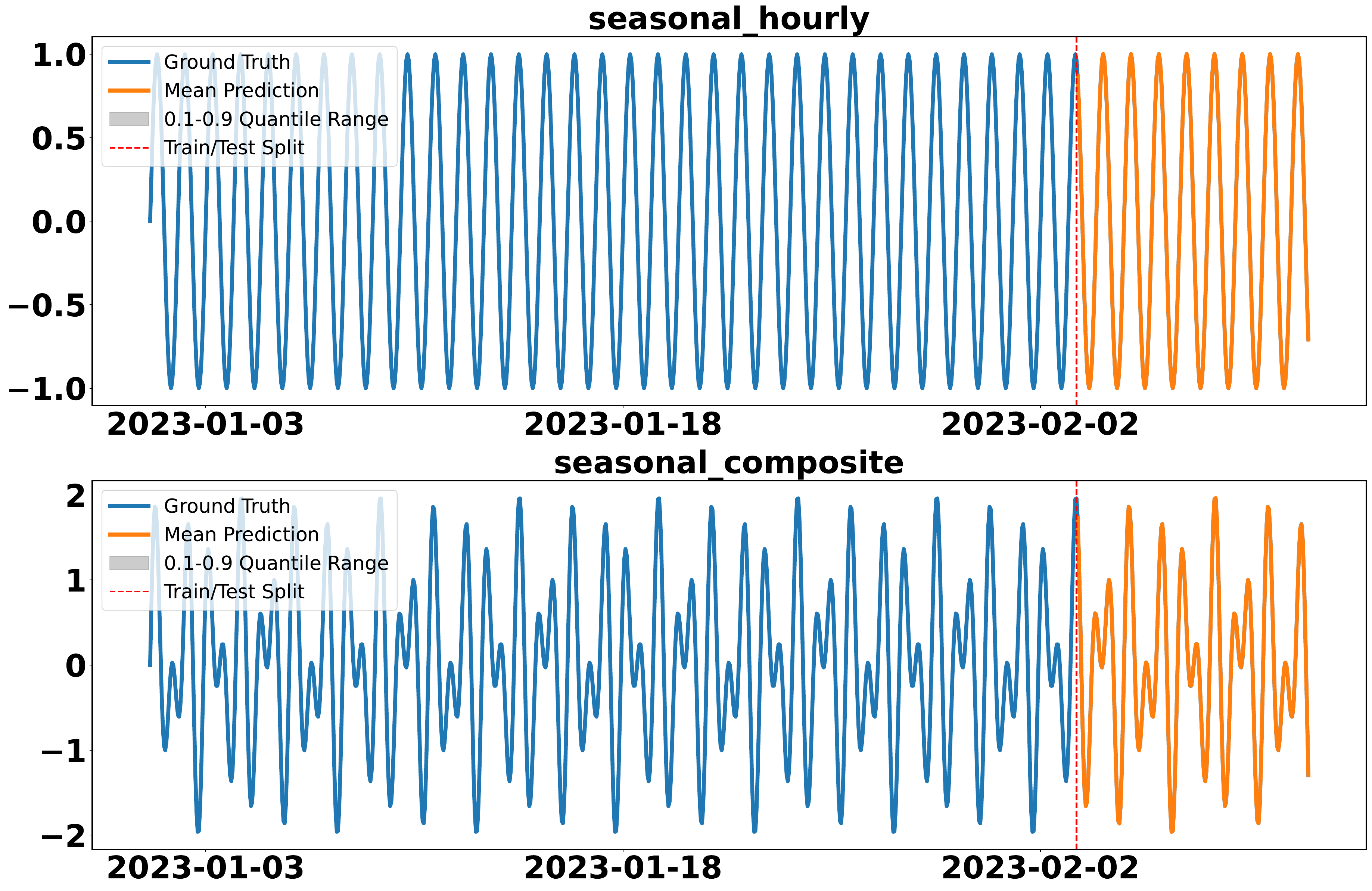}
        \caption{On seasonal patterns}
    \end{subfigure}
    \hspace{0.25cm}
    \begin{subfigure}{0.45\textwidth}
        \centering
        \includegraphics[width=0.92\textwidth]{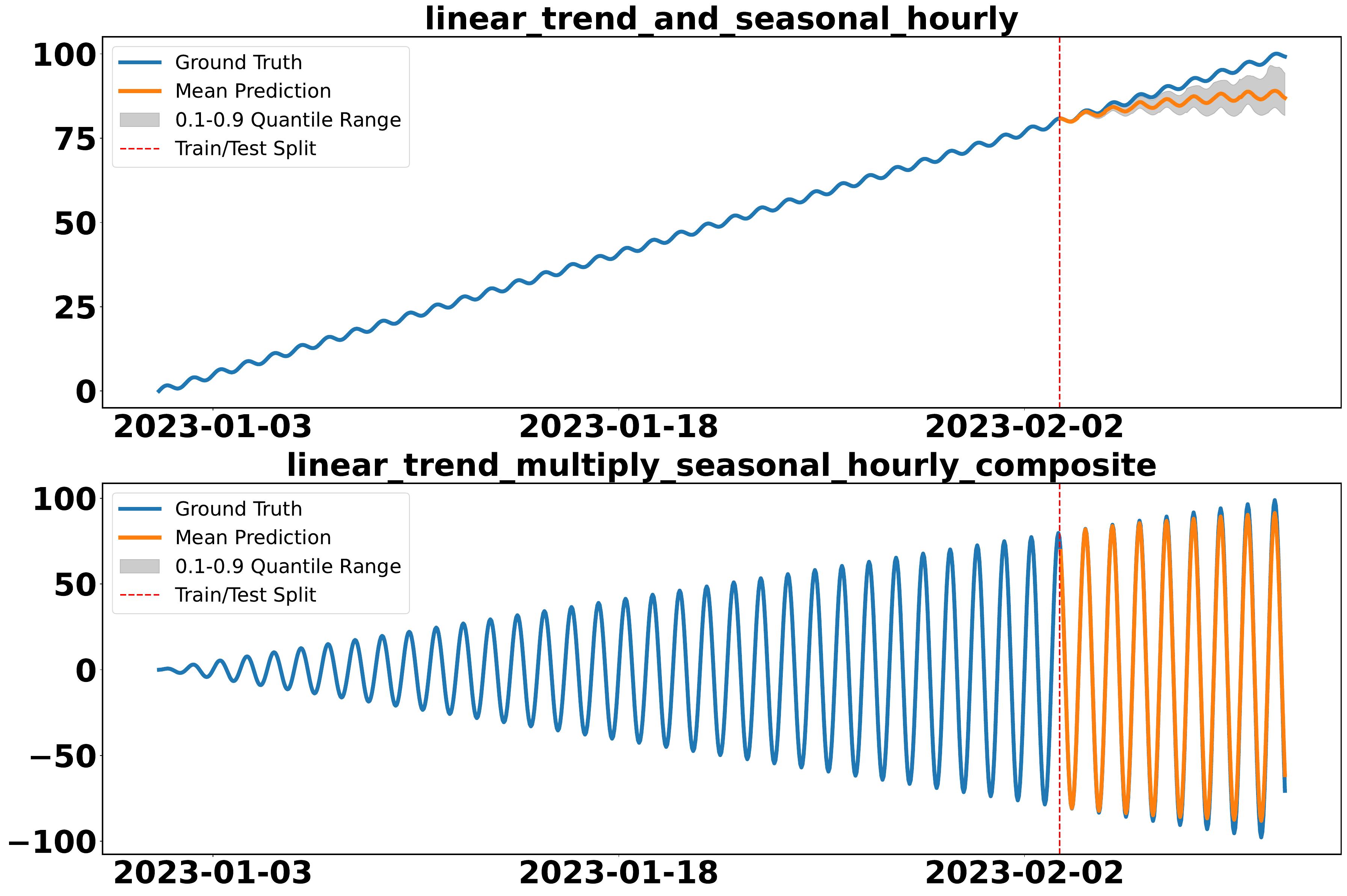}
        \caption{On additive and multiplicative combinations}
    \end{subfigure}    
    \caption{Qualitative performance of \tabpfn across noise, trend, and seasonal structures. The model handles noise and periodicity well but cannot extrapolate linear or exponential trend beyond the observed target range.}
    \label{fig-qualitative}
\end{figure}

We complement the quantitative benchmarks (Section~\ref{sec:experiment}) with qualitative evaluations on controlled synthetic series, following the setups proposed by \citet{ansari2024chronos}.
In all cases, the first $800$ points are provided as context and the model forecasts the next $200$ steps.
These experiments illustrate the characteristic strengths and limitations of \tabpfnts in isolation from dataset-specific confounders.

\paragraph{I.I.D. Noise.}
Figure~\ref{fig-qualitative}a shows the model’s predictions when the input is i.i.d.\ noise sampled from $\mathcal{N}(0,1)$ and $\mathcal{N}(100,10)$ at hourly and weekly resolution.
\tabpfnts does not overfit: it predicts the sample mean and produces predictive intervals that closely align with the corresponding Gaussian quantiles.
This demonstrates stable behavior and
reasonable uncertainty when no temporal structure is present.

\paragraph{Seasonality. }
\tabpfnts models seasonal structure exceptionally well (Figures~\ref{fig-qualitative}c–d).
When periodic patterns are represented in the features, the model not only reconstructs the signal accurately but also produces very tight predictive intervals that closely follow the ground truth.
This strong behaviour is further discussed in Section~\ref{sec:ablation-model-analysis}, where we show that the sinusoidal embedding provides an expressive phase representation that enables precise interpolation of periodic components.

\paragraph{Trend.}
Figure~\ref{fig-qualitative}b illustrates a clear limitation:
\tabpfnts does not extrapolate trends when future values fall outside the range of the conditioning set (i.e. beyond the observed target range).
This behavior appears both for linear and exponential trends.
Since the model operates via interpolation in feature space (discussed later in Section~\ref{sec:ablation-model-analysis}), it cannot extend predictions beyond the observed target domain---an important limitation for time series with sustained growth or decay.

\paragraph{Takeaway.}

These qualitative results reveal a consistent pattern: \tabpfnts behaves reliably when the future targets lie within the range of the conditioning set—capturing noise and seasonal patterns with accurate means and tight uncertainty intervals that align well with the underlying data distribution---but fails to extrapolate when forecasting requires moving beyond the observed target domain, as in linear or exponential trends.
This contrast highlights that the model’s strengths and limitations depend heavily on how temporal structure is represented, motivating the deeper analysis in Section~\ref{sec:ablation-model-analysis} and Section~\ref{sec:limitation-extrapolation}.

\subsection{How \tabpfnts Interprets Temporal Structure} \label{sec:ablation-model-analysis}

To understand why \tabpfn handles seasonality effectively (see Section~\ref{sec:ablation-qualitative-analysis}) despite being pretrained only on tabular data, we analyze how its backbone (\tabpfn) responds to minimal temporal featurization when forecasting periodic signals.
This reveals the precise mechanism by which the model leverages temporal cues.

\paragraph{Setup.}

\begin{figure}
    \centering
    \begin{subfigure}[b]{0.49\textwidth}
        \centering
        \includegraphics[width=\textwidth]{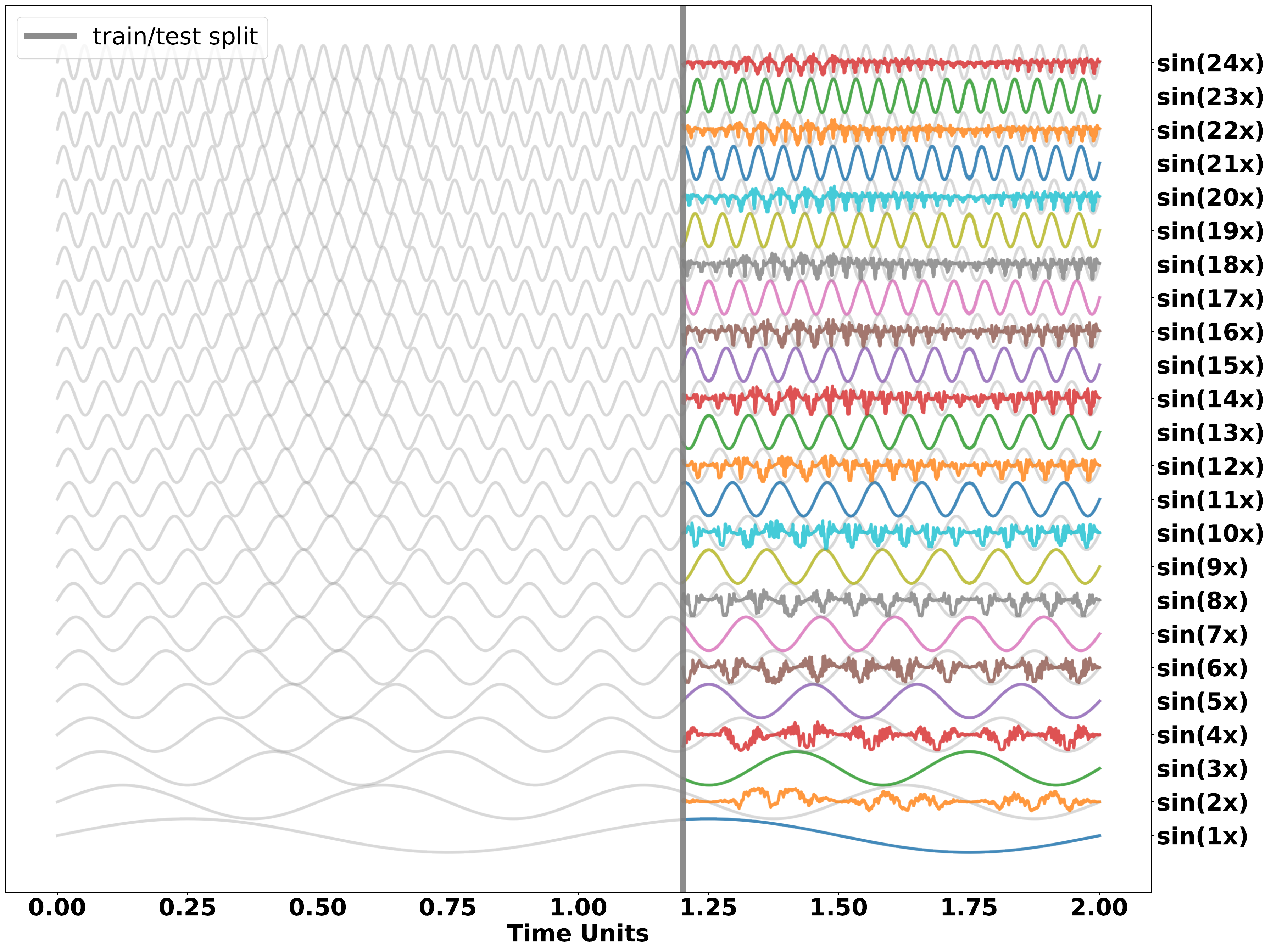}
        \caption{Conditioning on $\sin(t)$ only}
    \end{subfigure}
    \hfill
    \begin{subfigure}[b]{0.49\textwidth}
        \centering
        \includegraphics[width=\textwidth]{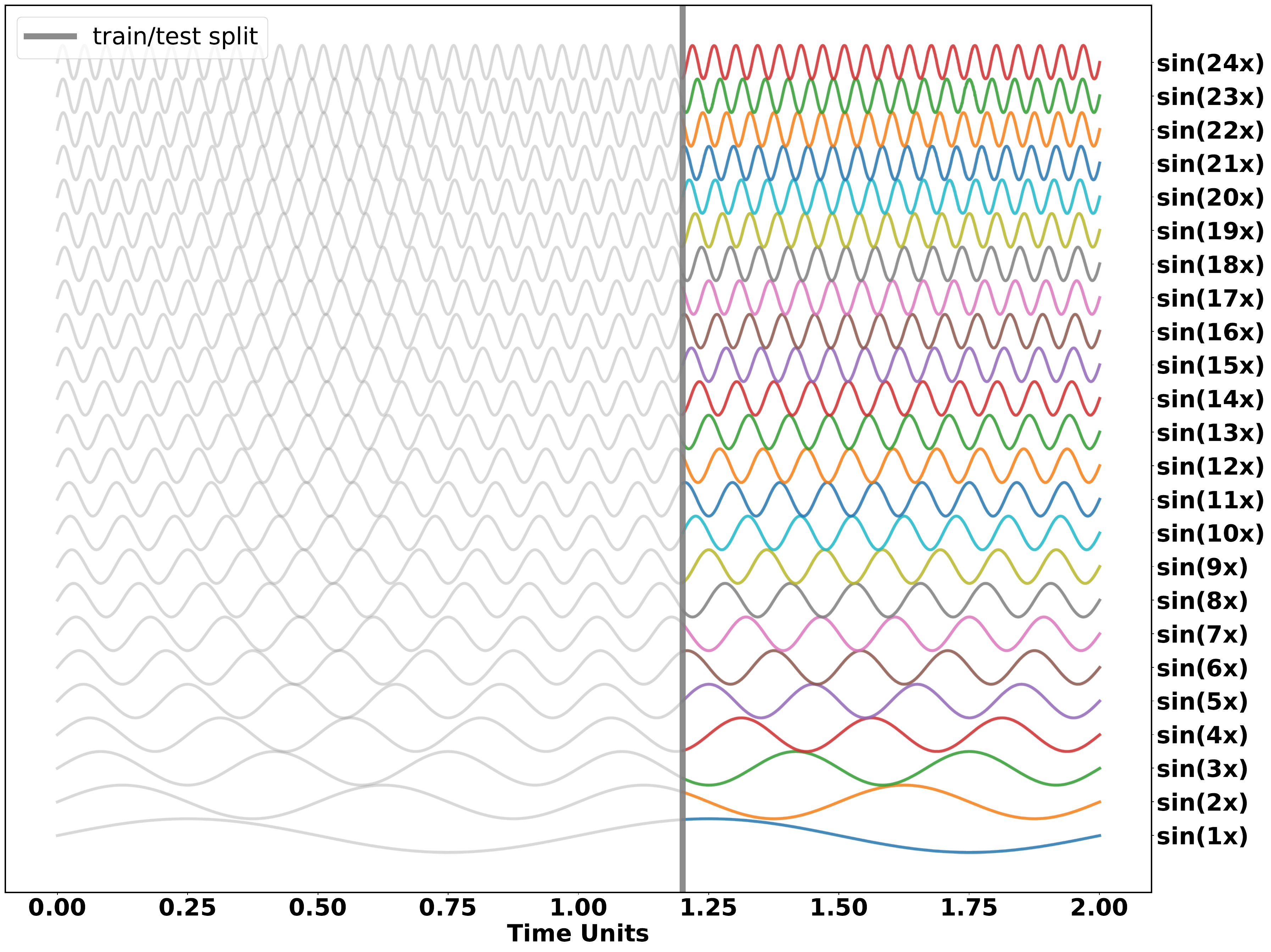}
        \caption{Conditioning on $\sin(t)$ and $\cos(t)$}
    \end{subfigure}
    \caption{
    Reconstruction of sinusoidal signals under different temporal embeddings. \tabpfn is evaluated on $\sin(n t)$ for $n = 1,\dots,24$.
    (a) With $\sin(t)$ as the sole temporal feature, the non-injective embedding causes different time points to collapse to the same feature representation, and the model therefore reconstructs odd but not even harmonics.
    (b) With $\{\sin(t),\cos(t)\}$, the injective embedding allows the model to interpolate reliably in this phase space and reconstruct all harmonics.
    These results suggest that periodic generalization arises from interpolation in an expressive temporal embedding.
    }
    \label{fig:sin-cos-generalization}
\end{figure}

We generate target series of the form $y_t = \sin(n t)$ for $n = 1,\dots,24$.
The model is evaluated under two featurization settings:
(i) using $\sin(t)$ alone, and
(ii) the pair $\left(\sin(t), \cos(t)\right)$.
This enables us to test whether the model can recover higher-order harmonics from a low-frequency base.
We then extend this to composite signals formed by summing multiple sinusoids with random frequencies, amplitudes, and phases.

\paragraph{Findings.}
With only $\sin(t)$, the model reconstructs odd harmonics but fails entirely on even ones (Figure~\ref{fig:sin-cos-generalization}a).
However, with the pair $\left(\sin(t), \cos(t)\right)$, the model accurately recovers the higher-order harmonics.

We hypothesize that this behavior stems from the \textbf{injectivity} of the temporal mapping.
When only $\sin(t)$ is present, the mapping $t \mapsto \sin(t)$ is not injective; distinct time points (e.g., on the rising vs. falling edges of a cycle) share identical feature values, making local interpolation inherently ambiguous.
The inclusion of $\cos(t)$ embeds time onto the unit circle, creating an injective mapping that uniquely identifies the phase (see Section~\ref{sec:app-phase-ambiguity} for a detailed symmetry analysis).
Under this representation, \tabpfn accurately recovers harmonics across a wide frequency range (Figure~\ref{fig:sin-cos-generalization}b), and successfully reconstructs complex composite signals (Figure~\ref{fig:comp-waves-generalization}).

We also observe a gradual deterioration in accuracy as the frequency increases (Figure~\ref{fig:sin-cos-generalization-full} and~\ref{fig:sin-cos-generalization-smape}).
This is consistent with the finite resolution of the temporal embedding and classical aliasing effect~\citep{oppenheim1999discrete}: as frequency increases, the angular distance between successive samples in the coordinate system decreases, making adjacent phases increasingly difficult to be distinguished.

While we considered an alternative hypothesis where the model performs global functional approximation (e.g., via Chebyshev expansion), our preliminary experiments (see Appendix~\ref{sec:app-functional-vs-retrieval}) suggest that \tabpfn's behavior is more consistent with a retrieval mechanism.

\paragraph{Takeaway.} These results suggest that when supplied with an injective sinusoidal embedding of time, \tabpfnts behaves as a \textbf{meta-learned retrieval engine}.
It interpolates effectively within a feature space where time has been mapped to a unique circular coordinate (i.e. phase-space), enabling phase-aware pattern matching against the observed history in the support set.

\subsection{How \tabpfnts Utilizes Covariates} \label{sec:ablation-covariates}

\begin{figure}
    \centering
    \includegraphics[width=1.0\linewidth]{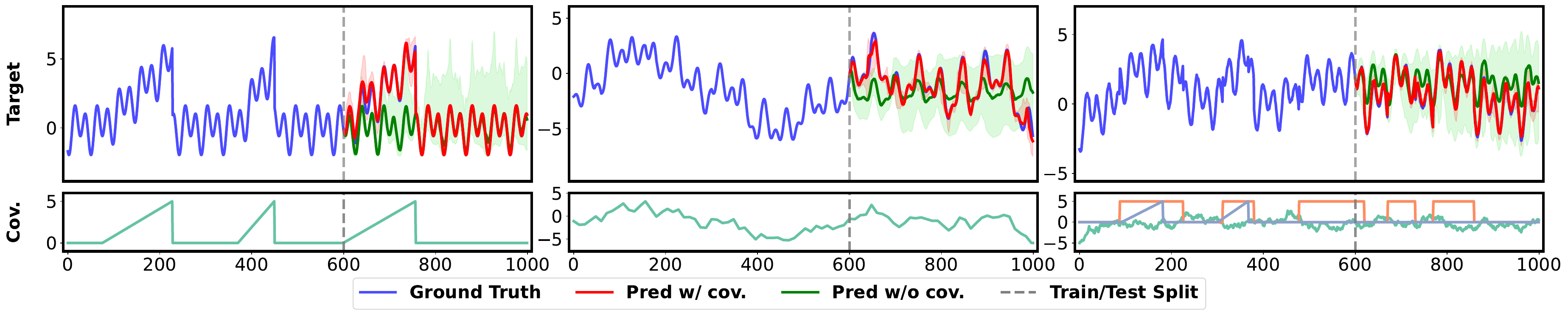}
    \caption{Forecasts of \tabpfnts on synthetic time series augmented by time-varying covariates. Median predictions are shown in red (conditioned on covariates) and green (unconditioned), with shaded bands indicating the $10^{th}$ - $90^{th}$ percentiles.}
    \label{fig-covariate-visualization}
\end{figure}

In Section~\ref{sec:covariate-forecasting}, we showed that \tabpfn achieves the strongest performance on covariate-informed forecasting tasks.
To better understand this effect, we conduct controlled synthetic experiments where the informativeness and interaction structure of each covariate are known.

\paragraph{Synthetic Setup.}
Similar to \citet{arango2025chronosx} and \citet{auer2025cosmic}, we generate synthetic time series where the relationship between the target and the covariates is fully controlled.
Each series consists of a stationary stochastic process \citep{brockwellStationaryProcesses2016}—implemented here as a sum of sinusoids with random amplitudes, phases, and periods—augmented with one or more covariates that follow common temporal patterns such as linear trends, pulses, ramps, or stochastic drifts.
Covariates influence the target through either additive or multiplicative coupling, resulting in ten interpretable synthetic classes (Figure~\ref{fig:covariate_type_visualization}).

\paragraph{Evaluation Protocol.}
For each class, we evaluate \tabpfnts under two conditions:
(i) using temporal features only, and
(ii) using both temporal features and covariates.
This isolates the incremental contribution of covariate information.

\paragraph{Findings.}

\begin{figure}
    \centering
    \includegraphics[width=0.9\linewidth]{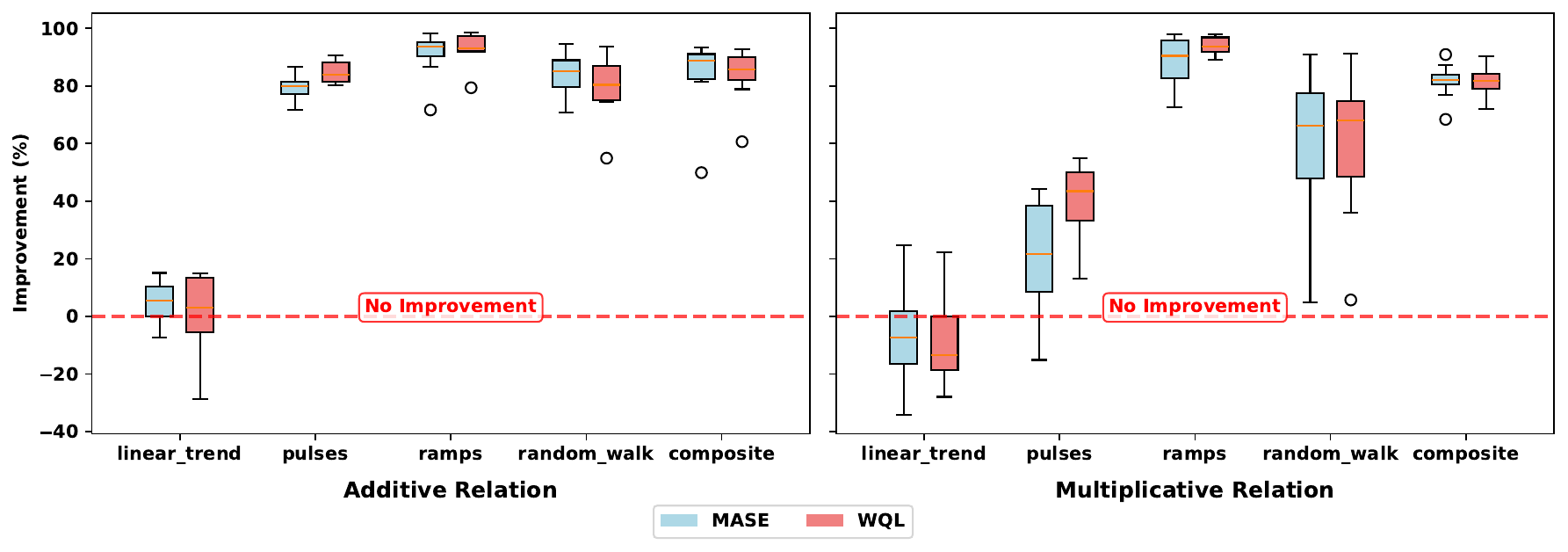}
    \caption{Improvement when incorporating covariates across synthetic classes. Boxplots show relative improvement (in percentage, \%) in MASE (blue) and WQL (red). Most covariate types lead to substantial improvement, particularly for pulses, ramps, random walks, and composite signals. In contrast, for both additive and multiplicative linear trend series, covariates fail to improve forecasts, reflecting the model’s limitation in extrapolating linear growth.}
    \label{fig:covariate_results}
\end{figure}

Figure~\ref{fig:covariate_results} shows that incorporating covariates substantially improves forecasting accuracy for most synthetic classes—pulses, ramps, random walks, and composite signals—reducing both MASE and WQL and producing sharper uncertainty estimates (Figure~\ref{fig:covariate_type_visualization}).
These gains are consistent with our benchmark results and indicate that \tabpfnts can reliably exploit informative auxiliary signals when their effects lie within the range covered by the conditioning set (detailed in Section~\ref{sec:limitation-extrapolation}).

The main exception is the \textit{linear-trend} setting, where covariates provide little benefit because the future targets fall outside the distribution observed in the context; \tabpfnts does not extrapolate linear growth (discussed later in Section~\ref{sec:limitation-extrapolation}).
Overall, these controlled experiments show that \tabpfnts effectively leverages covariates across a wide range of scenarios, with extrapolation remaining the primary limitation.

\subsection{Choice of Tabular Backbone} \label{sec:ablation-tabular-regressor}

\begin{wrapfigure}{r}{0.38\textwidth}
    \vspace{-5.2em}
    \begin{minipage}{\linewidth}
        \centering
        \includegraphics[width=\linewidth]{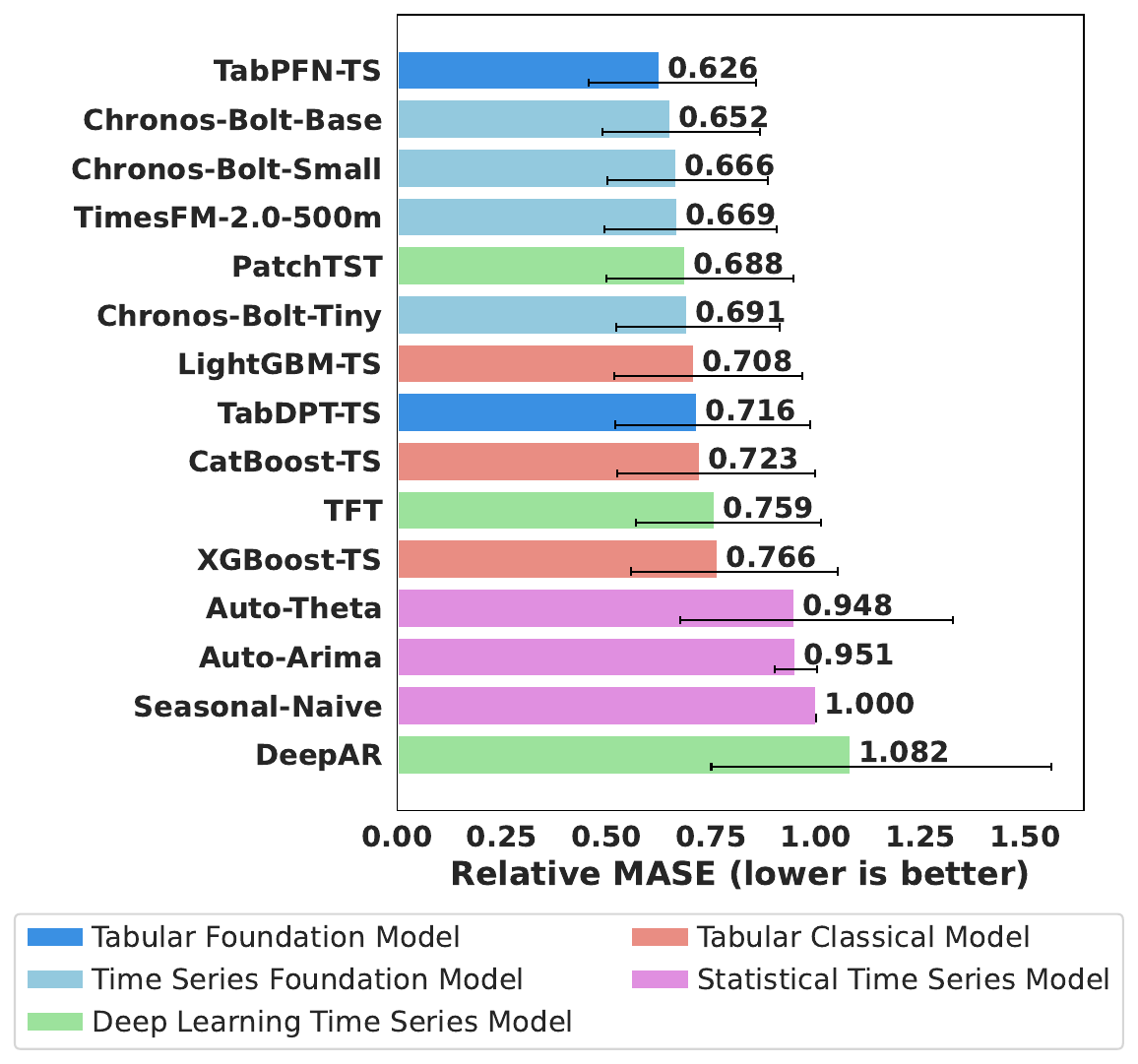}
        \vspace{-2.35em}
        \caption{
        MASE comparison of different tabular backbones on $67$ GIFT-Eval tasks.
        \tabpfnts outperforms all alternatives by a clear margin, suggesting \tabpfn's synthetic pretraining contributes beyond featurization alone.
        }
        \label{fig:tabdpt-result}
    \end{minipage}
    \vspace{-3.0em}
\end{wrapfigure}

To assess whether our forecasting formulation depends on the choice of tabular backbone, we evaluate two categories of alternative backbones under the same temporal featurization: (i) classical tabular regressors XGBoost~\citep{Chen_2016_xgboost}, LightGBM~\citep{ke_lightgbm}, and CatBoost~\citep{prokhorenkova2018catboost}; and (ii) TabDPT~\citep{ma2025tabdptscalingtabularfoundation}, one of the few open-source tabular foundation models that supports regression~\citep{erickson2025tabarenalivingbenchmarkmachine}.
As most of these models only produce point predictions, we evaluate on point forecasting (MASE) across $61$ of the $97$ GIFT-Eval tasks (the remainder exceeded our compute budget).



Figure~\ref{fig:tabdpt-result} shows that all the alternative backbones achieve meaningful forecasting performance, yet fall short of TabPFN-TS.
We attribute this to the pretraining: \tabpfn TabPFN-v2 is trained on millions of synthetic tasks spanning diverse functional relationships and noise regimes, while the gradient-boosted models are fit from scratch per series and TabDPT is pretrained on real-world tabular data with narrower task diversity. 


Taken together, this experiment suggests that: (i) the \textbf{temporal featurization generalizes across tabular foundation models}---different tabular foundation models can perform zero-shot forecasting under the same representation; and (ii) the \textbf{regression backbone matters}---models pretrained on broader and more diverse task distributions (such as \tabpfn) appear to transfer more effectively to the time-series forecasting.

\subsection{Impact of Temporal Featurization} \label{sec:ablation-featurization}

To assess the contribution of the each temporal feature type (Section~\ref{sec:from-ts-to-tabular}), we evaluate \tabpfnts under different combinations of these features on the $81$ smallest tasks (out of 97) in GIFT-Eval.

\begin{figure}
    \centering
    \includegraphics[width=0.8\textwidth]{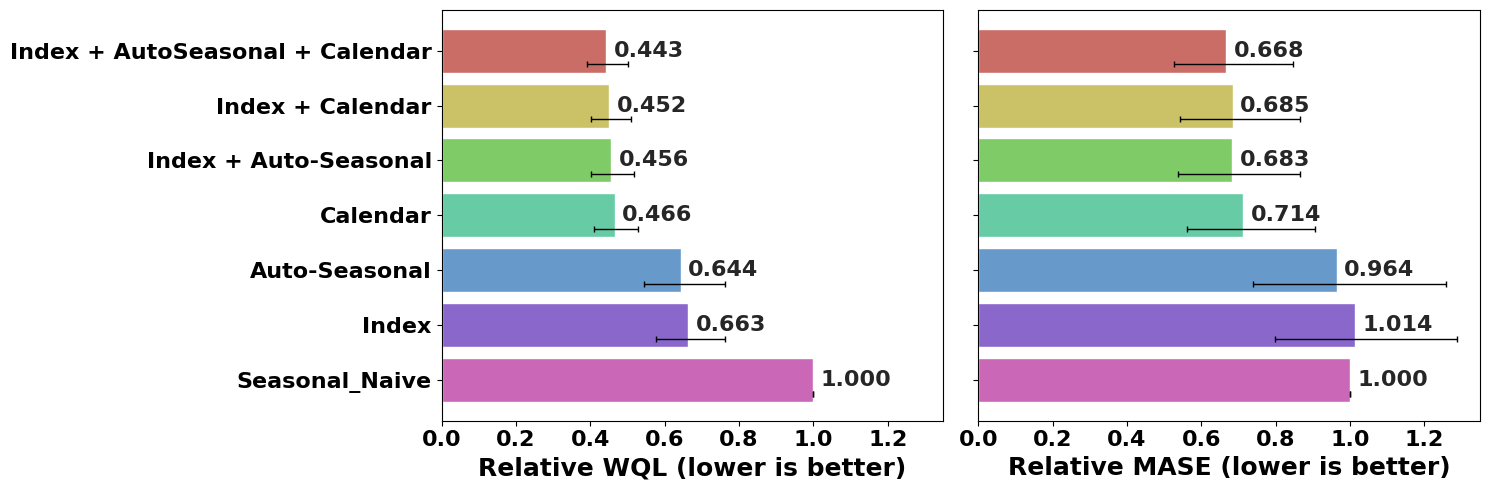}
    \caption{Ablation of temporal featurization components in \tabpfnts.
    Relative WQL (left) and MASE (right) show that all components contribute meaningfully to performance.
    Using only the index or only the auto-seasonal features yields substantially worse results, while combining all feature types achieves the best accuracy, underscoring the value of the proposed featurization scheme.
    }
    \label{fig:feature-ablation}
\end{figure}

The results in Figure~\ref{fig:feature-ablation} highlight the importance of \textbf{encoding time progression}.
Without an explicit index, calendar features provide both periodic structure and a weak notion of temporal progression, since some calendar attributes present monotonically within the timeline.
Automatic seasonal features, by contrast, encode oscillatory components but offer no information about global temporal progression.
When the index is present, it provides the primary temporal axis, while both calendar and automatic seasonal features contribute the complementary periodic structure.

The best results are obtained when all feature types are combined.
This indicates that calendar and automatic seasonal features capture different---and complementary---seasonal patterns.
Together with the index, they provide a broad and expressive temporal representation.

Overall, \tabpfnts benefits from temporal features that jointly encode \textbf{monotonic progression} and \textbf{multi-scale seasonality}.
The framework is naturally extensible, allowing practitioners to incorporate domain-specific periodic features to further improve forecasting accuracy when such knowledge is available.

\section{Limitation}
\subsection{Extrapolation Beyond the Conditioning Range}
\label{sec:limitation-extrapolation}

The analyses in Sections~\ref{sec:ablation-qualitative-analysis} and \ref{sec:ablation-covariates} reveal a persistent limitation of \tabpfnts:
it struggles in extrapolating trends, where the future target values lie outside the range observed in the conditioning set.

This behavior is a direct consequence of how \tabpfn is pretrained.
During pretraining, each synthetic task is generated by sampling a tabular dataset and randomly partitioning it into a context set $(X, y)$ and a query set $X’$.
The model is then trained to predict the corresponding targets $y’$.
Since both context and query are sampled from the same generating distribution, the query targets typically lie within the convex hull of the observed targets.
Consequently, \tabpfn is primarily optimized for conditional interpolation and lacks the inductive bias required for unbounded trend extrapolation.

\tabpfnts inherits this inductive bias: when future targets lie outside the range of observed target, it has little basis---under its interpolation-oriented objective---to project the trajectory further.
Predictions therefore remain within or close to the span of the observed targets.

In summary, \tabpfn is a strong conditional interpolator but lacks a mechanism for systematic trend extrapolation.
This limitation is structural, stemming from the pretraining objective, rather than from the tabular regression formulation used for forecasting.

\subsection{Inference Time} \label{sec:inference-time}

\begin{wrapfigure}{r}{0.40\textwidth}
    \vspace{-1.5em}
    \begin{minipage}{\linewidth}
        \centering
        \includegraphics[width=0.8\linewidth]{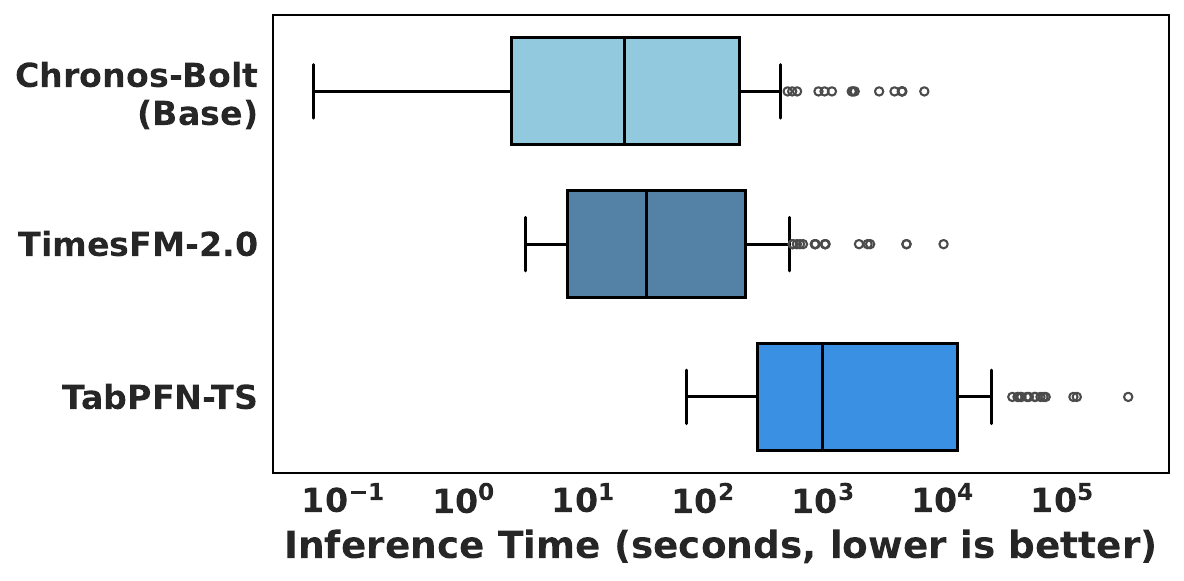}
        \caption{Per-task inference time on 97 GIFT-Eval tasks (log scale; lower is better). Chronos-Bolt-Base and TimesFM-2.0 run on one NVIDIA T4; \tabpfnts runs across four T4s, and we report time normalized to one GPU.}
        \label{fig:inference-time-comparison}
    \end{minipage}
    \vspace{-1.5em}
\end{wrapfigure}

A second practical limitation of \tabpfnts is its higher computational cost at inference.
Although the model supports data-parallel evaluation across multiple GPUs, this only increases throughput but does not reduce the per-time-series cost.
Figure~\ref{fig:inference-time-comparison} reports single-GPU-equivalent inference times, normalized by GPU count to allow direct comparison with Chronos-Bolt-Base and TimesFM-2.0.
Under equal hardware assumptions, \tabpfnts requires roughly $30\times$ more inference time.

This overhead arises from how \tabpfn performs prediction.
In tabular settings, one forward pass produces the prediction for an entire dataset.
In time-series forecasting, however, a dataset contains many individual series---often hundreds or thousands---and each series becomes an independent query instance under our formulation.
\tabpfnts must therefore execute a separate forward pass for every series.

Moreover, \tabpfnts retains the full-attention architecture of \tabpfn, which incurs $O(L^{2})$ computational complexity in sequence length $L$.
Recent time-series foundation models mitigate this cost by processing large batches of series and by using architectural techniques such as patching or lightweight decoders that reduce dependence on sequence length.
Consequently, they achieve substantially faster inference.

In summary, the higher inference cost of \tabpfnts arises structurally from the combination of (i) its one-forward-pass-per-series formulation and (ii) its quadratic attention mechanism.
Improving inference efficiency---through patching, batching, and/or architectural refinement (discussed in Section~\ref{sec:inference-optimization})---remains an important direction for future work.

\section{Conclusion}

\paragraph{Summary of Contributions.}
This work demonstrates that a tabular foundation model---pretrained solely on synthetic, non-temporal data---can be adapted into a competitive zero-shot forecaster through a simple reformulation of time-series prediction as tabular regression. Combining \tabpfn with a lightweight temporal featurization yields \tabpfnts, a compact ($11$M) model that requires no time-series–specific pretraining or fine-tuning, yet achieves strong performance on univariate forecasting and state-of-the-art accuracy on covariate-informed tasks.

\paragraph{Key Empirical Insights.}
Our ablations highlight three central findings.
(1) Temporal featurization is essential: performance improves substantially when monotonic progression and complementary periodicities are jointly encoded.
(2) Pretraining induces distinct inductive biases: \tabpfnts excels at interpolation inside the observed target range, but---consistent with its pretraining objective---fails when forecasting requires systematic extrapolation.
(3) Forecasting capability is model-agnostic: substituting the backbone with TabDPT still produces meaningful zero-shot forecasts, indicating that the overall formulation generalizes across tabular foundation models.

\paragraph{Implications.}
These results indicate that tabular foundation models constitute a surprisingly effective and data-efficient alternative to specialized time-series architectures. The tabular formulation naturally unifies univariate and covariate-aware forecasting, treats all inputs as general features, and enables probabilistic forecasting without architectural modifications to the backbone.

\paragraph{Vision and Future Directions.}
More broadly, this work suggests a promising direction toward unified foundation models for structured data that operate seamlessly over tabular, temporal, and hybrid inputs. Addressing the core limitation---the inability to extrapolate trends beyond the conditioning range---may require new pretraining objectives that blend interpolation, controlled extrapolation, and explicit handling of out-of-support targets. Such advances open a path toward compact, domain-agnostic predictors that bridge long-standing gaps between tabular learning and time-series forecasting.




\bibliography{main}

@article{auer2025cosmic,
  title={Zero-Shot Time Series Forecasting with Covariates via In-Context Learning},
  author={Auer, Andreas and Parthipan, Raghul and Mercado, Pedro and Ansari, Abdul Fatir and Stella, Lorenzo and Wang, Bernie and Bohlke-Schneider, Michael and Rangapuram, Syama Sundar},
  journal={arXiv preprint arXiv:2506.03128},
  year={2025}
}

@article{hann-window,
  title   = {On the use of windows for harmonic analysis with the discrete Fourier transform},
  author  = {Harris, Fredric J.},
  journal = {Proceedings of the IEEE},
  volume  = {66},
  number  = {1},
  pages   = {51--83},
  year    = {1978}
}

@book{discrete-signal-processing,
  title     = {Discrete-Time Signal Processing},
  author    = {Oppenheim, Alan V. and Schafer, Ronald W.},
  publisher = {Prentice Hall},
  year      = {1989}
}

@misc{erickson2025tabarenalivingbenchmarkmachine,
      title={TabArena: A Living Benchmark for Machine Learning on Tabular Data}, 
      author={Nick Erickson and Lennart Purucker and Andrej Tschalzev and David Holzmüller and Prateek Mutalik Desai and David Salinas and Frank Hutter},
      year={2025},
      eprint={2506.16791},
      archivePrefix={arXiv},
      primaryClass={cs.LG},
      url={https://arxiv.org/abs/2506.16791}, 
}

@misc{auer2025tirexzeroshotforecastinglong,
      title={TiRex: Zero-Shot Forecasting Across Long and Short Horizons with Enhanced In-Context Learning}, 
      author={Andreas Auer and Patrick Podest and Daniel Klotz and Sebastian Böck and Günter Klambauer and Sepp Hochreiter},
      year={2025},
      eprint={2505.23719},
      archivePrefix={arXiv},
      primaryClass={cs.LG},
      url={https://arxiv.org/abs/2505.23719}, 
}

@misc{shchur2025fevbenchrealisticbenchmarktime,
      title={fev-bench: A Realistic Benchmark for Time Series Forecasting}, 
      author={Oleksandr Shchur and Abdul Fatir Ansari and Caner Turkmen and Lorenzo Stella and Nick Erickson and Pablo Guerron and Michael Bohlke-Schneider and Yuyang Wang},
      year={2025},
      eprint={2509.26468},
      archivePrefix={arXiv},
      primaryClass={cs.LG},
      url={https://arxiv.org/abs/2509.26468}, 
}

@article{gift-eval,
      title={GIFT-Eval: A Benchmark For General Time Series Forecasting Model Evaluation}, 
      author={Taha Aksu and Gerald Woo and Juncheng Liu and Xu Liu and Chenghao Liu and Silvio Savarese and Caiming Xiong and Doyen Sahoo},
      journal = {arxiv preprint arxiv:2410.10393},
      year={2024},
}

@inproceedings{hollmanntabpfn,
  title={TabPFN: A Transformer That Solves Small Tabular Classification Problems in a Second},
  author={Hollmann, Noah and M{\"u}ller, Samuel and Eggensperger, Katharina and Hutter, Frank},
  year={2023},
  booktitle={The Eleventh International Conference on Learning Representations}
}

@article{lim2021temporal,
  title={Temporal fusion transformers for interpretable multi-horizon time series forecasting},
  author={Lim, Bryan and Ar{\i}k, Sercan {\"O} and Loeff, Nicolas and Pfister, Tomas},
  journal={International Journal of Forecasting},
  volume={37},
  number={4},
  pages={1748--1764},
  year={2021},
  publisher={Elsevier}
}

@article{ansari2024chronos,
  title={Chronos: Learning the language of time series},
  author={Ansari, Abdul Fatir and Stella, Lorenzo and Turkmen, Caner and Zhang, Xiyuan and Mercado, Pedro and Shen, Huibin and Shchur, Oleksandr and Rangapuram, Syama Sundar and Arango, Sebastian Pineda and Kapoor, Shubham and others},
  journal={arXiv preprint arXiv:2403.07815},
  year={2024}
}

@misc{garza2022statsforecast,
    author = {Garza, Federico and Mergenthaler Canseco, Max and Challú, Cristian and Olivares, Kin G.},
    title = {{StatsForecast}: Lightning fast forecasting with statistical and econometric models},
    year={2022},
    howpublished={{PyCon} Salt Lake City, Utah, US 2022},
    url={https://github.com/Nixtla/statsforecast}
}

@article{salinas2020deepar,
  title={DeepAR: Probabilistic forecasting with autoregressive recurrent networks},
  author={Salinas, David and Flunkert, Valentin and Gasthaus, Jan and Januschowski, Tim},
  journal={International journal of forecasting},
  volume={36},
  number={3},
  pages={1181--1191},
  year={2020},
  publisher={Elsevier}
}

@inproceedings{agtimeseries,
  title={{AutoGluon-TimeSeries}: {AutoML} for Probabilistic Time Series Forecasting},
  author={Shchur, Oleksandr and Turkmen, Caner and Erickson, Nick and Shen, Huibin and Shirkov, Alexander and Hu, Tony and Wang, Yuyang},
  booktitle={International Conference on Automated Machine Learning},
  year={2023}
}

@article{hollmann2025accurate,
  title={Accurate predictions on small data with a tabular foundation model},
  author={Hollmann, Noah and M{\"u}ller, Samuel and Purucker, Lennart and Krishnakumar, Arjun and K{\"o}rfer, Max and Hoo, Shi Bin and Schirrmeister, Robin Tibor and Hutter, Frank},
  journal={Nature},
  volume={637},
  number={8045},
  pages={319--326},
  year={2025},
  publisher={Nature Publishing Group UK London}
}

@misc{arango2025chronosx,
      title={ChronosX: Adapting Pretrained Time Series Models with Exogenous Variables}, 
      author={Sebastian Pineda Arango and Pedro Mercado and Shubham Kapoor and Abdul Fatir Ansari and Lorenzo Stella and Huibin Shen and Hugo Senetaire and Caner Turkmen and Oleksandr Shchur and Danielle C. Maddix and Michael Bohlke-Schneider and Yuyang Wang and Syama Sundar Rangapuram},
      year={2025},
      eprint={2503.12107},
      archivePrefix={arXiv},
      primaryClass={cs.LG},
      url={https://arxiv.org/abs/2503.12107}, 
}

@misc{ma2025tabdptscalingtabularfoundation,
      title={TabDPT: Scaling Tabular Foundation Models on Real Data}, 
      author={Junwei Ma and Valentin Thomas and Rasa Hosseinzadeh and Hamidreza Kamkari and Alex Labach and Jesse C. Cresswell and Keyvan Golestan and Guangwei Yu and Anthony L. Caterini and Maksims Volkovs},
      year={2025},
      eprint={2410.18164},
      archivePrefix={arXiv},
      primaryClass={cs.LG},
      url={https://arxiv.org/abs/2410.18164}, 
}

@book{oppenheim1999discrete,
  title={Discrete-Time Signal Processing},
  author={Oppenheim, A.V.},
  isbn={9788131704929},
  series={Pearson education signal processing series},
  url={https://books.google.de/books?id=geTn5W47KEsC},
  year={1999},
  publisher={Pearson Education}
}

@misc{woo2024unifiedtraininguniversaltime,
      title={Unified Training of Universal Time Series Forecasting Transformers}, 
      author={Gerald Woo and Chenghao Liu and Akshat Kumar and Caiming Xiong and Silvio Savarese and Doyen Sahoo},
      year={2024},
      eprint={2402.02592},
      archivePrefix={arXiv},
      primaryClass={cs.LG},
      url={https://arxiv.org/abs/2402.02592}, 
}

@incollection{brockwellStationaryProcesses2016,
  title = {Stationary {{Processes}}},
  booktitle = {Introduction to {{Time Series}} and {{Forecasting}}},
  author = {Brockwell, Peter J. and Davis, Richard A.},
  editor = {Brockwell, Peter J. and Davis, Richard A.},
  year = 2016,
  pages = {39--71},
  publisher = {Springer International Publishing},
  address = {Cham},
  doi = {10.1007/978-3-319-29854-2_2},
  urldate = {2025-11-15},
  isbn = {978-3-319-29854-2},
  langid = {english}
}

@inproceedings{das2024timesfm,
  title={A decoder-only foundation model for time-series forecasting},
  author={Das, Abhimanyu and Kong, Weihao and Sen, Rajat and Zhou, Yichen},
  booktitle={Forty-first International Conference on Machine Learning},
  year={2024}
}

@misc{liu2025sundialfamilyhighlycapable,
      title={Sundial: A Family of Highly Capable Time Series Foundation Models}, 
      author={Yong Liu and Guo Qin and Zhiyuan Shi and Zhi Chen and Caiyin Yang and Xiangdong Huang and Jianmin Wang and Mingsheng Long},
      year={2025},
      eprint={2502.00816},
      archivePrefix={arXiv},
      primaryClass={cs.LG},
      url={https://arxiv.org/abs/2502.00816}, 
}

@misc{cohen2024tototimeseriesoptimized,
      title={Toto: Time Series Optimized Transformer for Observability}, 
      author={Ben Cohen and Emaad Khwaja and Kan Wang and Charles Masson and Elise Ramé and Youssef Doubli and Othmane Abou-Amal},
      year={2024},
      eprint={2407.07874},
      archivePrefix={arXiv},
      primaryClass={cs.LG},
      url={https://arxiv.org/abs/2407.07874}, 
}

@article{prokhorenkova2018catboost,
  title={CatBoost: unbiased boosting with categorical features},
  author={Prokhorenkova, Liudmila and Gusev, Gleb and Vorobev, Aleksandr and Dorogush, Anna Veronika and Gulin, Andrey},
  journal={Advances in neural information processing systems},
  volume={31},
  year={2018}
}

@inproceedings{Chen_2016_xgboost, series={KDD ’16},
   title={XGBoost: A Scalable Tree Boosting System},
   url={http://dx.doi.org/10.1145/2939672.2939785},
   DOI={10.1145/2939672.2939785},
   booktitle={Proceedings of the 22nd ACM SIGKDD International Conference on Knowledge Discovery and Data Mining},
   publisher={ACM},
   author={Chen, Tianqi and Guestrin, Carlos},
   year={2016},
   month=aug, pages={785–794},
   collection={KDD ’16}   
}

@inproceedings{ke_lightgbm,
 author = {Ke, Guolin and Meng, Qi and Finley, Thomas and Wang, Taifeng and Chen, Wei and Ma, Weidong and Ye, Qiwei and Liu, Tie-Yan},
 booktitle = {Advances in Neural Information Processing Systems},
 editor = {I. Guyon and U. Von Luxburg and S. Bengio and H. Wallach and R. Fergus and S. Vishwanathan and R. Garnett},
 pages = {},
 publisher = {Curran Associates, Inc.},
 title = {LightGBM: A Highly Efficient Gradient Boosting Decision Tree},
 url = {https://proceedings.neurips.cc/paper_files/paper/2017/file/6449f44a102fde848669bdd9eb6b76fa-Paper.pdf},
 volume = {30},
 year = {2017}
}

@article{gluonts_jmlr,
  author  = {Alexander Alexandrov and Konstantinos Benidis and Michael Bohlke-Schneider
    and Valentin Flunkert and Jan Gasthaus and Tim Januschowski and Danielle C. Maddix
    and Syama Rangapuram and David Salinas and Jasper Schulz and Lorenzo Stella and
    Ali Caner Türkmen and Yuyang Wang},
  title   = {{GluonTS: Probabilistic and Neural Time Series Modeling in Python}},
  journal = {Journal of Machine Learning Research},
  year    = {2020},
  volume  = {21},
  number  = {116},
  pages   = {1-6},
  url     = {http://jmlr.org/papers/v21/19-820.html}
}

@article{gneiting2007probabilistic,
  author  = {Gneiting, Tilmann and Balabdaoui, Fadoua and Raftery, Adrian E.},
  title   = {Probabilistic Forecasts, Calibration and Sharpness},
  journal = {Journal of the Royal Statistical Society: Series B (Statistical Methodology)},
  year    = {2007},
  volume  = {69},
  number  = {2},
  pages   = {243--268},
  doi     = {10.1111/j.1467-9868.2007.00587.x}
}
\bibliographystyle{tmlr}

\appendix
\newpage
\section{Appendix}
\subsection{Implementation of Calendar Features} \label{append-calendar-feat}

\begin{algorithm}[ht]
    \caption{Detailed Calendar Features Implementation}
    


\begin{algorithmic}[1]
\Require
\Statex Time-indexed table $\mathcal{D}$ with index level \texttt{timestamp}.
\Ensure
\Statex $\mathcal{D}$ augmented with:
\Statex \quad $\bullet$ \texttt{year} column;
\Statex \quad $\bullet$ $\sin$ and $\cos$ embeddings for each of:
\Statex \quad\quad
$\bigl(\mathtt{second\_of\_minute}, 60\bigr)$,
$\bigl(\mathtt{minute\_of\_hour}, 60\bigr)$,
$\bigl(\mathtt{hour\_of\_day}, 24\bigr)$,
$\bigl(\mathtt{day\_of\_week}, 7\bigr)$,
$\bigl(\mathtt{day\_of\_month}, 30.5\bigr)$,
$\bigl(\mathtt{day\_of\_year}, 365\bigr)$,
$\bigl(\mathtt{week\_of\_year}, 52\bigr)$,
$\bigl(\mathtt{month\_of\_year}, 12\bigr)$.
\Statex
\State \(\mathcal{D}\!\leftarrow\!\mathcal{D}.\text{copy()}\)
\State \(\mathbf{T}\!\leftarrow\!\mathcal{D}.\text{index.get\_level\_values}(\texttt{"timestamp"})\)
\medskip

\Statex
\Statex \textbf{Extract year component}
\State \(\mathcal{D}[\texttt{"year"}]\;\gets\;\mathbf{T}.\texttt{year}\)
\medskip


\Statex
\Statex
\textbf{Extract calendar-based seasonality}
\State \(\mathcal{S}\;\gets\;\{\)
  \Statex \quad \quad \((\texttt{"second\_of\_minute"},60),\)
  \Statex \quad \quad \((\texttt{"minute\_of\_hour"},  60),\)
  \Statex \quad \quad \((\texttt{"hour\_of\_day"},     24),\)
  \Statex \quad \quad \((\texttt{"day\_of\_week"},      7),\)
  \Statex \quad \quad \((\texttt{"day\_of\_month"}, 30.5),\)
  \Statex \quad \quad \((\texttt{"day\_of\_year"},    365),\)
  \Statex \quad \quad \((\texttt{"week\_of\_year"},    52),\)
  \Statex \quad \quad \((\texttt{"month\_of\_year"},   12)\)
  \Statex \(\}\)
\medskip
\Comment{List of seasonal features with their natural periods}

\ForAll{\((\textit{name},P)\) in \(\mathcal{S}\)}
  \State \(\mathbf{f}\!\gets\!\text{time\_feature}(\textit{name}).\text{index}(\mathbf{T})\)
  \Comment{integer cycle index}
  \State \(\tilde P\!\gets\!P-1\)
  \State \(\mathcal{D}[\textit{name}\|\_\,\texttt{sin}]\;\gets\;\sin\bigl(2\pi\,\mathbf{f}/\tilde P\bigr)\)
  \State \(\mathcal{D}[\textit{name}\|\_\,\texttt{cos}]\;\gets\;\cos\bigl(2\pi\,\mathbf{f}/\tilde P\bigr)\)
\EndFor

\State \Return \(\mathcal{D}\)
\end{algorithmic}

\end{algorithm}

\newpage

\subsection{Implementation of Automatic Seasonal Features}
\label{append-auto-seasonal-features}

\begin{algorithm}[ht]
    \caption{Detailed Extract top-\(k\) Seasonalities Algorithm}
    \label{alg:auto-seasonal-feature}

\begin{algorithmic}[1]
\Require
\Statex \quad $\bullet$ Time series $\mathbf{x}_t = \{x_1, x_2, \dots, x_N\}$
\Statex \quad $\bullet$ Integer $k$ (max number of periods)
\Statex \quad $\bullet$ Hann window length $L$
\Ensure 
Set $\mathcal{P}$ of up to $k$ dominant periods

\Statex
\Statex
\textbf{Preprocessing:}
\State Detrend $\mathbf{x}_t$ via linear regression \[\tilde{x}_t = x_t - (\alpha t + \beta) \quad \text{, where $\alpha$ and $\beta$ are found using least squares}\]
\State Apply Hann window:
   \[
   w_t' = 0.5 \left(1 - \cos\left(\frac{2\pi t'}{L}\right)\right) \quad \text{for} \quad t \in \{0,\dots,L\}
   \]
   \[
   \breve{x} = conv(\tilde{x}, w)
   \]
\State Symmetrically zero-pad to length $2N$:
   \[
   \mathbf{y} = [0, \dots, 0, \breve{x}_1, \dots, \breve{x}_N, 0, \dots, 0]
   \]

\Statex \textbf{Spectral Analysis:}
\State Compute fast fourier transform:
   \[
   Y_k = \sum_{t=1}^{2N} y_t e^{-i2\pi (k-1)t/(2N)} \quad \text{for } k = 1,\dots,N
   \]
   \[
   \text{Magnitudes: } A_k = |Y_k|
   \]
   \[
   \text{Frequencies: } f_k = \frac{k-1}{2N} \quad \text{(normalized to Nyquist)}
   \]
\State Remove DC component:
   \[
   A_1 \leftarrow 0
   \]

\Statex \textbf{Peak Selection:}
\State Identify local maxima (peaks larger than immediate neighbors, taking midpoint of multi-point peaks in practice):
\[
\mathcal{L} = \left\{ i \in \{2, \dots, N-1\} \,\Big|\, A_i > A_{i-1} \text{ and } A_i > A_{i+1} \right\}
\]

\Statex \textbf{Period Conversion:}
\State Convert frequencies to periods and round to integers:
   \[
   p_i = \left\lfloor \frac{1}{f_i} \right\rceil \text{, for } i \in L
   \]
\State Remove duplicates and $0.$ periods, yielding a new set of indexes $\mathcal{I}$
\State Build top $k$ index set $\mathcal{T} = \{i \in \mathcal{I} | A_i \in \text{topk}_{k}(\{A_i|i \in \mathcal{I}\})\}$\\
\Return $\{p_i | i \in \mathcal{T}\}$
\end{algorithmic}

\end{algorithm}

\newpage

\subsection{\tabpfnts Configuration} \label{sec:append-tabpfn-ts-config}

We describe the configuration used for all \tabpfnts experiments.
Unless otherwise noted, all settings are fixed across benchmarks.

\paragraph{Data Preprocessing.}
For time series with missing observations, we remove the affected time points from the conditioning set.
For efficiency, we restrict the conditioning window to the most recent $4096$ observations (after removing missing points), which we found to provide a good balance between accuracy and computational cost (see Appendix~\ref{append-context-length}).
No additional normalization is applied to the target values, as \tabpfn performs internal z-normalization during inference.

\paragraph{\tabpfn Model Configuration.}
\tabpfn offers several publicly available regression checkpoints.
For all experiments, we use the \texttt{tabpfn-v2-regression-2noar4o2.ckpt}\footnote{https://huggingface.co/Prior-Labs/TabPFN-v2-reg/blob/9ff6a7be140b2663cdd21387a5decc3a8018ad6b/tabpfn-v2-regression-2noar4o2.ckpt}, which showed stable behavior across both point and probabilistic forecasting in preliminary evaluations.
We do not perform hyperparameter tuning or checkpoint selection beyond this fixed choice, as our focus is on evaluating the utility of the forecasting formulation rather than optimizing the backbone.

\paragraph{Temporal Featurization.}
We apply the full featurization pipeline described in Section~\ref{sec:from-ts-to-tabular}.
For Automatic Seasonal Features, we retain the top $k=5$ periodicities identified by the spectral analysis described in Section~\ref{sec:from-ts-to-tabular}.  
This value provides a good balance between expressiveness and computational efficiency: increasing $k$ adds additional feature dimensions, but we found in preliminary checks that the model's forecasting accuracy is relatively stable across reasonable values of $k$.  
A more extensive study of this parameter is left for future work.

\newpage

\subsection{GIFT-Eval Benchmark Datasets and Corresponding Statistics} \label{sec:append-dataset}

Each benchmarking task in GIFT-Eval corresponds to a unique combination of dataset, prediction horizon (short-, medium-, or long-term), and sampling frequency (where applicable).
For a given dataset, a benchmarking task is defined only if sufficient historical data is available to support the specified window size and forecast length, as shown in the short-, medium-, and long-term columns of Table~\ref{table-dataset-statistics}.
In total, GIFT-Eval comprises 97 such tasks that span diverse domains, temporal resolutions, and forecasting lengths.

These 97 tasks are used in the main experimental evaluation.
For the ablation studies, we exclude datasets marked with an asterisk (*) due to their relatively large size and higher resource requirements.

\vspace{0.5em}

\begin{table}[h]
    \caption{Statistics of datasets from the GIFT-Eval benchmark (reproduced from \citet{gift-eval} under a CC BY 4.0 license). Datasets marked with an asterisk (*) are excluded from ablation studies due to the large size.}
    \label{table-dataset-statistics}
    \centering
    \resizebox{\textwidth}{!}{%
\begin{tabular}{cccccccccccccccc}
\toprule
\multirow{2}{*}{\textbf{Dataset}} &
  \multirow{2}{*}{\textbf{Domain}} &
  \multirow{2}{*}{\textbf{Frequency}} &
  \multirow{2}{*}{\textbf{\# Series}} &
  \multicolumn{3}{c}{\textbf{Series Length}} &
  \multirow{2}{*}{\textbf{\# Obs}} &
  \multirow{2}{*}{\textbf{\# Target Variates}} &
  \multicolumn{2}{c}{\textbf{Short-term}} &
  \multicolumn{2}{c}{\textbf{Med-term}} &
  \multicolumn{2}{c}{\textbf{Long-term}} \\ 
  \cmidrule(r){5-7}
  \cmidrule(r){10-11}
  \cmidrule(r){12-13}
  \cmidrule(r){14-15}
  &  &  &  &
  \textbf{Avg} &
  \textbf{Min} &
  \textbf{Max} &
    &
    &
  \textbf{Pred Length} &
  \textbf{Windows} &
  \textbf{Pred Length} &
  \textbf{Windows} &
  \textbf{Pred Length} &
  \textbf{Windows} \\
\midrule
Jena Weather &
  Nature &
  10T &
  1 &
  52,704 &
  52,704 &
  52,704 &
  52,704 &
  21 &
  48 &
  20 &
  480 &
  11 &
  720 &
  8 \\
Jena Weather &
 Nature &
  H &
  1 &
  8,784 &
  8,784 &
  8,784 &
  8,784 &
  21 &
  48 &
  19 &
  480 &
  2 &
  720 &
  2 \\
Jena Weather &
 Nature &
  D &
  1 &
  366 &
  366 &
  366 &
  366 &
  21 &
  30 &
  2 &
   &
   &
   &
   \\
BizITObs - Application &
  Web/CloudOps &
  10S &
  1 &
  8,834 &
  8,834 &
  8,834 &
  8,834 &
  2 &
  60 &
  15 &
  600 &
  2 &
  900 &
  1 \\
BizITObs - Service &
  Web/CloudOps &
  10S &
  21 &
  8,835 &
  8,835 &
  8,835 &
  185,535 &
  2 &
  60 &
  15 &
  600 &
  2 &
  900 &
  1 \\
BizITObs - L2C &
  Web/CloudOps &
  5T &
  1 &
  31,968 &
  31,968 &
  31,968 &
  31,968 &
  7 &
  48 &
  20 &
  480 &
  7 &
  720 &
  5 \\
BizITObs - L2C &
 Web/CloudOps &
  H &
  1 &
  2,664 &
  2,664 &
  2,664 &
  2,664 &
  7 &
  48 &
  6 &
  480 &
  1 &
  720 &
  1 \\
Bitbrains - Fast Storage &
  Web/CloudOps &
  5T* &
  1,250 &
  8,640 &
  8,640 &
  8,640 &
  10,800,000 &
  2 &
  48 &
  18 &
  480 &
  2 &
  720 &
  2 \\
Bitbrains - Fast Storage &
  Web/CloudOps &
  H &
  1,250 &
  721 &
  721 &
  721 &
  901,250 &
  2 &
  48 &
  2 &
   &
   &
   &
   \\
Bitbrains - rnd* &
  Web/CloudOps &
  5T &
  500 &
  8,640 &
  8,640 &
  8,640 &
  4,320,000 &
  2 &
  48 &
  18 &
  480 &
  2 &
  720 &
  2 \\
Bitbrains - rnd &
  Web/CloudOps &
  H &
  500 &
  720 &
  720 &
  720 &
  360,000 &
  2 &
  48 &
  2 &
   &
   &
   &
   \\
Restaurant &
  Sales &
  D &
  807 &
  358 &
  67 &
  478 &
  289,303 &
  1 &
  30 &
  1 &
   &
   &
   &
   \\
ETT1 &
  Energy &
  15T &
  1 &
  69,680 &
  69,680 &
  69,680 &
  69,680 &
  7 &
  48 &
  20 &
  480 &
  15 &
  720 &
  10 \\
ETT1 &
  Energy &
  H &
  1 &
  17,420 &
  17,420 &
  17,420 &
  17,420 &
  7 &
  48 &
  20 &
  480 &
  4 &
  720 &
  3 \\
ETT1 &
  Energy &
  D &
  1 &
  725 &
  725 &
  725 &
  725 &
  7 &
  30 &
  3 &
   &
   &
   &
   \\
ETT1 &
  Energy &
  W-THU &
  1 &
  103 &
  103 &
  103 &
  103 &
  7 &
  8 &
  2 &
   &
   &
   &
   \\
ETT2 &
  Energy &
  15T &
  1 &
  69,680 &
  69,680 &
  69,680 &
  69,680 &
  7 &
  48 &
  20 &
  480 &
  15 &
  720 &
  10 \\
ETT2 &
  Energy &
  H &
  1 &
  17,420 &
  17,420 &
  17,420 &
  17,420 &
  7 &
  48 &
  20 &
  480 &
  4 &
  720 &
  3 \\
ETT2 &
  Energy &
  D &
  1 &
  725 &
  725 &
  725 &
  725 &
  7 &
  30 &
  3 &
   &
   &
   &
   \\
ETT2 &
  Energy &
  W-THU &
  1 &
  103 &
  103 &
  103 &
  103 &
  7 &
  8 &
  2 &
   &
   &
   &
   \\
Loop Seattle* &
  Transport &
  5T &
  323 &
  105,120 &
  105,120 &
  105,120 &
  33,953,760 &
  1 &
  48 &
  20 &
  480 &
  20 &
  720 &
  15 \\
Loop Seattle* &
  Transport &
  H &
  323 &
  8,760 &
  8,760 &
  8,760 &
  2,829,480 &
  1 &
  48 &
  19 &
  480 &
  2 &
  720 &
  2 \\
Loop Seattle &
  Transport &
  D &
  323 &
  365 &
  365 &
  365 &
  117,895 &
  1 &
  30 &
  2 &
   &
   &
   &
   \\
SZ-Taxi &
  Transport &
  15T &
  156 &
  2,976 &
  2,976 &
  2,976 &
  464,256 &
  1 &
  48 &
  7 &
  480 &
  1 &
  720 &
  1 \\
SZ-Taxi &
  Transport &
  H &
  156 &
  744 &
  744 &
  744 &
  116,064 &
  1 &
  48 &
  2 &
   &
   &
   &
   \\
M\_DENSE &
  Transport &
  H &
  30 &
  17,520 &
  17,520 &
  17,520 &
  525,600 &
  1 &
  48 &
  20 &
  480 &
  4 &
  720 &
  3 \\
M\_DENSE &
  Transport &
  D &
  30 &
  730 &
  730 &
  730 &
  21,900 &
  1 &
  30 &
  3 &
   &
   &
   &
   \\
Solar &
  Energy &
  10T &
  137 &
  52,560 &
  52,560 &
  52,560 &
  7,200,720 &
  1 &
  48 &
  20 &
  480 &
  11 &
  720 &
  8 \\
Solar &
  Energy &
  H &
  137 &
  8,760 &
  8,760 &
  8,760 &
  1,200,120 &
  1 &
  48 &
  19 &
  480 &
  2 &
  720 &
  2 \\
Solar &
  Energy &
  D &
  137 &
  365 &
  365 &
  365 &
  50,005 &
  1 &
  30 &
  2 &
   &
   &
   &
   \\
Solar &
  Energy &
  W-FRI &
  137 &
  52 &
  52 &
  52 &
  7,124 &
  1 &
  8 &
  1 &
   &
   &
   &
   \\
Hierarchical Sales &
  Sales &
  D &
  118 &
  1,825 &
  1,825 &
  1,825 &
  215,350 &
  1 &
  30 &
  7 &
   &
   &
   &
   \\
Hierarchical Sales &
  Sales &
  W-WED &
  118 &
  260 &
  260 &
  260 &
  30,680 &
  1 &
  8 &
  4 &
   &
   &
   &
   \\
M4 Yearly &
  Econ/Fin &
  A-DEC &
  22,974 &
  37 &
  19 &
  284 &
  845,109 &
  1 &
  6 &
  1 &
   &
   &
   &
   \\
M4 Quarterly &
  Econ/Fin &
  Q-DEC &
  24,000 &
  100 &
  24 &
  874 &
  2,406,108 &
  1 &
  8 &
  1 &
   &
   &
   &
   \\
M4 Monthly &
  Econ/Fin &
  M &
  48,000 &
  234 &
  60 &
  2,812 &
  11,246,411 &
  1 &
  18 &
  1 &
   &
   &
   &
   \\
M4 Weekly &
  Econ/Fin &
  W-SUN &
  359 &
  1,035 &
  93 &
  2,610 &
  371,579 &
  1 &
  13 &
  1 &
   &
   &
   &
   \\
M4 Daily &
  Econ/Fin &
  D &
  4,227 &
  2,371 &
  107 &
  9,933 &
  10,023,836 &
  1 &
  14 &
  1 &
   &
   &
   &
   \\
M4 Hourly &
  Econ/Fin &
  H &
  414 &
  902 &
  748 &
  1,008 &
  373,372 &
  1 &
  48 &
  2 &
   &
   &
   &
   \\
Hospital &
  Healthcare &
  M &
  767 &
  84 &
  84 &
  84 &
  64,428 &
  1 &
  12 &
  1 &
   &
   &
   &
   \\
COVID Deaths &
  Healthcare &
  D &
  266 &
  212 &
  212 &
  212 &
  56,392 &
  1 &
  30 &
  1 &
   &
   &
   &
   \\
US Births &
  Healthcare &
  D &
  1 &
  7,305 &
  7,305 &
  7,305 &
  7,305 &
  1 &
  30 &
  20 &
   &
   &
   &
   \\
US Births &
  Healthcare &
  W-TUE &
  1 &
  1,043 &
  1,043 &
  1,043 &
  1,043 &
  1 &
  8 &
  14 &
   &
   &
   &
   \\
US Births &
  Healthcare &
  M &
  1 &
  240 &
  240 &
  240 &
  240 &
  1 &
  12 &
  2 &
   &
   &
   &
   \\
Saugeen &
  Nature &
  D &
  1 &
  23,741 &
  23,741 &
  23,741 &
  23,741 &
  1 &
  30 &
  20 &
   &
   &
   &
   \\
Saugeen &
  Nature &
  W-THU &
  1 &
  3,391 &
  3,391 &
  3,391 &
  3,391 &
  1 &
  8 &
  20 &
   &
   &
   &
   \\
Saugeen &
  Nature &
  M &
  1 &
  780 &
  780 &
  780 &
   &
  1 &
  12 &
  7 &
   &
   &
   &
   \\
Temperature Rain* &
  Nature &
  D &
  32,072 &
  725 &
  725 &
  725 &
  780 &
  1 &
  30 &
  3 &
   &
   &
   &
   \\
KDD Cup 2018 &
  Nature &
  H &
  270 &
  10,898 &
  9,504 &
  10,920 &
  2,942,364 &
  1 &
  48 &
  20 &
  480 &
  2 &
  720 &
  2 \\
KDD Cup 2018 &
  Nature &
  D &
  270 &
  455 &
  396 &
  455 &
  122,791 &
  1 &
  30 &
  2 &
   &
   &
   &
   \\
Car Parts &
  Sales &
  M &
  2,674 &
  51 &
  51 &
  51 &
  136,374 &
  1 &
  12 &
  1 &
   &
   &
   &
   \\
Electricity* &
  Energy &
  15T &
  370 &
  140,256 &
  140,256 &
  140,256 &
  51,894,720 &
  1 &
  48 &
  20 &
  480 &
  20 &
  720 &
  20 \\
Electricity &
  Energy &
  H &
  370 &
  35,064 &
  35,064 &
  35,064 &
  12,973,680 &
  1 &
  48 &
  20 &
  480 &
  8 &
  720 &
  5 \\
Electricity &
  Energy &
  D &
  370 &
  1,461 &
  1,461 &
  1,461 &
  540,570 &
  1 &
  30 &
  5 &
   &
   &
   &
   \\
Electricity &
  Energy &
  W-FRI &
  370 &
  208 &
  208 &
  208 &
  76,960 &
  1 &
  8 &
  3 &
   &
   &
   &
   \\
\bottomrule
\end{tabular}%
}

\end{table}

\pagebreak

\subsection{Additional Results on GIFT-Eval} \label{append-additional-res}


Figures~\ref{append-vis-short}–\ref{append-vis-long} present example predictions from \tabpfnts on randomly selected samples from short-, medium-, and long-term forecasting tasks, respectively.

\vspace*{1em} 

\def\width{0.48}
\def\gap{4pt}
\begin{figure}[H]
    \centering
    \begin{minipage}{\width\textwidth}
        \centering
        \includegraphics[width=\linewidth]{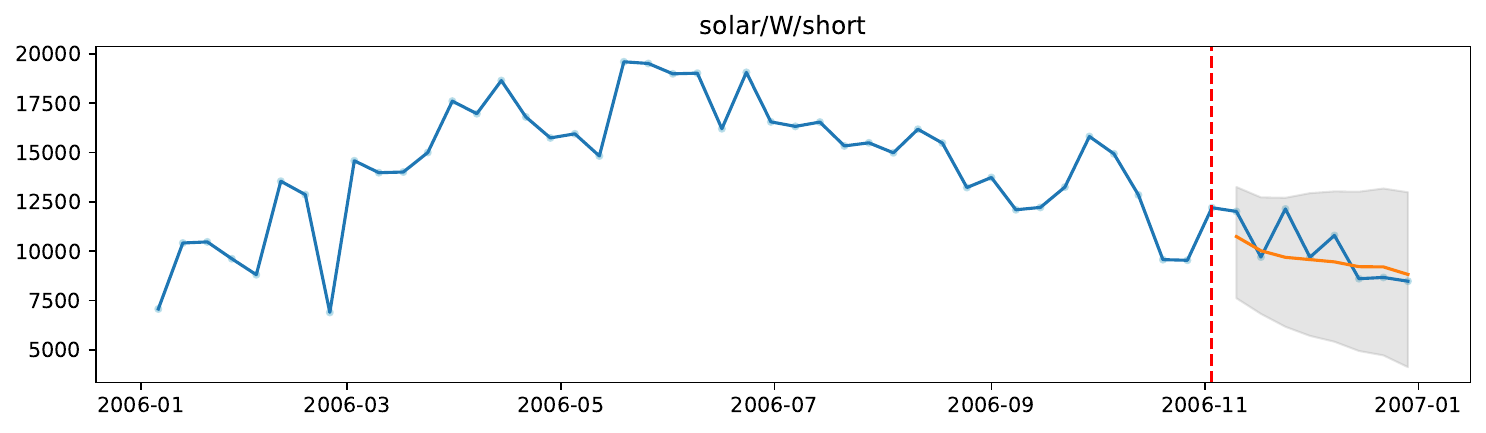}
    \end{minipage}
    \hfill
    \begin{minipage}{\width\textwidth}
        \centering
        \includegraphics[width=\linewidth]{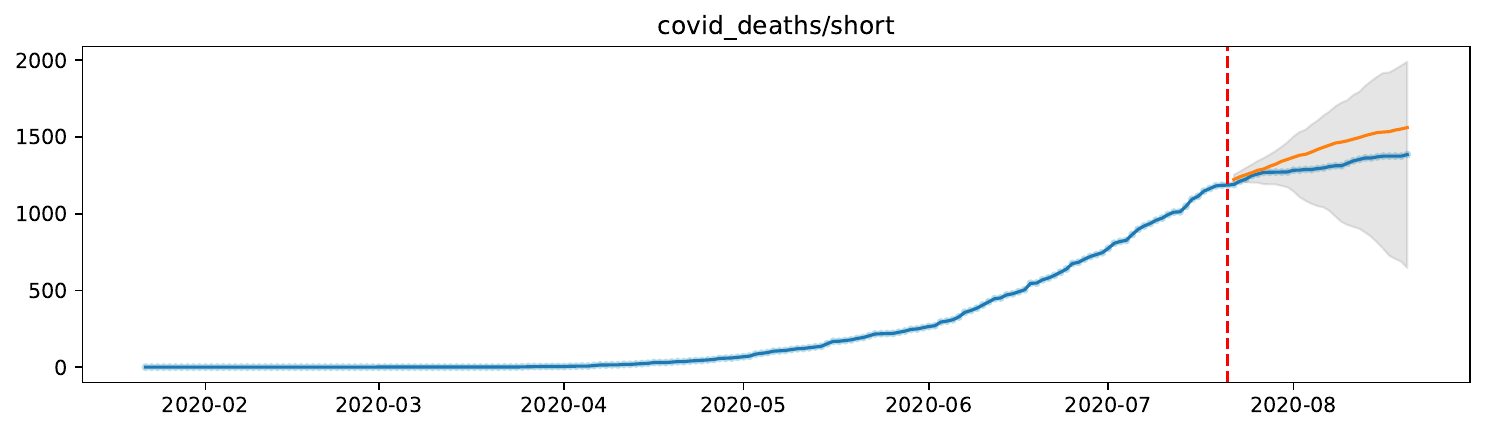}
    \end{minipage}

    \vspace{\gap} 

    \begin{minipage}{\width\textwidth}
        \centering
        \includegraphics[width=\linewidth]{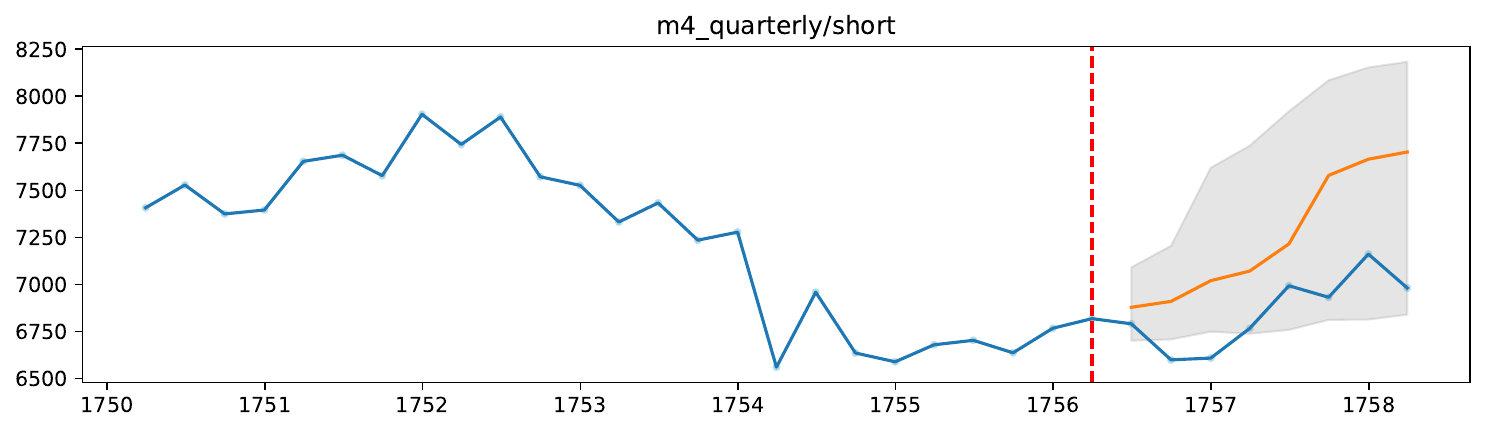}
    \end{minipage}
    \hfill
    \begin{minipage}{\width\textwidth}
        \centering
        \includegraphics[width=\linewidth]{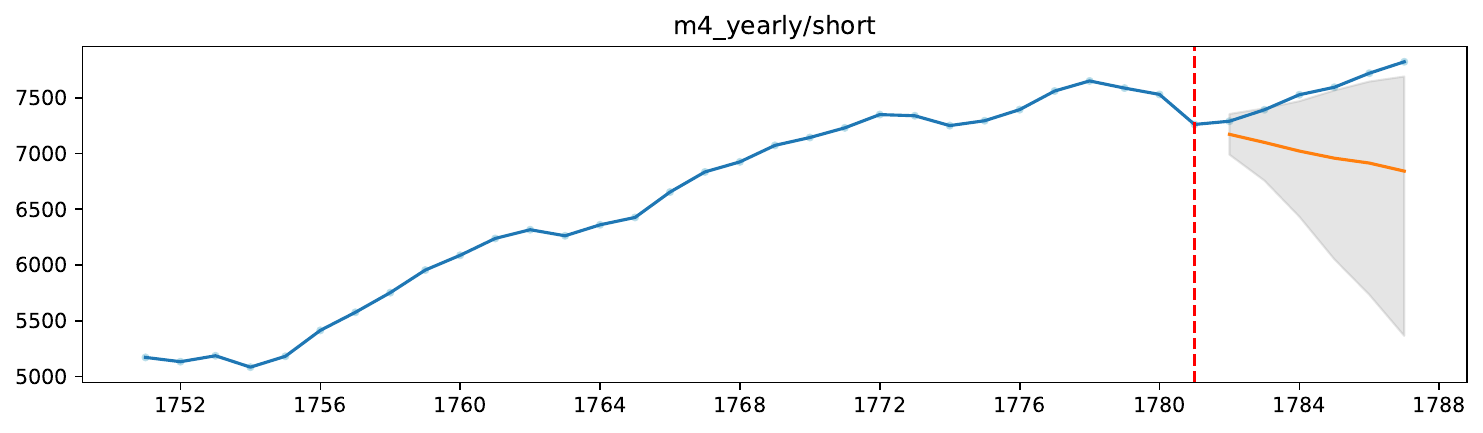}
    \end{minipage}

    \vspace{\gap} 

    \begin{minipage}{\width\textwidth}
        \centering
        \includegraphics[width=\linewidth]{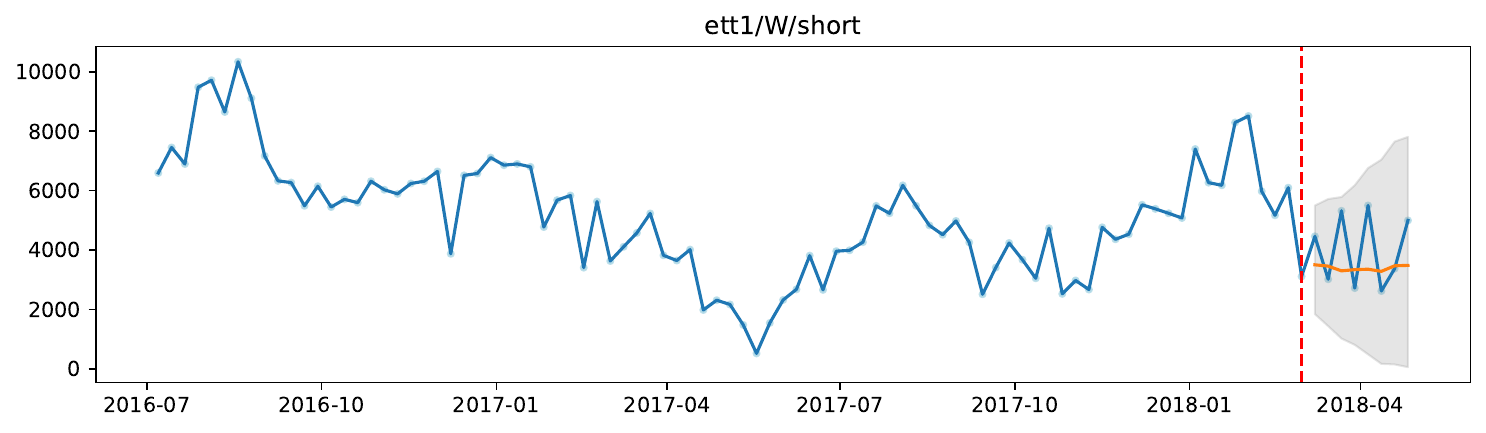}
    \end{minipage}
    \hfill
    \begin{minipage}{\width\textwidth}
        \centering
        \includegraphics[width=\linewidth]{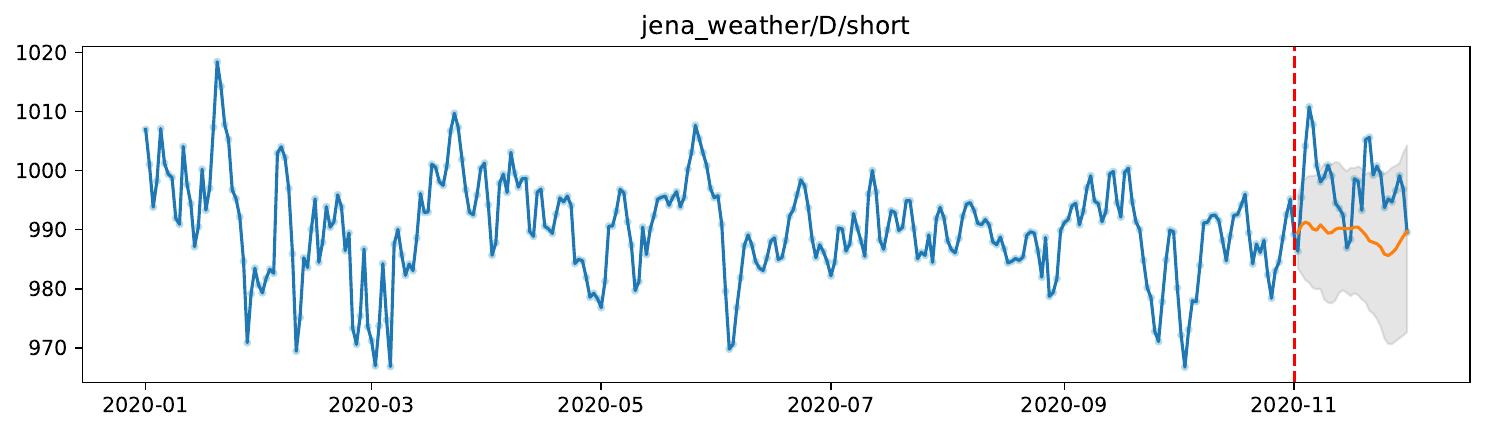}
    \end{minipage}

    \vspace{\gap} 

    \begin{minipage}{\width\textwidth}
        \centering
        \includegraphics[width=\linewidth]{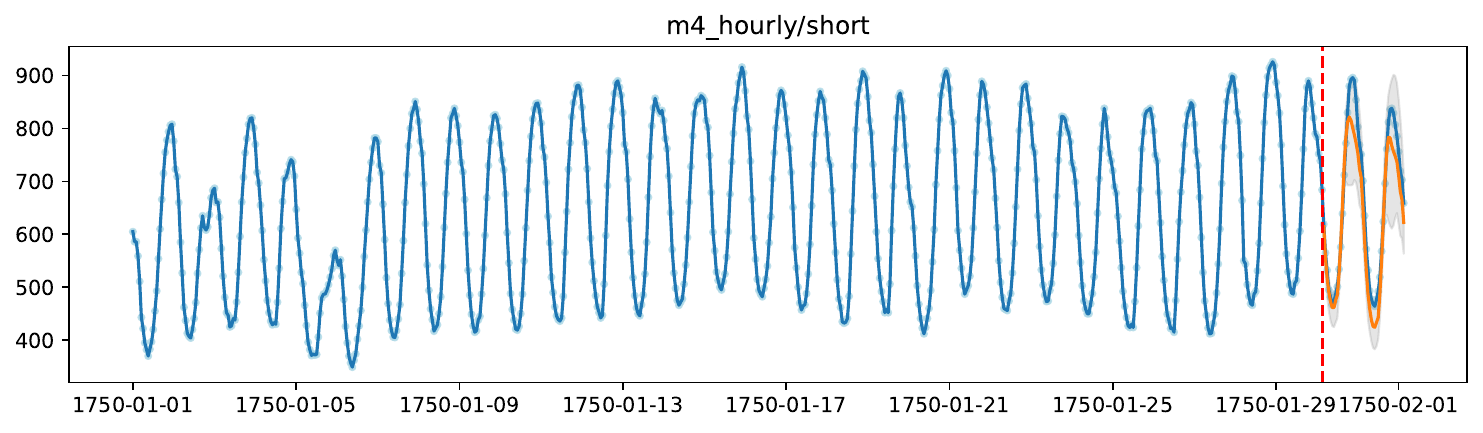}
    \end{minipage}
    \hfill
    \begin{minipage}{\width\textwidth}
        \centering
        \includegraphics[width=\linewidth]{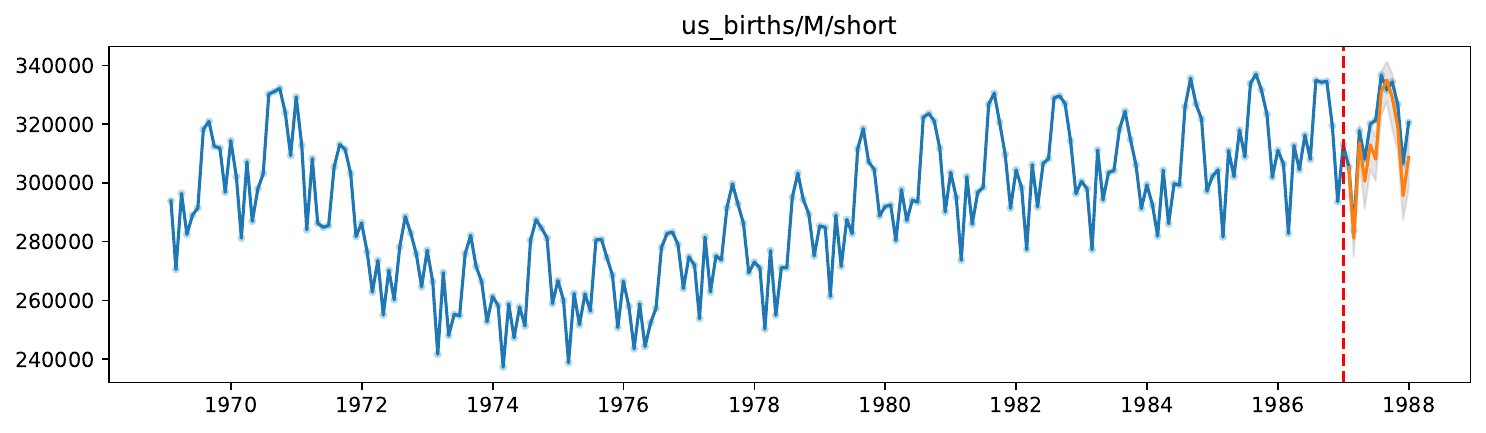}
    \end{minipage}

    \vspace{\gap} 

    \begin{minipage}{\width\textwidth}
        \centering
        \includegraphics[width=\linewidth]{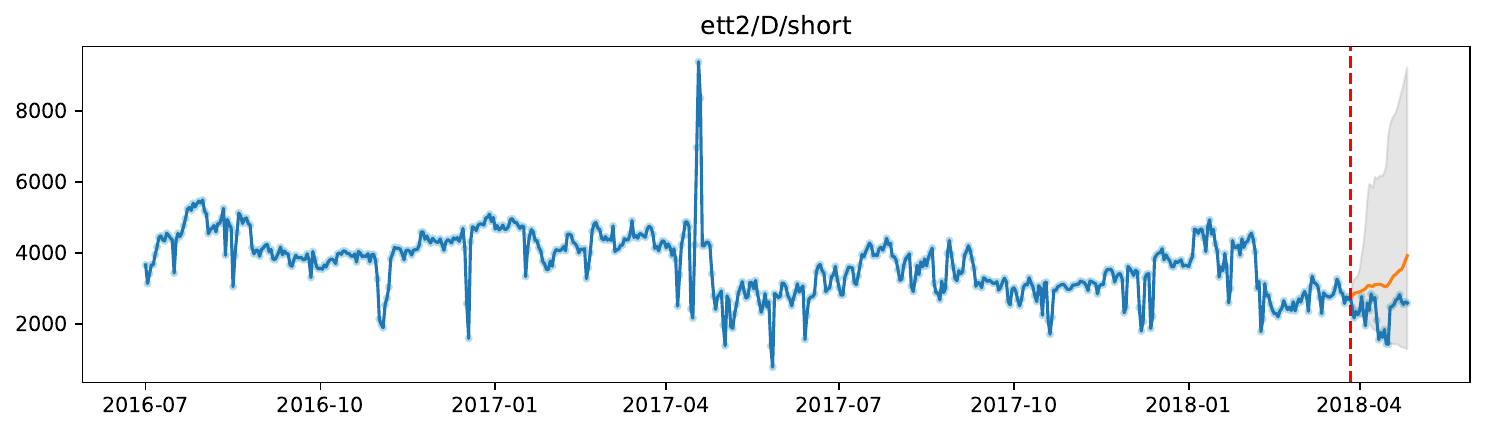}
    \end{minipage}
    \hfill
    \begin{minipage}{\width\textwidth}
        \centering
        \includegraphics[width=\linewidth]{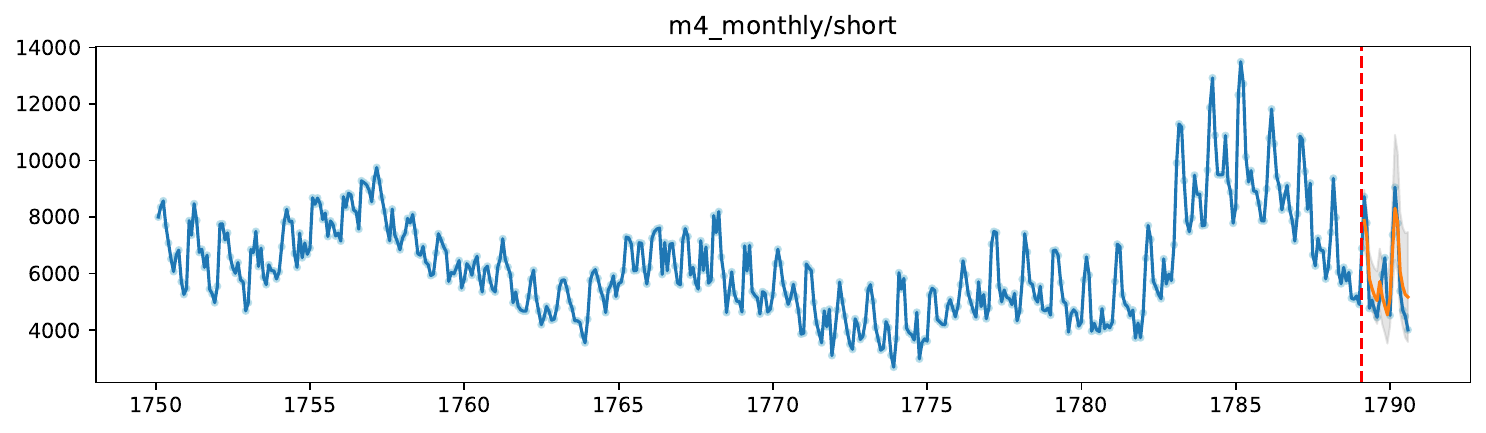}
    \end{minipage}

    \vspace{\gap} 

    \begin{minipage}{\width\textwidth}
        \centering
        \includegraphics[width=\linewidth]{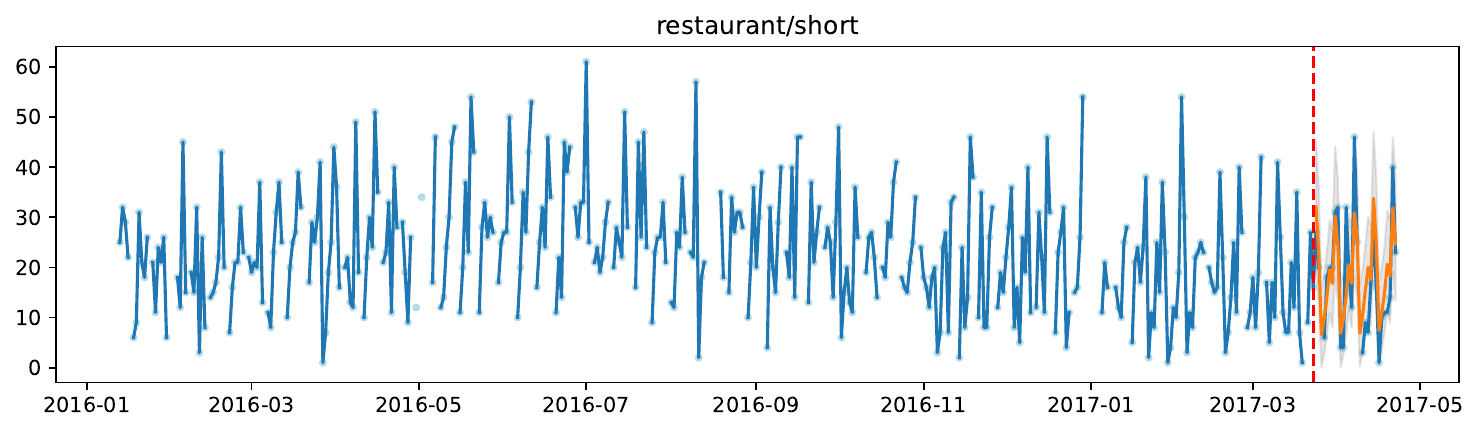}
    \end{minipage}
    \hfill
    \begin{minipage}{\width\textwidth}
        \centering
        \includegraphics[width=\linewidth]{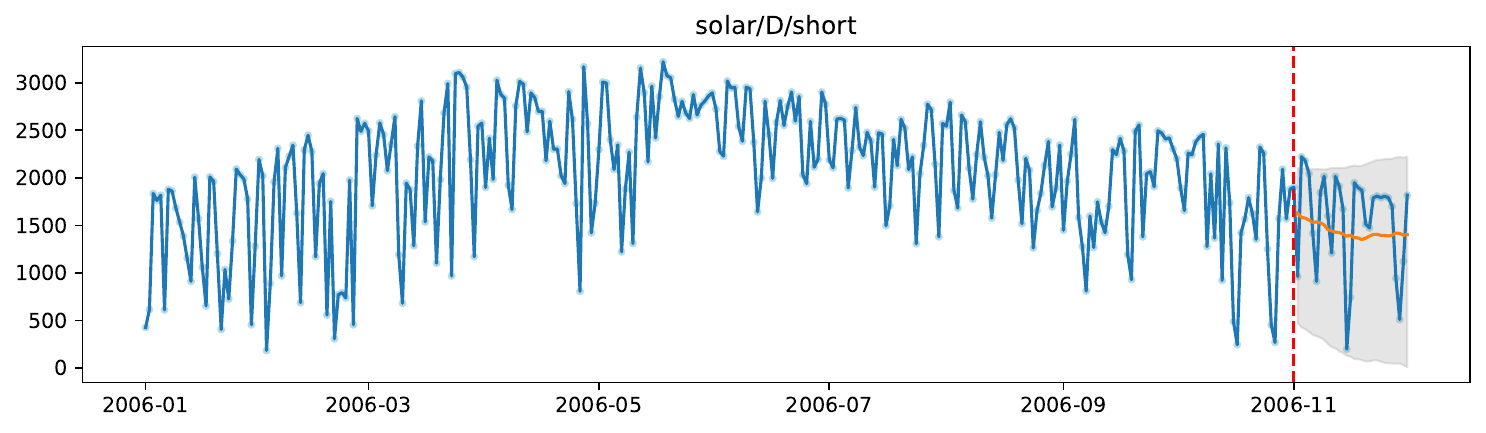}
    \end{minipage}

    \vspace{\gap} 

    \begin{minipage}{\width\textwidth}
        \centering
        \includegraphics[width=\linewidth]{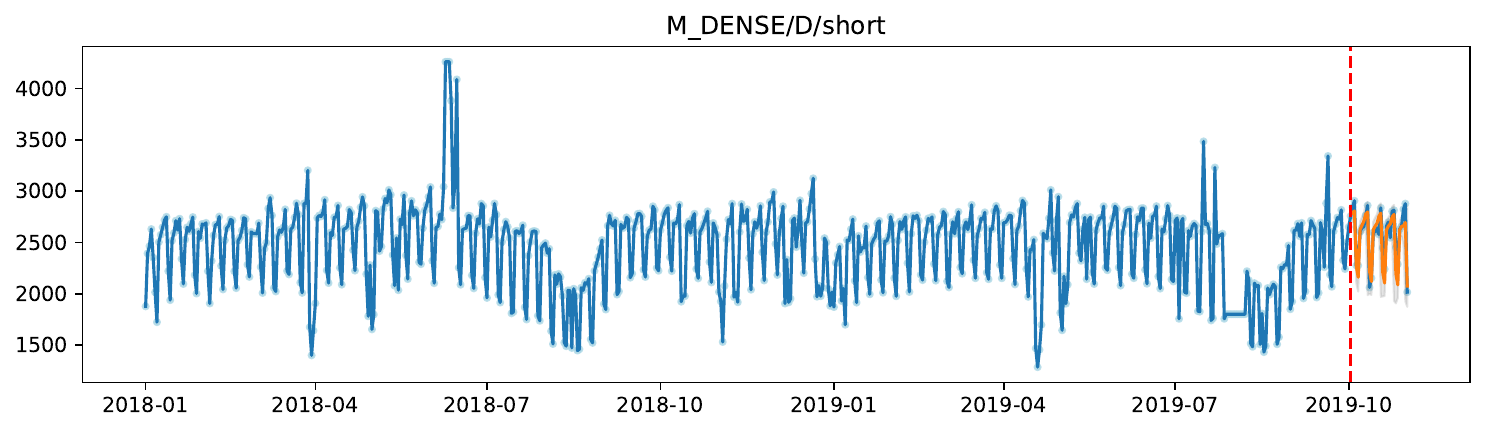}
    \end{minipage}
    \hfill
    \begin{minipage}{\width\textwidth}
        \centering
        \includegraphics[width=\linewidth]{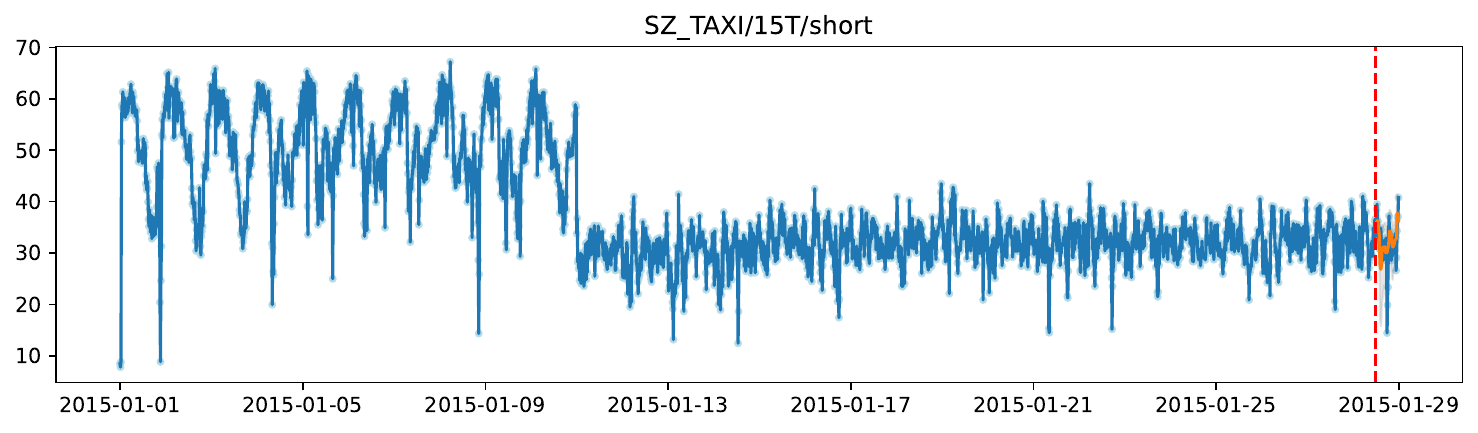}
    \end{minipage}
    
    \caption{Visualization of the \tabpfnts' predictions on some of the \textbf{short-term} benchmarking tasks.}
    \label{append-vis-short}
\end{figure}
\def\width{0.48}
\def\gap{4pt}
\begin{figure}[H]
    \centering
    \begin{minipage}{\width\textwidth}
        \centering
        \includegraphics[width=\linewidth]{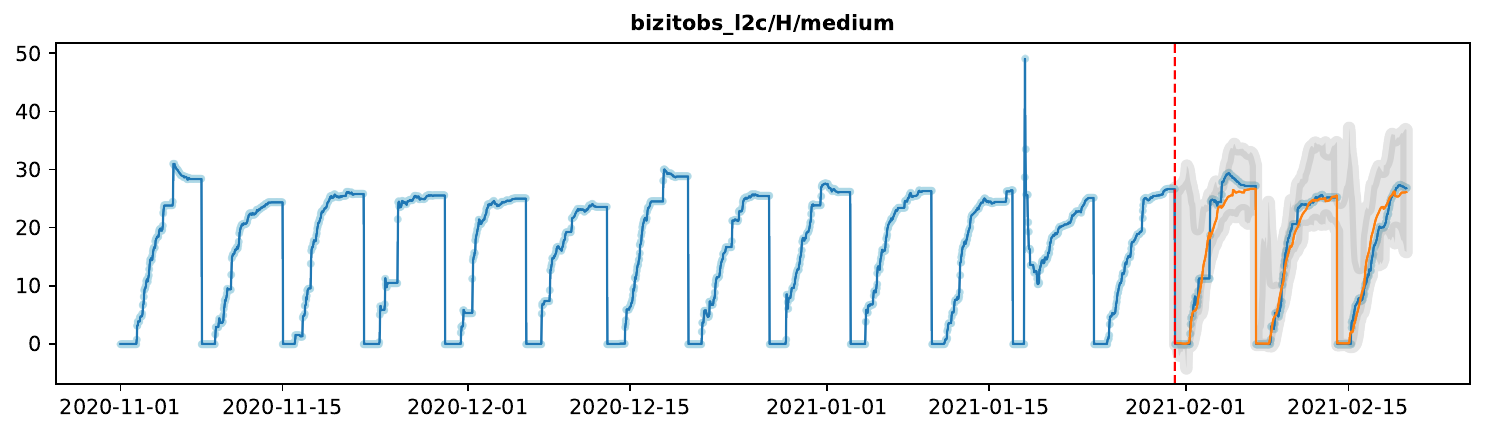}
    \end{minipage}
    \hfill
    \begin{minipage}{\width\textwidth}
        \centering
        \includegraphics[width=\linewidth]{figures/medium/bizitobs_l2c-H-medium.pdf}
    \end{minipage}

    \vspace{\gap} 

    \begin{minipage}{\width\textwidth}
        \centering
        \includegraphics[width=\linewidth]{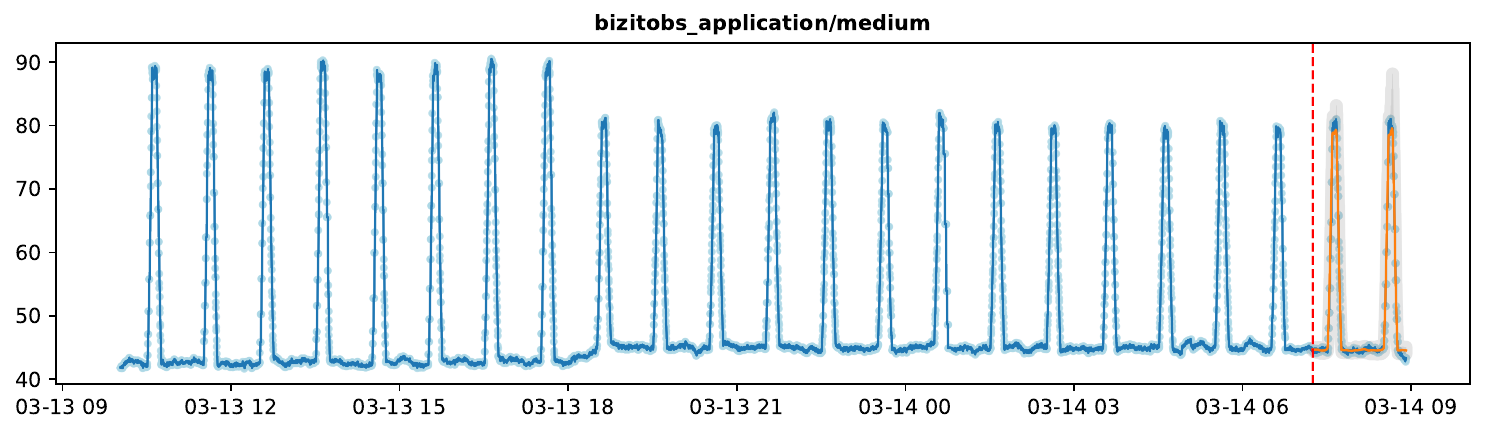}
    \end{minipage}
    \hfill
    \begin{minipage}{\width\textwidth}
        \centering
        \includegraphics[width=\linewidth]{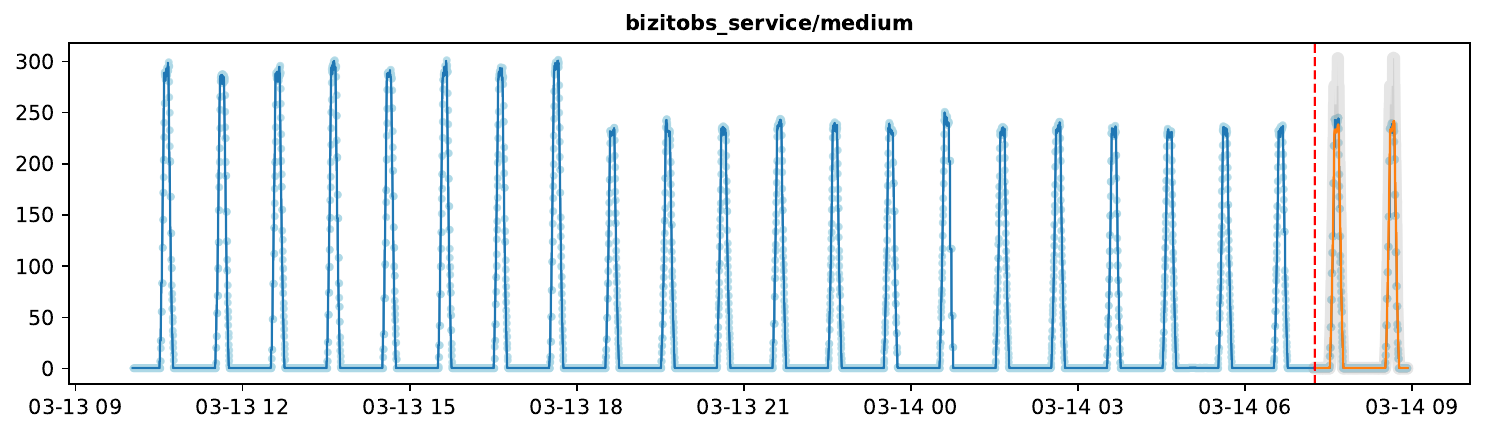}
    \end{minipage}

    \vspace{\gap} 


    \vspace{\gap} 

    \begin{minipage}{\width\textwidth}
        \centering
        \includegraphics[width=\linewidth]{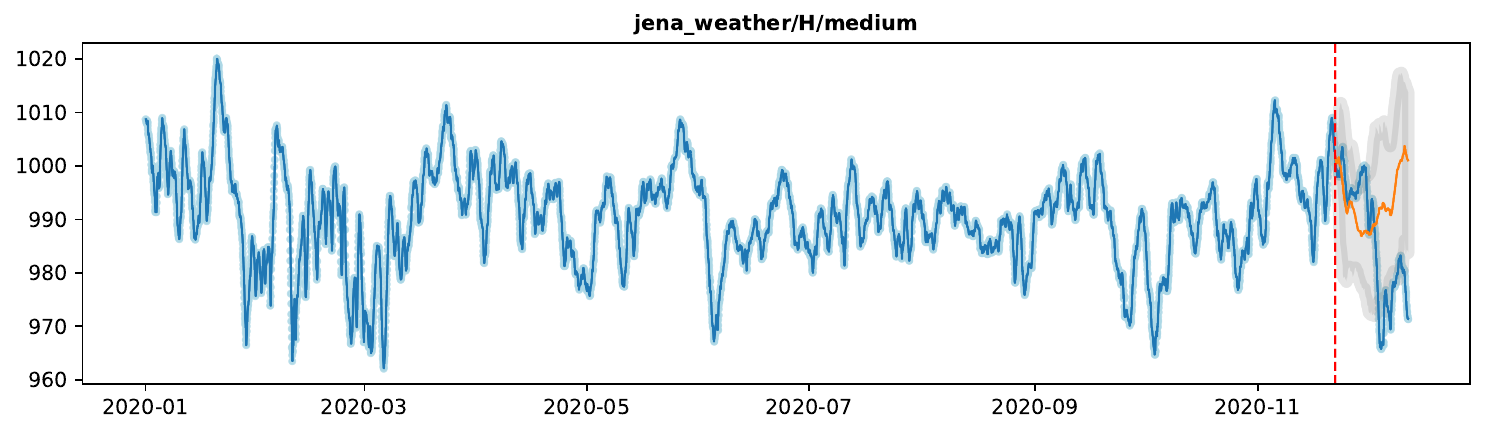}
    \end{minipage}
    \hfill
    \begin{minipage}{\width\textwidth}
        \centering
        \includegraphics[width=\linewidth]{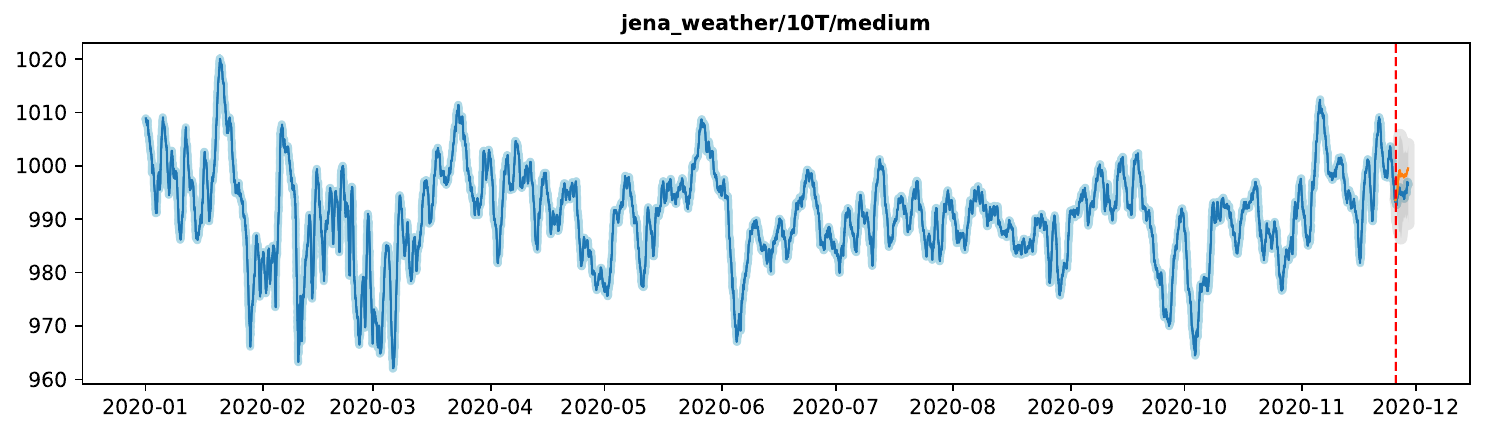}
    \end{minipage}

    \vspace{\gap} 

    \begin{minipage}{\width\textwidth}
        \centering
        \includegraphics[width=\linewidth]{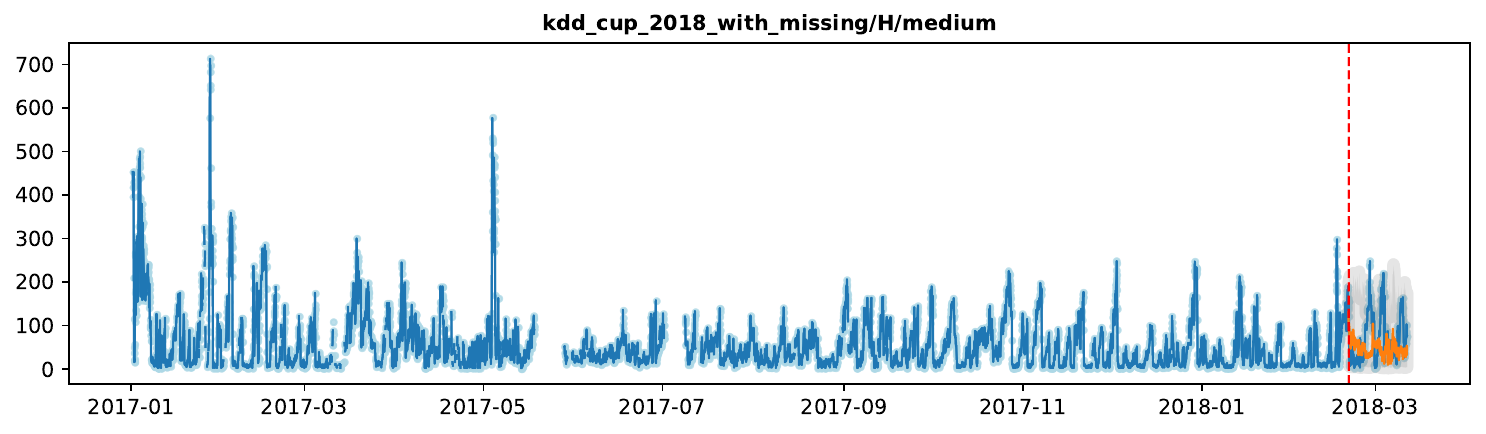}
    \end{minipage}
    \hfill
    \begin{minipage}{\width\textwidth}
        \centering
        \includegraphics[width=\linewidth]{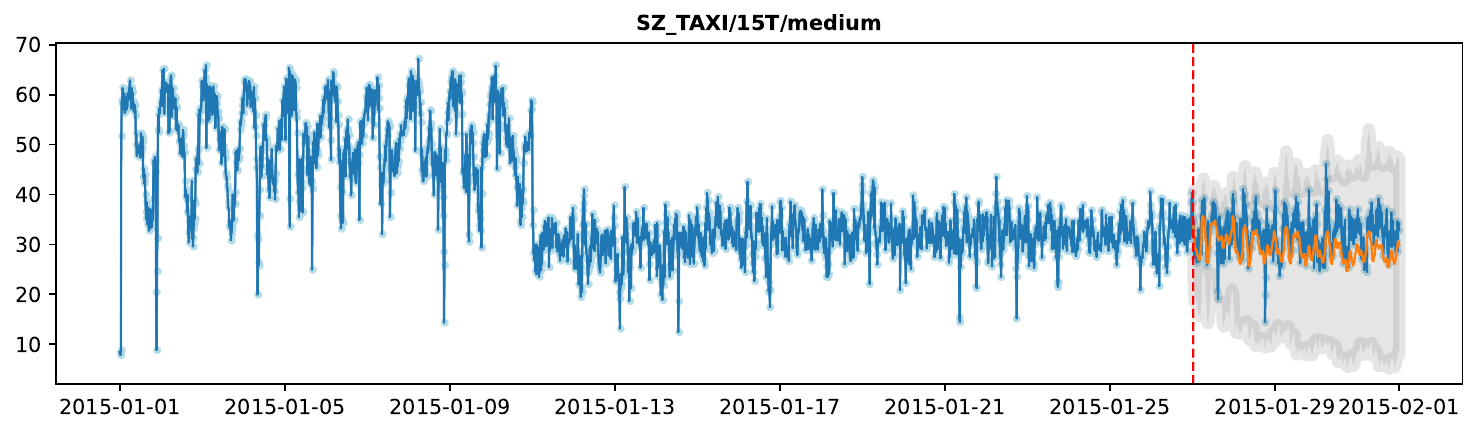}
    \end{minipage}

    \caption{Visualization of the \tabpfnts' predictions on some of the \textbf{medium-term} benchmarking tasks.}
    \label{append-vis-medium}
\end{figure}
\vspace{5em}
\def\width{0.48}
\def\gap{4pt}
\begin{figure}[H]
    \centering
    \begin{minipage}{\width\textwidth}
        \centering
        \includegraphics[width=\linewidth]{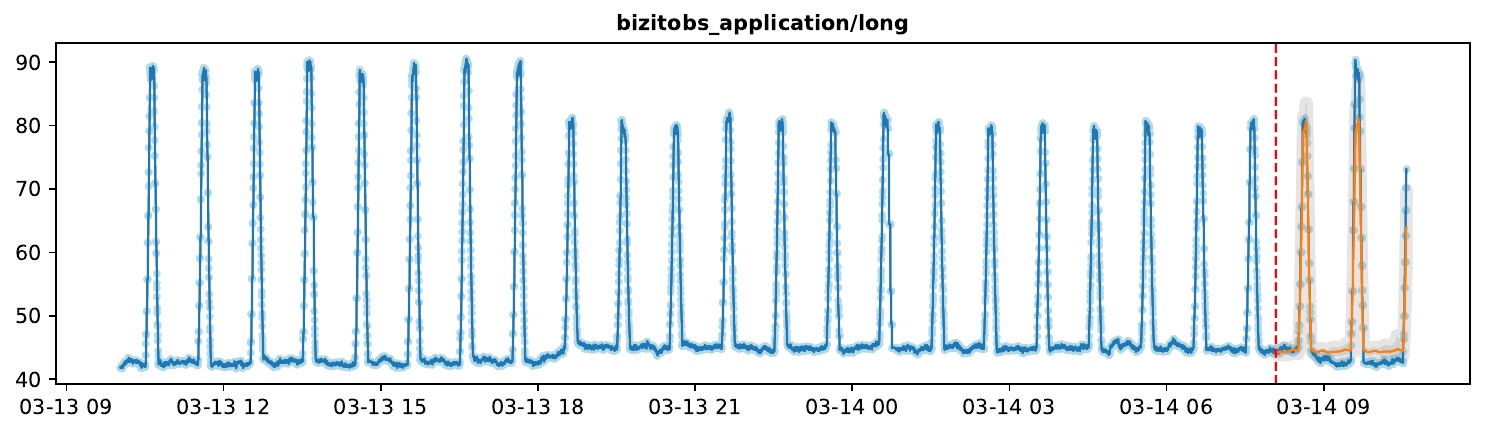}
    \end{minipage}
    \hfill
    \begin{minipage}{\width\textwidth}
        \centering
        \includegraphics[width=\linewidth]{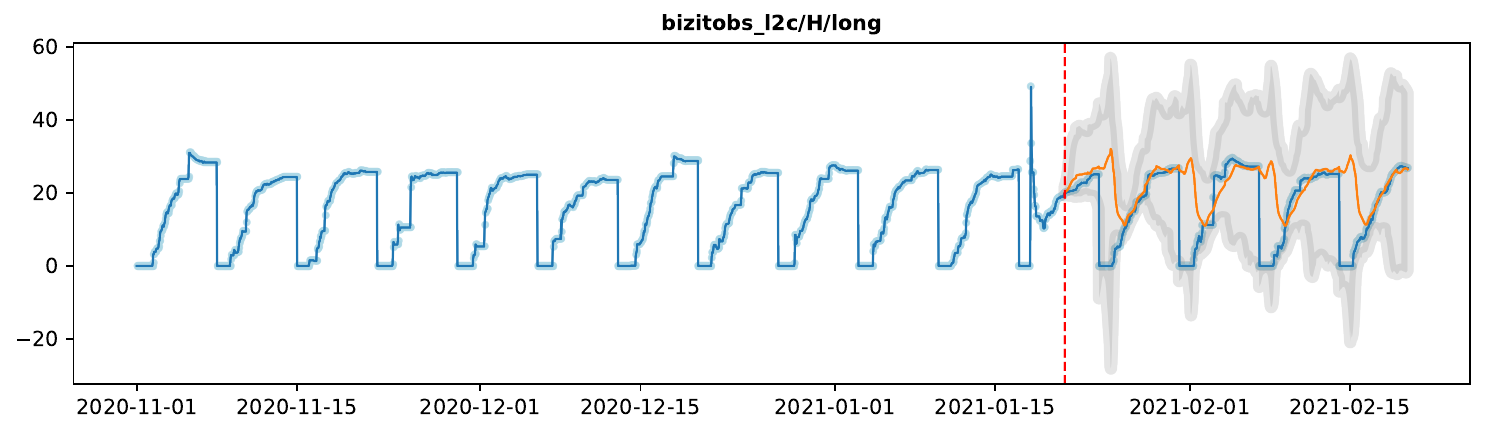}
    \end{minipage}

    \vspace{\gap} 

    \begin{minipage}{\width\textwidth}
        \centering
        \includegraphics[width=\linewidth]{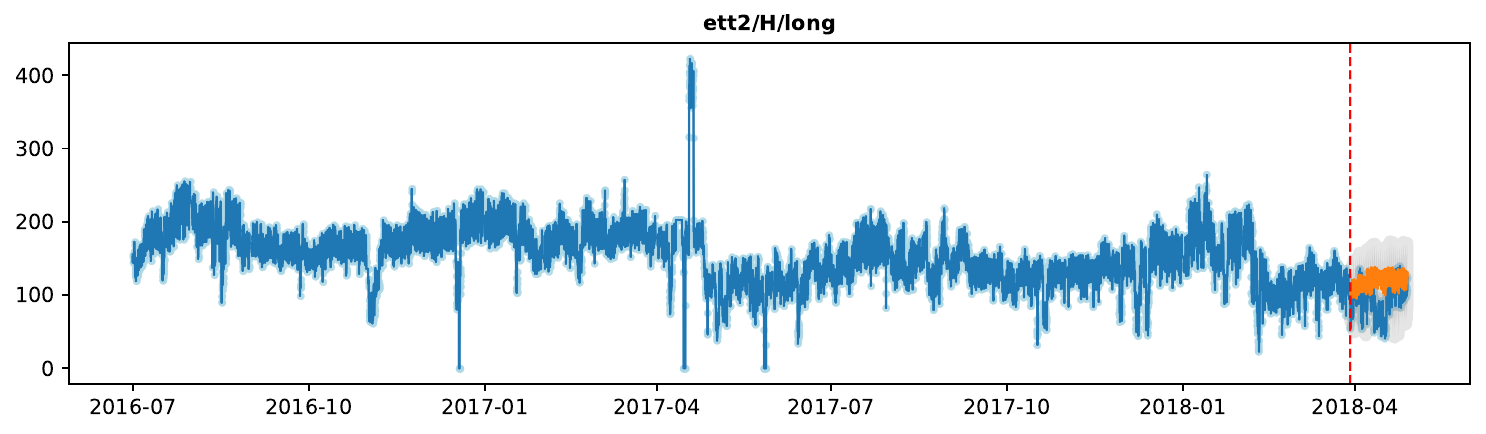}
    \end{minipage}
    \hfill
    \begin{minipage}{\width\textwidth}
        \centering
        \includegraphics[width=\linewidth]{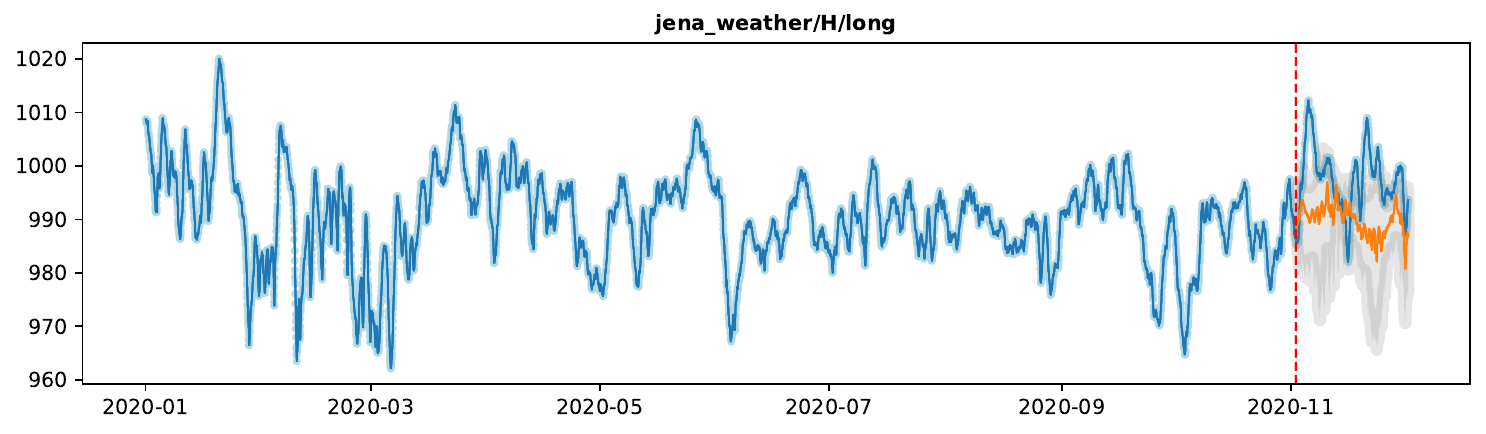}
    \end{minipage}

    \vspace{\gap} 

    \begin{minipage}{\width\textwidth}
        \centering
        \includegraphics[width=\linewidth]{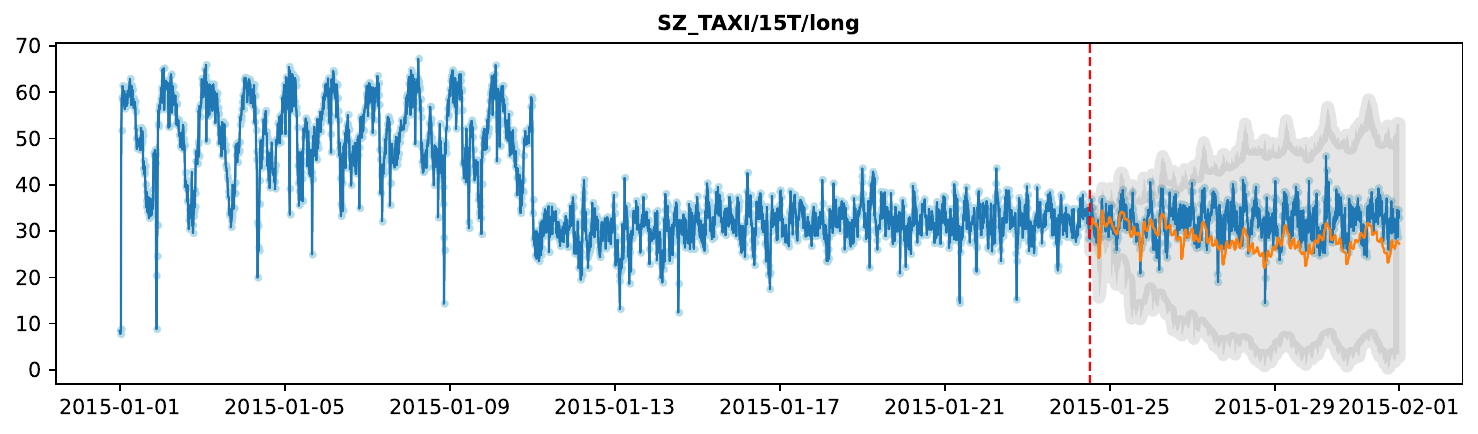}
    \end{minipage}
    \hfill
    \begin{minipage}{\width\textwidth}
        \centering
        \includegraphics[width=\linewidth]{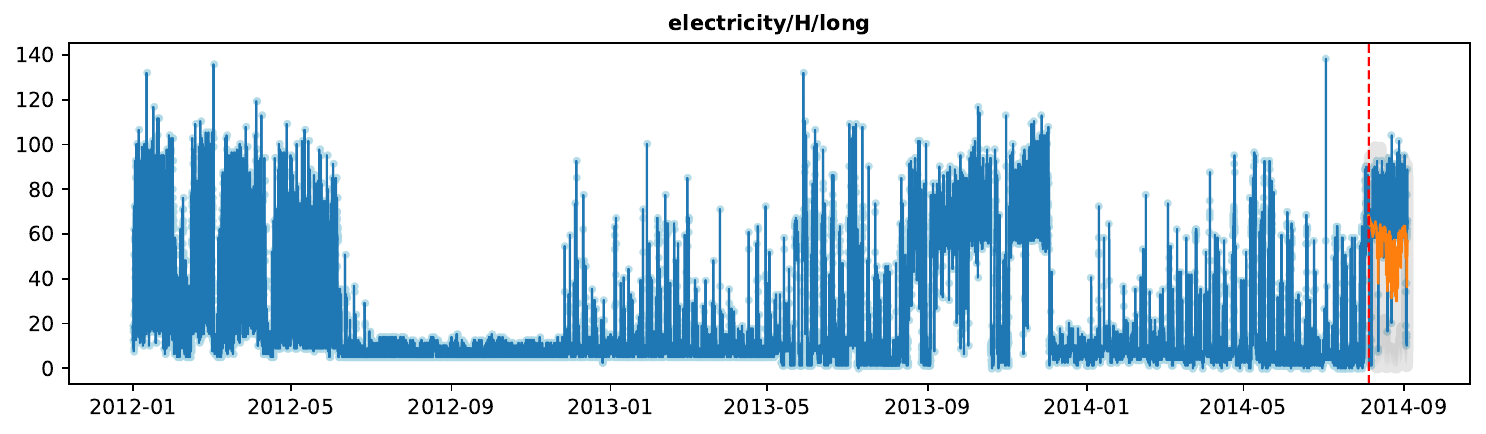}
    \end{minipage}

    \caption{Visualization of the \tabpfnts' predictions on some of the \textbf{long-term} benchmarking tasks.}
    \label{append-vis-long}
\end{figure}

\pagebreak

\begin{table}[H]
    \centering
    \caption{Probabilistic forecasting performance (WQL scores) of all models. Lower is better.}
    \label{table-wql-scores}
    \resizebox{\textwidth}{!}{

\begin{tabular}{rrrccccccccccccccc}
\toprule
 &  &  & Tabular Foundation Model & \multicolumn{8}{c}{Time-Series Foundation Model} & \multicolumn{3}{c}{Deep Learning Time-Series Model} & \multicolumn{3}{c}{Statistical Time-Series Model} \\
\cmidrule(lr){4-4} \cmidrule(lr){5-12} \cmidrule(lr){13-15} \cmidrule(lr){16-18}
 &  &  & \rotatebox{45}{TabPFN-TS} & \rotatebox{45}{Chronos-Bolt Base} & \rotatebox{45}{Chronos-Bolt Small} & \rotatebox{45}{Chronos-Bolt Tiny} & \rotatebox{45}{Moirai2} & \rotatebox{45}{Sundial} & \rotatebox{45}{TiRex} & \rotatebox{45}{TimesFM2.0-500M} & \rotatebox{45}{Toto} & \rotatebox{45}{DeepAR} & \rotatebox{45}{PatchTST} & \rotatebox{45}{TFT} & \rotatebox{45}{AutoARIMA} & \rotatebox{45}{AutoTheta} & \rotatebox{45}{Seasonal Naive} \\
Dataset & Freq. & Term &  &  &  &  &  &  &  &  &  &  &  &  &  &  &  \\
\midrule
\multirow[t]{4}{*}{bitbrains\_fast\_storage} & \multirow[t]{3}{*}{5T} & long & 0.885 & 0.748 & 0.753 & 0.750 & 0.807 & 0.811 & 0.672 & 0.908 & \textbf{0.669} & 1.010 & 0.669 & 0.734 & 1.290 & 1.360 & 1.290 \\
 &  & medium & 0.949 & 0.755 & 0.867 & 0.814 & 0.694 & 0.728 & 0.638 & 0.881 & 0.629 & 0.990 & 0.642 & \textbf{0.610} & 1.270 & 1.450 & 1.270 \\
 &  & short & 0.662 & 0.454 & 0.435 & 0.420 & 0.427 & 0.462 & 0.380 & 0.447 & \textbf{0.371} & 0.493 & 0.471 & 0.451 & 1.210 & 0.731 & 1.210 \\
\cline{2-18}
 & H & short & 0.670 & 0.774 & 0.589 & 0.593 & 0.615 & 0.764 & 0.700 & 0.688 & 0.623 & 0.778 & \textbf{0.549} & 0.595 & 0.844 & 1.150 & 1.080 \\
\cline{1-18} \cline{2-18}
\multirow[t]{4}{*}{bitbrains\_rnd} & \multirow[t]{3}{*}{5T} & long & 0.819 & 0.756 & 0.756 & 0.917 & \textbf{0.570} & 0.715 & 0.632 & 0.706 & 0.589 & 0.672 & 0.664 & 0.624 & 1.290 & 1.600 & 1.290 \\
 &  & medium & 0.819 & 0.605 & 0.792 & 0.697 & \textbf{0.596} & 0.730 & 0.604 & 0.727 & 0.628 & 0.647 & 0.620 & 0.628 & 1.260 & 1.470 & 1.260 \\
 &  & short & 0.608 & 0.438 & 0.453 & 0.482 & 0.404 & 0.433 & 0.404 & 0.461 & \textbf{0.399} & 0.557 & 0.474 & 0.486 & 1.100 & 0.741 & 1.100 \\
\cline{2-18}
 & H & short & 0.742 & 0.624 & 0.623 & 0.604 & 0.670 & 0.725 & 0.611 & 0.649 & 0.593 & \textbf{0.585} & 0.603 & 0.650 & 0.874 & 1.380 & 1.300 \\
\cline{1-18} \cline{2-18}
\multirow[t]{3}{*}{bizitobs\_application} & \multirow[t]{3}{*}{10S} & long & 0.049 & 0.109 & 0.092 & 0.137 & 0.056 & 0.061 & 0.052 & 0.057 & 0.053 & 0.083 & 0.054 & 0.056 & 0.973 & \textbf{0.035} & 0.973 \\
 &  & medium & 0.041 & 0.104 & 0.085 & 0.115 & 0.037 & 0.046 & 0.038 & 0.033 & 0.034 & 0.053 & 0.047 & 0.047 & 0.042 & \textbf{0.024} & 0.042 \\
 &  & short & 0.015 & 0.054 & 0.035 & 0.070 & 0.013 & 0.016 & 0.011 & 0.014 & 0.012 & 0.064 & 0.022 & 0.090 & 0.035 & \textbf{0.010} & 0.035 \\
\cline{1-18} \cline{2-18}
\multirow[t]{6}{*}{bizitobs\_l2c} & \multirow[t]{3}{*}{5T} & long & 0.306 & 0.738 & 0.790 & 0.722 & 0.300 & 0.310 & \textbf{0.269} & 0.748 & 0.533 & 0.719 & 0.324 & 0.472 & 0.674 & 0.632 & 0.674 \\
 &  & medium & 0.261 & 0.445 & 0.462 & 0.420 & 0.261 & \textbf{0.234} & 0.251 & 0.529 & 0.316 & 0.589 & 0.332 & 0.346 & 0.530 & 0.415 & 0.530 \\
 &  & short & 0.084 & 0.074 & 0.073 & 0.075 & 0.084 & \textbf{0.067} & 0.076 & 0.084 & 0.069 & 0.179 & 0.074 & 0.077 & 0.262 & 0.080 & 0.262 \\
\cline{2-18}
 & \multirow[t]{3}{*}{H} & long & 0.292 & 0.278 & 0.295 & 0.306 & 0.321 & 0.325 & \textbf{0.268} & 0.728 & 0.369 & 0.338 & 0.291 & 0.286 & 0.787 & 0.819 & 1.820 \\
 &  & medium & \textbf{0.237} & 0.254 & 0.285 & 0.304 & 0.274 & 0.276 & 0.252 & 0.640 & 0.356 & 0.345 & 0.263 & 0.345 & 0.813 & 0.892 & 1.420 \\
 &  & short & 0.210 & \textbf{0.189} & 0.204 & 0.203 & 0.235 & 0.223 & 0.212 & 0.345 & 0.199 & 0.789 & 0.217 & 0.401 & 0.547 & 0.507 & 0.536 \\
\cline{1-18} \cline{2-18}
\multirow[t]{3}{*}{bizitobs\_service} & \multirow[t]{3}{*}{10S} & long & 0.052 & 0.113 & 0.095 & 0.133 & 0.054 & 0.057 & 0.053 & 0.062 & \textbf{0.051} & 0.070 & 0.057 & 0.056 & 0.056 & 0.052 & 0.056 \\
 &  & medium & 0.041 & 0.096 & 0.081 & 0.113 & 0.034 & 0.044 & \textbf{0.023} & 0.038 & 0.027 & 0.044 & 0.045 & 0.044 & 0.049 & 0.027 & 0.049 \\
 &  & short & 0.019 & 0.051 & 0.031 & 0.065 & 0.014 & 0.016 & 0.012 & 0.015 & \textbf{0.011} & 0.032 & 0.025 & 0.025 & 0.040 & 0.013 & 0.040 \\
\cline{1-18} \cline{2-18}
car\_parts & M & short & 0.970 & 0.995 & 1.007 & 1.001 & 0.936 & 1.189 & 0.995 & 1.046 & 0.899 & 0.953 & 1.000 & \textbf{0.890} & 1.290 & 1.340 & 1.720 \\
\cline{1-18} \cline{2-18}
covid\_deaths & D & short & 0.041 & 0.047 & 0.043 & 0.067 & 0.028 & 0.131 & 0.032 & 0.062 & \textbf{0.027} & 0.177 & 0.067 & 0.037 & 0.030 & 0.095 & 0.125 \\
\cline{1-18} \cline{2-18}
\multirow[t]{8}{*}{electricity} & \multirow[t]{3}{*}{15T} & long & 0.081 & 0.084 & 0.086 & 0.092 & 0.083 & 0.082 & \textbf{0.078} & 0.083 & 0.086 & 0.155 & 0.081 & 0.084 & 0.129 & 0.401 & 0.129 \\
 &  & medium & 0.083 & 0.083 & 0.087 & 0.089 & 0.080 & 0.082 & \textbf{0.079} & 0.080 & 0.086 & 0.119 & 0.086 & 0.094 & 0.124 & 0.328 & 0.124 \\
 &  & short & 0.097 & 0.082 & 0.082 & 0.086 & \textbf{0.077} & 0.084 & 0.091 & 0.079 & 0.099 & 0.152 & 0.134 & 0.184 & 0.165 & 0.140 & 0.165 \\
\cline{2-18}
 & D & short & 0.063 & 0.055 & 0.058 & 0.057 & \textbf{0.054} & 0.064 & 0.054 & 0.060 & 0.059 & 0.078 & 0.083 & 0.084 & 0.083 & 0.088 & 0.122 \\
\cline{2-18}
 & \multirow[t]{3}{*}{H} & long & 0.108 & 0.098 & 0.102 & 0.102 & 0.097 & 0.093 & 0.094 & 0.089 & \textbf{0.083} & 0.176 & 0.104 & 0.094 & 0.190 & 0.300 & 0.190 \\
 &  & medium & 0.088 & 0.081 & 0.084 & 0.092 & 0.080 & 0.080 & 0.079 & \textbf{0.073} & 0.075 & 0.454 & 0.081 & 0.091 & 0.156 & 0.254 & 0.156 \\
 &  & short & 0.072 & 0.064 & 0.067 & 0.072 & 0.065 & 0.069 & 0.064 & \textbf{0.054} & 0.069 & 0.094 & 0.079 & 0.089 & 0.109 & 0.177 & 0.109 \\
\cline{2-18}
 & W & short & 0.055 & 0.047 & 0.048 & \textbf{0.047} & 0.066 & 0.072 & 0.048 & 0.049 & 0.064 & 0.092 & 0.095 & 0.107 & 0.100 & 0.101 & 0.099 \\
\cline{1-18} \cline{2-18}
\multirow[t]{8}{*}{ett1} & \multirow[t]{3}{*}{15T} & long & 0.259 & 0.298 & 0.296 & 0.332 & 0.268 & 0.253 & \textbf{0.234} & 0.283 & 0.251 & 2.220 & 0.247 & 0.280 & 0.396 & 1.390 & 0.396 \\
 &  & medium & 0.253 & 0.281 & 0.288 & 0.299 & 0.260 & 0.260 & \textbf{0.237} & 0.278 & 0.260 & 0.315 & 0.250 & 0.247 & 0.352 & 1.130 & 0.352 \\
 &  & short & 0.167 & \textbf{0.158} & 0.169 & 0.179 & 0.160 & 0.177 & 0.161 & 0.168 & 0.162 & 0.320 & 0.191 & 0.245 & 0.241 & 0.410 & 0.241 \\
\cline{2-18}
 & D & short & 0.298 & 0.287 & 0.283 & 0.301 & 0.287 & 0.373 & \textbf{0.277} & 0.281 & 0.284 & 0.293 & 0.304 & 0.330 & 0.279 & 0.341 & 0.515 \\
\cline{2-18}
 & \multirow[t]{3}{*}{H} & long & 0.295 & 0.311 & 0.337 & 0.317 & 0.323 & 0.283 & \textbf{0.258} & 0.310 & 0.267 & 0.469 & 0.297 & 0.313 & 0.430 & 1.940 & 0.616 \\
 &  & medium & 0.283 & 0.303 & 0.295 & 0.280 & 0.287 & 0.269 & \textbf{0.252} & 0.282 & 0.254 & 0.535 & 0.273 & 0.316 & 0.384 & 1.650 & 0.540 \\
 &  & short & 0.194 & 0.181 & 0.189 & 0.195 & 0.185 & 0.190 & \textbf{0.176} & 0.192 & 0.194 & 0.233 & 0.190 & 0.199 & 0.223 & 0.668 & 0.250 \\
\cline{2-18}
 & W & short & 0.284 & 0.296 & 0.293 & 0.275 & \textbf{0.249} & 0.404 & 0.278 & 0.272 & 0.263 & 0.686 & 0.323 & 0.406 & 0.305 & 0.319 & 0.338 \\
\cline{1-18} \cline{2-18}
\multirow[t]{8}{*}{ett2} & \multirow[t]{3}{*}{15T} & long & 0.101 & 0.111 & 0.118 & 0.119 & 0.102 & 0.098 & 0.092 & 0.106 & \textbf{0.088} & 0.304 & 0.098 & 0.109 & 0.165 & 0.169 & 0.165 \\
 &  & medium & 0.100 & 0.110 & 0.119 & 0.113 & 0.098 & 0.096 & \textbf{0.089} & 0.105 & 0.093 & 0.258 & 0.094 & 0.104 & 0.143 & 0.150 & 0.143 \\
 &  & short & 0.073 & 0.067 & 0.070 & 0.070 & 0.066 & 0.069 & 0.066 & \textbf{0.065} & 0.068 & 0.378 & 0.076 & 0.081 & 0.096 & 0.077 & 0.096 \\
\cline{2-18}
 & D & short & 0.126 & 0.094 & \textbf{0.091} & 0.095 & 0.093 & 0.103 & 0.094 & 0.108 & 0.111 & 0.207 & 0.131 & 0.096 & 0.125 & 0.164 & 0.205 \\
\cline{2-18}
 & \multirow[t]{3}{*}{H} & long & 0.139 & 0.117 & 0.121 & 0.124 & 0.109 & 0.117 & 0.114 & 0.125 & \textbf{0.108} & 0.196 & 0.130 & 0.138 & 0.272 & 0.336 & 0.287 \\
 &  & medium & 0.121 & 0.115 & 0.118 & 0.116 & 0.111 & 0.114 & 0.107 & 0.110 & \textbf{0.102} & 0.281 & 0.125 & 0.122 & 0.245 & 0.284 & 0.241 \\
 &  & short & 0.073 & \textbf{0.063} & 0.065 & 0.065 & 0.064 & 0.072 & 0.064 & 0.066 & 0.065 & 0.122 & 0.074 & 0.078 & 0.089 & 0.102 & 0.094 \\
\cline{2-18}
 & W & short & 0.099 & 0.088 & 0.094 & 0.095 & \textbf{0.085} & 0.098 & 0.087 & 0.110 & 0.106 & 0.728 & 0.142 & 0.160 & 0.136 & 0.160 & 0.169 \\
\cline{1-18} \cline{2-18}
\multirow[t]{2}{*}{hierarchical\_sales} & D & short & 0.592 & 0.576 & 0.582 & 0.581 & 0.577 & 0.649 & \textbf{0.570} & 0.576 & 0.570 & 0.600 & 0.590 & 0.600 & 0.735 & 0.967 & 2.360 \\
\cline{2-18}
 & W & short & 0.345 & 0.353 & 0.354 & 0.353 & 0.352 & 0.390 & 0.348 & \textbf{0.330} & 0.356 & 0.379 & 0.358 & 0.382 & 0.485 & 0.474 & 1.030 \\
\cline{1-18} \cline{2-18}
hospital & M & short & 0.054 & 0.057 & 0.058 & 0.059 & 0.052 & 0.061 & 0.051 & \textbf{0.050} & 0.052 & 0.062 & 0.064 & 0.058 & 0.060 & 0.055 & 0.062 \\
\cline{1-18} \cline{2-18}
\multirow[t]{7}{*}{jena\_weather} & \multirow[t]{3}{*}{10T} & long & 0.053 & 0.064 & 0.063 & 0.079 & 0.060 & 0.056 & 0.049 & \textbf{0.035} & 0.050 & 0.143 & 0.066 & 0.052 & 0.304 & 0.424 & 0.304 \\
 &  & medium & 0.054 & 0.057 & 0.060 & 0.068 & 0.059 & 0.054 & 0.048 & \textbf{0.031} & 0.049 & 0.073 & 0.065 & 0.052 & 0.277 & 0.350 & 0.277 \\
 &  & short & 0.034 & 0.033 & 0.037 & 0.042 & 0.036 & 0.031 & 0.027 & \textbf{0.016} & 0.027 & 0.063 & 0.064 & 0.069 & 0.155 & 0.130 & 0.155 \\
\cline{2-18}
 & D & short & 0.047 & 0.045 & 0.047 & 0.047 & \textbf{0.043} & 0.048 & 0.044 & 0.058 & 0.051 & 0.062 & 0.053 & 0.069 & 0.080 & 0.082 & 0.297 \\
\cline{2-18}
 & \multirow[t]{3}{*}{H} & long & 0.103 & 0.062 & 0.068 & 0.066 & 0.058 & 0.066 & 0.059 & 0.068 & \textbf{0.057} & 0.197 & 0.076 & 0.090 & 0.230 & 1.290 & 0.598 \\
 &  & medium & 0.058 & 0.054 & 0.058 & 0.058 & 0.055 & 0.058 & 0.053 & 0.066 & \textbf{0.053} & 0.078 & 0.069 & 0.073 & 0.211 & 0.832 & 0.486 \\
 &  & short & 0.042 & 0.042 & 0.043 & 0.042 & 0.042 & 0.050 & \textbf{0.041} & 0.045 & 0.042 & 0.699 & 0.050 & 0.048 & 0.143 & 0.296 & 0.173 \\
\cline{1-18} \cline{2-18}
\multirow[t]{4}{*}{kdd\_cup\_2018} & D & short & \textbf{0.362} & 0.372 & 0.373 & 0.365 & 0.389 & 0.396 & 0.376 & 0.378 & 0.387 & 0.383 & 0.401 & 0.380 & 0.393 & 0.459 & 0.888 \\
\cline{2-18}
 & \multirow[t]{3}{*}{H} & long & 0.478 & \textbf{0.300} & 0.419 & 0.472 & 0.518 & 0.375 & 0.441 & 0.518 & 0.457 & 1.090 & 0.477 & 0.503 & 1.050 & 0.970 & 1.250 \\
 &  & medium & 0.450 & \textbf{0.301} & 0.364 & 0.416 & 0.495 & 0.377 & 0.421 & 0.466 & 0.441 & 0.442 & 0.442 & 0.472 & 0.851 & 0.791 & 0.949 \\
 &  & short & 0.418 & \textbf{0.246} & 0.267 & 0.313 & 0.426 & 0.351 & 0.378 & 0.376 & 0.403 & 0.517 & 0.457 & 0.467 & 0.559 & 0.531 & 0.559 \\
\cline{1-18} \cline{2-18}
\multirow[t]{7}{*}{loop\_seattle} & \multirow[t]{3}{*}{5T} & long & 0.090 & 0.129 & 0.125 & 0.121 & 0.087 & 0.084 & 0.084 & 0.114 & \textbf{0.077} & 0.184 & 0.095 & 0.088 & 0.137 & 0.231 & 0.137 \\
 &  & medium & 0.087 & 0.116 & 0.119 & 0.116 & 0.080 & 0.077 & 0.078 & 0.110 & \textbf{0.072} & 0.118 & 0.095 & 0.092 & 0.123 & 0.240 & 0.123 \\
 &  & short & 0.053 & 0.055 & 0.055 & 0.055 & \textbf{0.046} & 0.050 & 0.048 & 0.051 & 0.048 & 0.072 & 0.066 & 0.065 & 0.081 & 0.082 & 0.081 \\
\cline{2-18}
 & D & short & 0.043 & 0.044 & 0.045 & 0.046 & 0.043 & 0.047 & 0.042 & \textbf{0.041} & 0.044 & 0.052 & 0.046 & 0.048 & 0.078 & 0.072 & 0.131 \\
\cline{2-18}
 & \multirow[t]{3}{*}{H} & long & 0.063 & 0.076 & 0.082 & 0.087 & 0.066 & 0.072 & \textbf{0.060} & 0.066 & 0.065 & 0.068 & 0.069 & 0.068 & 0.193 & 0.468 & 0.245 \\
 &  & medium & 0.067 & 0.076 & 0.082 & 0.087 & 0.069 & 0.075 & \textbf{0.064} & 0.067 & 0.064 & 0.072 & 0.071 & 0.069 & 0.154 & 0.390 & 0.206 \\
 &  & short & 0.063 & 0.065 & 0.066 & 0.071 & 0.062 & 0.067 & \textbf{0.058} & 0.059 & 0.063 & 0.066 & 0.076 & 0.073 & 0.108 & 0.165 & 0.108 \\
\cline{1-18} \cline{2-18}
m4\_daily & D & short & 0.023 & 0.021 & 0.021 & 0.021 & \textbf{0.020} & 0.027 & 0.021 & 0.021 & 0.022 & 0.030 & 0.023 & 0.023 & 0.023 & 0.024 & 0.026 \\
\cline{1-18} \cline{2-18}
m4\_hourly & H & short & 0.030 & 0.025 & 0.020 & 0.021 & 0.024 & 0.023 & 0.020 & \textbf{0.011} & 0.035 & 0.133 & 0.039 & 0.040 & 0.034 & 0.041 & 0.040 \\
\cline{1-18} \cline{2-18}
m4\_monthly & M & short & 0.094 & 0.094 & 0.094 & 0.095 & 0.095 & 0.116 & 0.091 & \textbf{0.067} & 0.097 & 0.184 & 0.102 & 0.113 & 0.098 & 0.098 & 0.126 \\
\cline{1-18} \cline{2-18}
m4\_quarterly & Q & short & 0.078 & 0.077 & 0.078 & 0.079 & 0.075 & 0.093 & 0.073 & \textbf{0.062} & 0.078 & 0.083 & 0.083 & 0.083 & 0.082 & 0.079 & 0.099 \\
\cline{1-18} \cline{2-18}
m4\_weekly & W & short & 0.037 & 0.038 & 0.038 & 0.041 & 0.040 & 0.043 & \textbf{0.037} & 0.042 & 0.049 & 0.062 & 0.040 & 0.049 & 0.050 & 0.053 & 0.073 \\
\cline{1-18} \cline{2-18}
m4\_yearly & A & short & 0.118 & 0.121 & 0.128 & 0.129 & 0.116 & 0.160 & 0.116 & \textbf{0.091} & 0.122 & 0.113 & 0.117 & 0.110 & 0.130 & 0.115 & 0.138 \\
\cline{1-18} \cline{2-18}
\multirow[t]{4}{*}{m\_dense} & D & short & 0.061 & 0.069 & 0.072 & 0.082 & 0.070 & 0.067 & 0.068 & \textbf{0.060} & 0.075 & 0.076 & 0.070 & 0.077 & 0.135 & 0.126 & 0.294 \\
\cline{2-18}
 & \multirow[t]{3}{*}{H} & long & 0.165 & 0.170 & 0.146 & 0.198 & 0.120 & 0.130 & 0.115 & 0.127 & 0.128 & 0.130 & 0.120 & \textbf{0.115} & 0.270 & 1.430 & 0.552 \\
 &  & medium & 0.160 & 0.157 & 0.134 & 0.155 & 0.120 & 0.128 & 0.116 & 0.127 & 0.121 & 0.118 & 0.127 & \textbf{0.114} & 0.255 & 1.210 & 0.479 \\
 &  & short & 0.155 & \textbf{0.125} & 0.133 & 0.140 & 0.132 & 0.133 & 0.128 & 0.139 & 0.148 & 0.128 & 0.173 & 0.139 & 0.281 & 0.549 & 0.281 \\
\cline{1-18} \cline{2-18}
restaurant & D & short & 0.263 & 0.264 & 0.264 & 0.276 & 0.260 & 0.286 & \textbf{0.255} & 0.261 & 0.297 & 0.270 & 0.262 & 0.284 & 0.362 & 0.329 & 0.907 \\
\cline{1-18} \cline{2-18}
\multirow[t]{3}{*}{saugeen} & D & short & 0.373 & 0.338 & 0.354 & 0.339 & \textbf{0.328} & 0.379 & 0.358 & 0.408 & 0.353 & 0.572 & 0.408 & 0.419 & 0.564 & 0.669 & 0.754 \\
\cline{2-18}
 & M & short & \textbf{0.276} & 0.296 & 0.293 & 0.288 & 0.291 & 0.332 & 0.297 & 0.342 & 0.299 & 0.689 & 0.372 & 0.340 & 0.326 & 0.373 & 0.445 \\
\cline{2-18}
 & W & short & 0.395 & 0.363 & 0.372 & 0.364 & 0.419 & 0.406 & \textbf{0.350} & 0.601 & 0.390 & 0.397 & 0.484 & 0.491 & 0.549 & 0.734 & 0.855 \\
\cline{1-18} \cline{2-18}
\multirow[t]{8}{*}{solar} & \multirow[t]{3}{*}{10T} & long & 0.331 & 0.443 & 0.497 & 0.534 & 0.408 & 0.365 & \textbf{0.321} & 0.498 & 0.352 & 0.549 & 0.339 & 0.379 & 0.786 & 6.640 & 0.786 \\
 &  & medium & \textbf{0.326} & 0.436 & 0.453 & 0.495 & 0.412 & 0.373 & 0.334 & 0.516 & 0.353 & 0.485 & 0.356 & 0.362 & 0.771 & 5.670 & 0.771 \\
 &  & short & 0.458 & 0.511 & 0.498 & 0.488 & \textbf{0.350} & 0.444 & 0.541 & 0.804 & 0.541 & 0.933 & 1.370 & 0.618 & 0.860 & 2.360 & 0.860 \\
\cline{2-18}
 & D & short & \textbf{0.269} & 0.287 & 0.286 & 0.282 & 0.303 & 0.324 & 0.273 & 0.278 & 0.290 & 0.682 & 0.287 & 0.277 & 0.282 & 0.286 & 0.757 \\
\cline{2-18}
 & \multirow[t]{3}{*}{H} & long & 0.351 & 0.405 & 0.373 & 0.354 & 0.332 & 0.293 & \textbf{0.287} & 0.493 & 0.331 & 0.381 & 0.353 & 0.401 & 0.607 & 7.320 & 1.470 \\
 &  & medium & 0.313 & 0.368 & 0.356 & 0.342 & 0.333 & 0.309 & \textbf{0.285} & 0.376 & 0.331 & 0.352 & 0.344 & 0.330 & 0.557 & 6.130 & 1.270 \\
 &  & short & 0.336 & 0.298 & 0.303 & 0.313 & 0.342 & 0.329 & \textbf{0.273} & 0.406 & 0.328 & 0.389 & 0.340 & 0.367 & 0.628 & 2.330 & 0.628 \\
\cline{2-18}
 & W & short & 0.120 & 0.133 & 0.136 & 0.132 & 0.163 & 0.148 & 0.144 & 0.171 & 0.186 & 0.242 & 0.162 & \textbf{0.114} & 0.152 & 0.155 & 0.236 \\
\cline{1-18} \cline{2-18}
\multirow[t]{4}{*}{sz\_taxi} & \multirow[t]{3}{*}{15T} & long & 0.242 & 0.248 & 0.245 & 0.240 & 0.217 & 0.221 & \textbf{0.198} & 0.227 & 0.202 & 0.286 & 0.281 & 0.241 & 0.398 & 0.629 & 0.554 \\
 &  & medium & 0.230 & 0.244 & 0.246 & 0.240 & 0.214 & 0.228 & \textbf{0.202} & 0.229 & 0.205 & 0.210 & 0.220 & 0.206 & 0.351 & 0.529 & 0.454 \\
 &  & short & 0.209 & 0.202 & 0.203 & 0.203 & 0.201 & 0.223 & 0.200 & \textbf{0.199} & 0.203 & 0.219 & 0.207 & 0.222 & 0.309 & 0.288 & 0.309 \\
\cline{2-18}
 & H & short & 0.140 & 0.136 & 0.137 & 0.137 & 0.136 & 0.154 & 0.135 & \textbf{0.135} & 0.137 & 0.139 & 0.144 & 0.144 & 0.170 & 0.232 & 0.229 \\
\cline{1-18} \cline{2-18}
temperature\_rain & D & short & 0.569 & \textbf{0.538} & 0.544 & 0.548 & 0.561 & 0.620 & 0.547 & 0.586 & 0.560 & 0.682 & 0.644 & 0.592 & 0.694 & 0.761 & 1.630 \\
\cline{1-18} \cline{2-18}
\multirow[t]{3}{*}{us\_births} & D & short & \textbf{0.016} & 0.026 & 0.028 & 0.037 & 0.020 & 0.022 & 0.023 & 0.019 & 0.026 & 0.028 & 0.025 & 0.016 & 0.074 & 0.075 & 0.144 \\
\cline{2-18}
 & M & short & 0.015 & 0.019 & 0.016 & 0.017 & 0.017 & 0.028 & 0.012 & 0.011 & 0.013 & 0.016 & 0.017 & 0.021 & \textbf{0.010} & 0.019 & 0.017 \\
\cline{2-18}
 & W & short & \textbf{0.011} & 0.013 & 0.013 & 0.014 & 0.011 & 0.017 & 0.012 & 0.013 & 0.014 & 0.017 & 0.015 & 0.019 & 0.018 & 0.018 & 0.022 \\
\cline{1-18} \cline{2-18}
\bottomrule
\end{tabular}
}
\end{table}


\begin{table}[H]
    \centering
    \caption{Point forecasting performance (MASE scores) of all models. Lower is better.}
    \label{table-mase-scores}
    \resizebox{\textwidth}{!}{

\begin{tabular}{rrrccccccccccccccc}
\toprule
 &  &  & Tabular Foundation Model & \multicolumn{8}{c}{Time-Series Foundation Model} & \multicolumn{3}{c}{Deep Learning Time-Series Model} & \multicolumn{3}{c}{Statistical Time-Series Model} \\
\cmidrule(lr){4-4} \cmidrule(lr){5-12} \cmidrule(lr){13-15} \cmidrule(lr){16-18}
 &  &  & \rotatebox{45}{TabPFN-TS} & \rotatebox{45}{Chronos-Bolt Base} & \rotatebox{45}{Chronos-Bolt Small} & \rotatebox{45}{Chronos-Bolt Tiny} & \rotatebox{45}{Moirai2} & \rotatebox{45}{Sundial} & \rotatebox{45}{TiRex} & \rotatebox{45}{TimesFM2.0-500M} & \rotatebox{45}{Toto} & \rotatebox{45}{DeepAR} & \rotatebox{45}{PatchTST} & \rotatebox{45}{TFT} & \rotatebox{45}{AutoARIMA} & \rotatebox{45}{AutoTheta} & \rotatebox{45}{Seasonal Naive} \\
Dataset & Freq. & Term &  &  &  &  &  &  &  &  &  &  &  &  &  &  &  \\
\midrule
\multirow[t]{4}{*}{bitbrains\_fast\_storage} & \multirow[t]{3}{*}{5T} & long & 0.885 & 0.748 & 0.753 & 0.750 & 0.807 & 0.811 & 0.672 & 0.908 & \textbf{0.669} & 1.010 & 0.669 & 0.734 & 1.290 & 1.360 & 1.290 \\
 &  & medium & 0.949 & 0.755 & 0.867 & 0.814 & 0.694 & 0.728 & 0.638 & 0.881 & 0.629 & 0.990 & 0.642 & \textbf{0.610} & 1.270 & 1.450 & 1.270 \\
 &  & short & 0.662 & 0.454 & 0.435 & 0.420 & 0.427 & 0.462 & 0.380 & 0.447 & \textbf{0.371} & 0.493 & 0.471 & 0.451 & 1.210 & 0.731 & 1.210 \\
\cline{2-18}
 & H & short & 0.670 & 0.774 & 0.589 & 0.593 & 0.615 & 0.764 & 0.700 & 0.688 & 0.623 & 0.778 & \textbf{0.549} & 0.595 & 0.844 & 1.150 & 1.080 \\
\cline{1-18} \cline{2-18}
\multirow[t]{4}{*}{bitbrains\_rnd} & \multirow[t]{3}{*}{5T} & long & 0.819 & 0.756 & 0.756 & 0.917 & \textbf{0.570} & 0.715 & 0.632 & 0.706 & 0.589 & 0.672 & 0.664 & 0.624 & 1.290 & 1.600 & 1.290 \\
 &  & medium & 0.819 & 0.605 & 0.792 & 0.697 & \textbf{0.596} & 0.730 & 0.604 & 0.727 & 0.628 & 0.647 & 0.620 & 0.628 & 1.260 & 1.470 & 1.260 \\
 &  & short & 0.608 & 0.438 & 0.453 & 0.482 & 0.404 & 0.433 & 0.404 & 0.461 & \textbf{0.399} & 0.557 & 0.474 & 0.486 & 1.100 & 0.741 & 1.100 \\
\cline{2-18}
 & H & short & 0.742 & 0.624 & 0.623 & 0.604 & 0.670 & 0.725 & 0.611 & 0.649 & 0.593 & \textbf{0.585} & 0.603 & 0.650 & 0.874 & 1.380 & 1.300 \\
\cline{1-18} \cline{2-18}
\multirow[t]{3}{*}{bizitobs\_application} & \multirow[t]{3}{*}{10S} & long & 0.049 & 0.109 & 0.092 & 0.137 & 0.056 & 0.061 & 0.052 & 0.057 & 0.053 & 0.083 & 0.054 & 0.056 & 0.973 & \textbf{0.035} & 0.973 \\
 &  & medium & 0.041 & 0.104 & 0.085 & 0.115 & 0.037 & 0.046 & 0.038 & 0.033 & 0.034 & 0.053 & 0.047 & 0.047 & 0.042 & \textbf{0.024} & 0.042 \\
 &  & short & 0.015 & 0.054 & 0.035 & 0.070 & 0.013 & 0.016 & 0.011 & 0.014 & 0.012 & 0.064 & 0.022 & 0.090 & 0.035 & \textbf{0.010} & 0.035 \\
\cline{1-18} \cline{2-18}
\multirow[t]{6}{*}{bizitobs\_l2c} & \multirow[t]{3}{*}{5T} & long & 0.306 & 0.738 & 0.790 & 0.722 & 0.300 & 0.310 & \textbf{0.269} & 0.748 & 0.533 & 0.719 & 0.324 & 0.472 & 0.674 & 0.632 & 0.674 \\
 &  & medium & 0.261 & 0.445 & 0.462 & 0.420 & 0.261 & \textbf{0.234} & 0.251 & 0.529 & 0.316 & 0.589 & 0.332 & 0.346 & 0.530 & 0.415 & 0.530 \\
 &  & short & 0.084 & 0.074 & 0.073 & 0.075 & 0.084 & \textbf{0.067} & 0.076 & 0.084 & 0.069 & 0.179 & 0.074 & 0.077 & 0.262 & 0.080 & 0.262 \\
\cline{2-18}
 & \multirow[t]{3}{*}{H} & long & 0.292 & 0.278 & 0.295 & 0.306 & 0.321 & 0.325 & \textbf{0.268} & 0.728 & 0.369 & 0.338 & 0.291 & 0.286 & 0.787 & 0.819 & 1.820 \\
 &  & medium & \textbf{0.237} & 0.254 & 0.285 & 0.304 & 0.274 & 0.276 & 0.252 & 0.640 & 0.356 & 0.345 & 0.263 & 0.345 & 0.813 & 0.892 & 1.420 \\
 &  & short & 0.210 & \textbf{0.189} & 0.204 & 0.203 & 0.235 & 0.223 & 0.212 & 0.345 & 0.199 & 0.789 & 0.217 & 0.401 & 0.547 & 0.507 & 0.536 \\
\cline{1-18} \cline{2-18}
\multirow[t]{3}{*}{bizitobs\_service} & \multirow[t]{3}{*}{10S} & long & 0.052 & 0.113 & 0.095 & 0.133 & 0.054 & 0.057 & 0.053 & 0.062 & \textbf{0.051} & 0.070 & 0.057 & 0.056 & 0.056 & 0.052 & 0.056 \\
 &  & medium & 0.041 & 0.096 & 0.081 & 0.113 & 0.034 & 0.044 & \textbf{0.023} & 0.038 & 0.027 & 0.044 & 0.045 & 0.044 & 0.049 & 0.027 & 0.049 \\
 &  & short & 0.019 & 0.051 & 0.031 & 0.065 & 0.014 & 0.016 & 0.012 & 0.015 & \textbf{0.011} & 0.032 & 0.025 & 0.025 & 0.040 & 0.013 & 0.040 \\
\cline{1-18} \cline{2-18}
car\_parts & M & short & 0.970 & 0.995 & 1.007 & 1.001 & 0.936 & 1.189 & 0.995 & 1.046 & 0.899 & 0.953 & 1.000 & \textbf{0.890} & 1.290 & 1.340 & 1.720 \\
\cline{1-18} \cline{2-18}
covid\_deaths & D & short & 0.041 & 0.047 & 0.043 & 0.067 & 0.028 & 0.131 & 0.032 & 0.062 & \textbf{0.027} & 0.177 & 0.067 & 0.037 & 0.030 & 0.095 & 0.125 \\
\cline{1-18} \cline{2-18}
\multirow[t]{8}{*}{electricity} & \multirow[t]{3}{*}{15T} & long & 0.081 & 0.084 & 0.086 & 0.092 & 0.083 & 0.082 & \textbf{0.078} & 0.083 & 0.086 & 0.155 & 0.081 & 0.084 & 0.129 & 0.401 & 0.129 \\
 &  & medium & 0.083 & 0.083 & 0.087 & 0.089 & 0.080 & 0.082 & \textbf{0.079} & 0.080 & 0.086 & 0.119 & 0.086 & 0.094 & 0.124 & 0.328 & 0.124 \\
 &  & short & 0.097 & 0.082 & 0.082 & 0.086 & \textbf{0.077} & 0.084 & 0.091 & 0.079 & 0.099 & 0.152 & 0.134 & 0.184 & 0.165 & 0.140 & 0.165 \\
\cline{2-18}
 & D & short & 0.063 & 0.055 & 0.058 & 0.057 & \textbf{0.054} & 0.064 & 0.054 & 0.060 & 0.059 & 0.078 & 0.083 & 0.084 & 0.083 & 0.088 & 0.122 \\
\cline{2-18}
 & \multirow[t]{3}{*}{H} & long & 0.108 & 0.098 & 0.102 & 0.102 & 0.097 & 0.093 & 0.094 & 0.089 & \textbf{0.083} & 0.176 & 0.104 & 0.094 & 0.190 & 0.300 & 0.190 \\
 &  & medium & 0.088 & 0.081 & 0.084 & 0.092 & 0.080 & 0.080 & 0.079 & \textbf{0.073} & 0.075 & 0.454 & 0.081 & 0.091 & 0.156 & 0.254 & 0.156 \\
 &  & short & 0.072 & 0.064 & 0.067 & 0.072 & 0.065 & 0.069 & 0.064 & \textbf{0.054} & 0.069 & 0.094 & 0.079 & 0.089 & 0.109 & 0.177 & 0.109 \\
\cline{2-18}
 & W & short & 0.055 & 0.047 & 0.048 & \textbf{0.047} & 0.066 & 0.072 & 0.048 & 0.049 & 0.064 & 0.092 & 0.095 & 0.107 & 0.100 & 0.101 & 0.099 \\
\cline{1-18} \cline{2-18}
\multirow[t]{8}{*}{ett1} & \multirow[t]{3}{*}{15T} & long & 0.259 & 0.298 & 0.296 & 0.332 & 0.268 & 0.253 & \textbf{0.234} & 0.283 & 0.251 & 2.220 & 0.247 & 0.280 & 0.396 & 1.390 & 0.396 \\
 &  & medium & 0.253 & 0.281 & 0.288 & 0.299 & 0.260 & 0.260 & \textbf{0.237} & 0.278 & 0.260 & 0.315 & 0.250 & 0.247 & 0.352 & 1.130 & 0.352 \\
 &  & short & 0.167 & \textbf{0.158} & 0.169 & 0.179 & 0.160 & 0.177 & 0.161 & 0.168 & 0.162 & 0.320 & 0.191 & 0.245 & 0.241 & 0.410 & 0.241 \\
\cline{2-18}
 & D & short & 0.298 & 0.287 & 0.283 & 0.301 & 0.287 & 0.373 & \textbf{0.277} & 0.281 & 0.284 & 0.293 & 0.304 & 0.330 & 0.279 & 0.341 & 0.515 \\
\cline{2-18}
 & \multirow[t]{3}{*}{H} & long & 0.295 & 0.311 & 0.337 & 0.317 & 0.323 & 0.283 & \textbf{0.258} & 0.310 & 0.267 & 0.469 & 0.297 & 0.313 & 0.430 & 1.940 & 0.616 \\
 &  & medium & 0.283 & 0.303 & 0.295 & 0.280 & 0.287 & 0.269 & \textbf{0.252} & 0.282 & 0.254 & 0.535 & 0.273 & 0.316 & 0.384 & 1.650 & 0.540 \\
 &  & short & 0.194 & 0.181 & 0.189 & 0.195 & 0.185 & 0.190 & \textbf{0.176} & 0.192 & 0.194 & 0.233 & 0.190 & 0.199 & 0.223 & 0.668 & 0.250 \\
\cline{2-18}
 & W & short & 0.284 & 0.296 & 0.293 & 0.275 & \textbf{0.249} & 0.404 & 0.278 & 0.272 & 0.263 & 0.686 & 0.323 & 0.406 & 0.305 & 0.319 & 0.338 \\
\cline{1-18} \cline{2-18}
\multirow[t]{8}{*}{ett2} & \multirow[t]{3}{*}{15T} & long & 0.101 & 0.111 & 0.118 & 0.119 & 0.102 & 0.098 & 0.092 & 0.106 & \textbf{0.088} & 0.304 & 0.098 & 0.109 & 0.165 & 0.169 & 0.165 \\
 &  & medium & 0.100 & 0.110 & 0.119 & 0.113 & 0.098 & 0.096 & \textbf{0.089} & 0.105 & 0.093 & 0.258 & 0.094 & 0.104 & 0.143 & 0.150 & 0.143 \\
 &  & short & 0.073 & 0.067 & 0.070 & 0.070 & 0.066 & 0.069 & 0.066 & \textbf{0.065} & 0.068 & 0.378 & 0.076 & 0.081 & 0.096 & 0.077 & 0.096 \\
\cline{2-18}
 & D & short & 0.126 & 0.094 & \textbf{0.091} & 0.095 & 0.093 & 0.103 & 0.094 & 0.108 & 0.111 & 0.207 & 0.131 & 0.096 & 0.125 & 0.164 & 0.205 \\
\cline{2-18}
 & \multirow[t]{3}{*}{H} & long & 0.139 & 0.117 & 0.121 & 0.124 & 0.109 & 0.117 & 0.114 & 0.125 & \textbf{0.108} & 0.196 & 0.130 & 0.138 & 0.272 & 0.336 & 0.287 \\
 &  & medium & 0.121 & 0.115 & 0.118 & 0.116 & 0.111 & 0.114 & 0.107 & 0.110 & \textbf{0.102} & 0.281 & 0.125 & 0.122 & 0.245 & 0.284 & 0.241 \\
 &  & short & 0.073 & \textbf{0.063} & 0.065 & 0.065 & 0.064 & 0.072 & 0.064 & 0.066 & 0.065 & 0.122 & 0.074 & 0.078 & 0.089 & 0.102 & 0.094 \\
\cline{2-18}
 & W & short & 0.099 & 0.088 & 0.094 & 0.095 & \textbf{0.085} & 0.098 & 0.087 & 0.110 & 0.106 & 0.728 & 0.142 & 0.160 & 0.136 & 0.160 & 0.169 \\
\cline{1-18} \cline{2-18}
\multirow[t]{2}{*}{hierarchical\_sales} & D & short & 0.592 & 0.576 & 0.582 & 0.581 & 0.577 & 0.649 & \textbf{0.570} & 0.576 & 0.570 & 0.600 & 0.590 & 0.600 & 0.735 & 0.967 & 2.360 \\
\cline{2-18}
 & W & short & 0.345 & 0.353 & 0.354 & 0.353 & 0.352 & 0.390 & 0.348 & \textbf{0.330} & 0.356 & 0.379 & 0.358 & 0.382 & 0.485 & 0.474 & 1.030 \\
\cline{1-18} \cline{2-18}
hospital & M & short & 0.054 & 0.057 & 0.058 & 0.059 & 0.052 & 0.061 & 0.051 & \textbf{0.050} & 0.052 & 0.062 & 0.064 & 0.058 & 0.060 & 0.055 & 0.062 \\
\cline{1-18} \cline{2-18}
\multirow[t]{7}{*}{jena\_weather} & \multirow[t]{3}{*}{10T} & long & 0.053 & 0.064 & 0.063 & 0.079 & 0.060 & 0.056 & 0.049 & \textbf{0.035} & 0.050 & 0.143 & 0.066 & 0.052 & 0.304 & 0.424 & 0.304 \\
 &  & medium & 0.054 & 0.057 & 0.060 & 0.068 & 0.059 & 0.054 & 0.048 & \textbf{0.031} & 0.049 & 0.073 & 0.065 & 0.052 & 0.277 & 0.350 & 0.277 \\
 &  & short & 0.034 & 0.033 & 0.037 & 0.042 & 0.036 & 0.031 & 0.027 & \textbf{0.016} & 0.027 & 0.063 & 0.064 & 0.069 & 0.155 & 0.130 & 0.155 \\
\cline{2-18}
 & D & short & 0.047 & 0.045 & 0.047 & 0.047 & \textbf{0.043} & 0.048 & 0.044 & 0.058 & 0.051 & 0.062 & 0.053 & 0.069 & 0.080 & 0.082 & 0.297 \\
\cline{2-18}
 & \multirow[t]{3}{*}{H} & long & 0.103 & 0.062 & 0.068 & 0.066 & 0.058 & 0.066 & 0.059 & 0.068 & \textbf{0.057} & 0.197 & 0.076 & 0.090 & 0.230 & 1.290 & 0.598 \\
 &  & medium & 0.058 & 0.054 & 0.058 & 0.058 & 0.055 & 0.058 & 0.053 & 0.066 & \textbf{0.053} & 0.078 & 0.069 & 0.073 & 0.211 & 0.832 & 0.486 \\
 &  & short & 0.042 & 0.042 & 0.043 & 0.042 & 0.042 & 0.050 & \textbf{0.041} & 0.045 & 0.042 & 0.699 & 0.050 & 0.048 & 0.143 & 0.296 & 0.173 \\
\cline{1-18} \cline{2-18}
\multirow[t]{4}{*}{kdd\_cup\_2018} & D & short & \textbf{0.362} & 0.372 & 0.373 & 0.365 & 0.389 & 0.396 & 0.376 & 0.378 & 0.387 & 0.383 & 0.401 & 0.380 & 0.393 & 0.459 & 0.888 \\
\cline{2-18}
 & \multirow[t]{3}{*}{H} & long & 0.478 & \textbf{0.300} & 0.419 & 0.472 & 0.518 & 0.375 & 0.441 & 0.518 & 0.457 & 1.090 & 0.477 & 0.503 & 1.050 & 0.970 & 1.250 \\
 &  & medium & 0.450 & \textbf{0.301} & 0.364 & 0.416 & 0.495 & 0.377 & 0.421 & 0.466 & 0.441 & 0.442 & 0.442 & 0.472 & 0.851 & 0.791 & 0.949 \\
 &  & short & 0.418 & \textbf{0.246} & 0.267 & 0.313 & 0.426 & 0.351 & 0.378 & 0.376 & 0.403 & 0.517 & 0.457 & 0.467 & 0.559 & 0.531 & 0.559 \\
\cline{1-18} \cline{2-18}
\multirow[t]{7}{*}{loop\_seattle} & \multirow[t]{3}{*}{5T} & long & 0.090 & 0.129 & 0.125 & 0.121 & 0.087 & 0.084 & 0.084 & 0.114 & \textbf{0.077} & 0.184 & 0.095 & 0.088 & 0.137 & 0.231 & 0.137 \\
 &  & medium & 0.087 & 0.116 & 0.119 & 0.116 & 0.080 & 0.077 & 0.078 & 0.110 & \textbf{0.072} & 0.118 & 0.095 & 0.092 & 0.123 & 0.240 & 0.123 \\
 &  & short & 0.053 & 0.055 & 0.055 & 0.055 & \textbf{0.046} & 0.050 & 0.048 & 0.051 & 0.048 & 0.072 & 0.066 & 0.065 & 0.081 & 0.082 & 0.081 \\
\cline{2-18}
 & D & short & 0.043 & 0.044 & 0.045 & 0.046 & 0.043 & 0.047 & 0.042 & \textbf{0.041} & 0.044 & 0.052 & 0.046 & 0.048 & 0.078 & 0.072 & 0.131 \\
\cline{2-18}
 & \multirow[t]{3}{*}{H} & long & 0.063 & 0.076 & 0.082 & 0.087 & 0.066 & 0.072 & \textbf{0.060} & 0.066 & 0.065 & 0.068 & 0.069 & 0.068 & 0.193 & 0.468 & 0.245 \\
 &  & medium & 0.067 & 0.076 & 0.082 & 0.087 & 0.069 & 0.075 & \textbf{0.064} & 0.067 & 0.064 & 0.072 & 0.071 & 0.069 & 0.154 & 0.390 & 0.206 \\
 &  & short & 0.063 & 0.065 & 0.066 & 0.071 & 0.062 & 0.067 & \textbf{0.058} & 0.059 & 0.063 & 0.066 & 0.076 & 0.073 & 0.108 & 0.165 & 0.108 \\
\cline{1-18} \cline{2-18}
m4\_daily & D & short & 0.023 & 0.021 & 0.021 & 0.021 & \textbf{0.020} & 0.027 & 0.021 & 0.021 & 0.022 & 0.030 & 0.023 & 0.023 & 0.023 & 0.024 & 0.026 \\
\cline{1-18} \cline{2-18}
m4\_hourly & H & short & 0.030 & 0.025 & 0.020 & 0.021 & 0.024 & 0.023 & 0.020 & \textbf{0.011} & 0.035 & 0.133 & 0.039 & 0.040 & 0.034 & 0.041 & 0.040 \\
\cline{1-18} \cline{2-18}
m4\_monthly & M & short & 0.094 & 0.094 & 0.094 & 0.095 & 0.095 & 0.116 & 0.091 & \textbf{0.067} & 0.097 & 0.184 & 0.102 & 0.113 & 0.098 & 0.098 & 0.126 \\
\cline{1-18} \cline{2-18}
m4\_quarterly & Q & short & 0.078 & 0.077 & 0.078 & 0.079 & 0.075 & 0.093 & 0.073 & \textbf{0.062} & 0.078 & 0.083 & 0.083 & 0.083 & 0.082 & 0.079 & 0.099 \\
\cline{1-18} \cline{2-18}
m4\_weekly & W & short & 0.037 & 0.038 & 0.038 & 0.041 & 0.040 & 0.043 & \textbf{0.037} & 0.042 & 0.049 & 0.062 & 0.040 & 0.049 & 0.050 & 0.053 & 0.073 \\
\cline{1-18} \cline{2-18}
m4\_yearly & A & short & 0.118 & 0.121 & 0.128 & 0.129 & 0.116 & 0.160 & 0.116 & \textbf{0.091} & 0.122 & 0.113 & 0.117 & 0.110 & 0.130 & 0.115 & 0.138 \\
\cline{1-18} \cline{2-18}
\multirow[t]{4}{*}{m\_dense} & D & short & 0.061 & 0.069 & 0.072 & 0.082 & 0.070 & 0.067 & 0.068 & \textbf{0.060} & 0.075 & 0.076 & 0.070 & 0.077 & 0.135 & 0.126 & 0.294 \\
\cline{2-18}
 & \multirow[t]{3}{*}{H} & long & 0.165 & 0.170 & 0.146 & 0.198 & 0.120 & 0.130 & 0.115 & 0.127 & 0.128 & 0.130 & 0.120 & \textbf{0.115} & 0.270 & 1.430 & 0.552 \\
 &  & medium & 0.160 & 0.157 & 0.134 & 0.155 & 0.120 & 0.128 & 0.116 & 0.127 & 0.121 & 0.118 & 0.127 & \textbf{0.114} & 0.255 & 1.210 & 0.479 \\
 &  & short & 0.155 & \textbf{0.125} & 0.133 & 0.140 & 0.132 & 0.133 & 0.128 & 0.139 & 0.148 & 0.128 & 0.173 & 0.139 & 0.281 & 0.549 & 0.281 \\
\cline{1-18} \cline{2-18}
restaurant & D & short & 0.263 & 0.264 & 0.264 & 0.276 & 0.260 & 0.286 & \textbf{0.255} & 0.261 & 0.297 & 0.270 & 0.262 & 0.284 & 0.362 & 0.329 & 0.907 \\
\cline{1-18} \cline{2-18}
\multirow[t]{3}{*}{saugeen} & D & short & 0.373 & 0.338 & 0.354 & 0.339 & \textbf{0.328} & 0.379 & 0.358 & 0.408 & 0.353 & 0.572 & 0.408 & 0.419 & 0.564 & 0.669 & 0.754 \\
\cline{2-18}
 & M & short & \textbf{0.276} & 0.296 & 0.293 & 0.288 & 0.291 & 0.332 & 0.297 & 0.342 & 0.299 & 0.689 & 0.372 & 0.340 & 0.326 & 0.373 & 0.445 \\
\cline{2-18}
 & W & short & 0.395 & 0.363 & 0.372 & 0.364 & 0.419 & 0.406 & \textbf{0.350} & 0.601 & 0.390 & 0.397 & 0.484 & 0.491 & 0.549 & 0.734 & 0.855 \\
\cline{1-18} \cline{2-18}
\multirow[t]{8}{*}{solar} & \multirow[t]{3}{*}{10T} & long & 0.331 & 0.443 & 0.497 & 0.534 & 0.408 & 0.365 & \textbf{0.321} & 0.498 & 0.352 & 0.549 & 0.339 & 0.379 & 0.786 & 6.640 & 0.786 \\
 &  & medium & \textbf{0.326} & 0.436 & 0.453 & 0.495 & 0.412 & 0.373 & 0.334 & 0.516 & 0.353 & 0.485 & 0.356 & 0.362 & 0.771 & 5.670 & 0.771 \\
 &  & short & 0.458 & 0.511 & 0.498 & 0.488 & \textbf{0.350} & 0.444 & 0.541 & 0.804 & 0.541 & 0.933 & 1.370 & 0.618 & 0.860 & 2.360 & 0.860 \\
\cline{2-18}
 & D & short & \textbf{0.269} & 0.287 & 0.286 & 0.282 & 0.303 & 0.324 & 0.273 & 0.278 & 0.290 & 0.682 & 0.287 & 0.277 & 0.282 & 0.286 & 0.757 \\
\cline{2-18}
 & \multirow[t]{3}{*}{H} & long & 0.351 & 0.405 & 0.373 & 0.354 & 0.332 & 0.293 & \textbf{0.287} & 0.493 & 0.331 & 0.381 & 0.353 & 0.401 & 0.607 & 7.320 & 1.470 \\
 &  & medium & 0.313 & 0.368 & 0.356 & 0.342 & 0.333 & 0.309 & \textbf{0.285} & 0.376 & 0.331 & 0.352 & 0.344 & 0.330 & 0.557 & 6.130 & 1.270 \\
 &  & short & 0.336 & 0.298 & 0.303 & 0.313 & 0.342 & 0.329 & \textbf{0.273} & 0.406 & 0.328 & 0.389 & 0.340 & 0.367 & 0.628 & 2.330 & 0.628 \\
\cline{2-18}
 & W & short & 0.120 & 0.133 & 0.136 & 0.132 & 0.163 & 0.148 & 0.144 & 0.171 & 0.186 & 0.242 & 0.162 & \textbf{0.114} & 0.152 & 0.155 & 0.236 \\
\cline{1-18} \cline{2-18}
\multirow[t]{4}{*}{sz\_taxi} & \multirow[t]{3}{*}{15T} & long & 0.242 & 0.248 & 0.245 & 0.240 & 0.217 & 0.221 & \textbf{0.198} & 0.227 & 0.202 & 0.286 & 0.281 & 0.241 & 0.398 & 0.629 & 0.554 \\
 &  & medium & 0.230 & 0.244 & 0.246 & 0.240 & 0.214 & 0.228 & \textbf{0.202} & 0.229 & 0.205 & 0.210 & 0.220 & 0.206 & 0.351 & 0.529 & 0.454 \\
 &  & short & 0.209 & 0.202 & 0.203 & 0.203 & 0.201 & 0.223 & 0.200 & \textbf{0.199} & 0.203 & 0.219 & 0.207 & 0.222 & 0.309 & 0.288 & 0.309 \\
\cline{2-18}
 & H & short & 0.140 & 0.136 & 0.137 & 0.137 & 0.136 & 0.154 & 0.135 & \textbf{0.135} & 0.137 & 0.139 & 0.144 & 0.144 & 0.170 & 0.232 & 0.229 \\
\cline{1-18} \cline{2-18}
temperature\_rain & D & short & 0.569 & \textbf{0.538} & 0.544 & 0.548 & 0.561 & 0.620 & 0.547 & 0.586 & 0.560 & 0.682 & 0.644 & 0.592 & 0.694 & 0.761 & 1.630 \\
\cline{1-18} \cline{2-18}
\multirow[t]{3}{*}{us\_births} & D & short & \textbf{0.016} & 0.026 & 0.028 & 0.037 & 0.020 & 0.022 & 0.023 & 0.019 & 0.026 & 0.028 & 0.025 & 0.016 & 0.074 & 0.075 & 0.144 \\
\cline{2-18}
 & M & short & 0.015 & 0.019 & 0.016 & 0.017 & 0.017 & 0.028 & 0.012 & 0.011 & 0.013 & 0.016 & 0.017 & 0.021 & \textbf{0.010} & 0.019 & 0.017 \\
\cline{2-18}
 & W & short & \textbf{0.011} & 0.013 & 0.013 & 0.014 & 0.011 & 0.017 & 0.012 & 0.013 & 0.014 & 0.017 & 0.015 & 0.019 & 0.018 & 0.018 & 0.022 \\
\cline{1-18} \cline{2-18}
\bottomrule
\end{tabular}
\textbf{}}
\end{table}

\newpage

\subsection{Covariate-Informed Forecasting Evaluation on \texttt{fev-bench}} \label{sec:append-fev-bench}

We evaluate on the subset of \texttt{fev-bench} tasks that include covariates in both the conditioning and prediction windows (commonly referred to in the time-series literature as past and known future covariates).
The tasks used are listed in Table~\ref{tab:fev-bench-tasks}.  
Detailed metadata are provided in the original benchmark paper \citep{shchur2025fevbenchrealisticbenchmarktime}.  
As these details are already standardized and publicly available, we refer readers to the benchmark documentation rather than replicate the full tables here.

\begin{table}[h!]
    \centering
    \small
    \begin{tabular}{cl}
\toprule
\textbf{\#} & \textbf{\texttt{fev-bench} Tasks} \\
\midrule
 1 & proenfo\_gfc12 \\
 2 & proenfo\_gfc14 \\
 3 & proenfo\_gfc17 \\
 4 & rohlik\_orders\_1D \\
 5 & rohlik\_sales\_1W \\
 6 & rohlik\_orders\_1W \\
 7 & entsoe\_15T \\
 8 & entsoe\_30T \\
 9 & entsoe\_1H \\
10 & epf\_be \\
11 & epf\_de \\
12 & epf\_fr \\
13 & epf\_np \\
14 & epf\_pjm \\
15 & rossmann\_1D \\
16 & rossmann\_1W \\
17 & hermes \\
18 & walmart \\
19 & m5\_1W \\
20 & m5\_1M \\
21 & favorita\_stores\_1D \\
22 & favorita\_stores\_1W \\
23 & favorita\_stores\_1M \\
24 & favorita\_transactions\_1D \\
25 & solar\_with\_weather\_15T \\
26 & solar\_with\_weather\_1H \\
27 & uci\_air\_quality\_1H \\
28 & uci\_air\_quality\_1D \\
\bottomrule
\end{tabular}    
    \caption{Subset of \texttt{fev-bench} tasks used for covariate-informed forecasting.  
    Detailed task characteristics are available in the original benchmark paper~\citep{shchur2025fevbenchrealisticbenchmarktime}.}
    \label{tab:fev-bench-tasks}
\end{table}

\pagebreak

\subsection{Additional Results for Harmonic Reconstruction Experiments} \label{sec:append:harmonic-reconstruction}

\begin{figure}[H]
    \centering
    \includegraphics[width=1.0\linewidth]{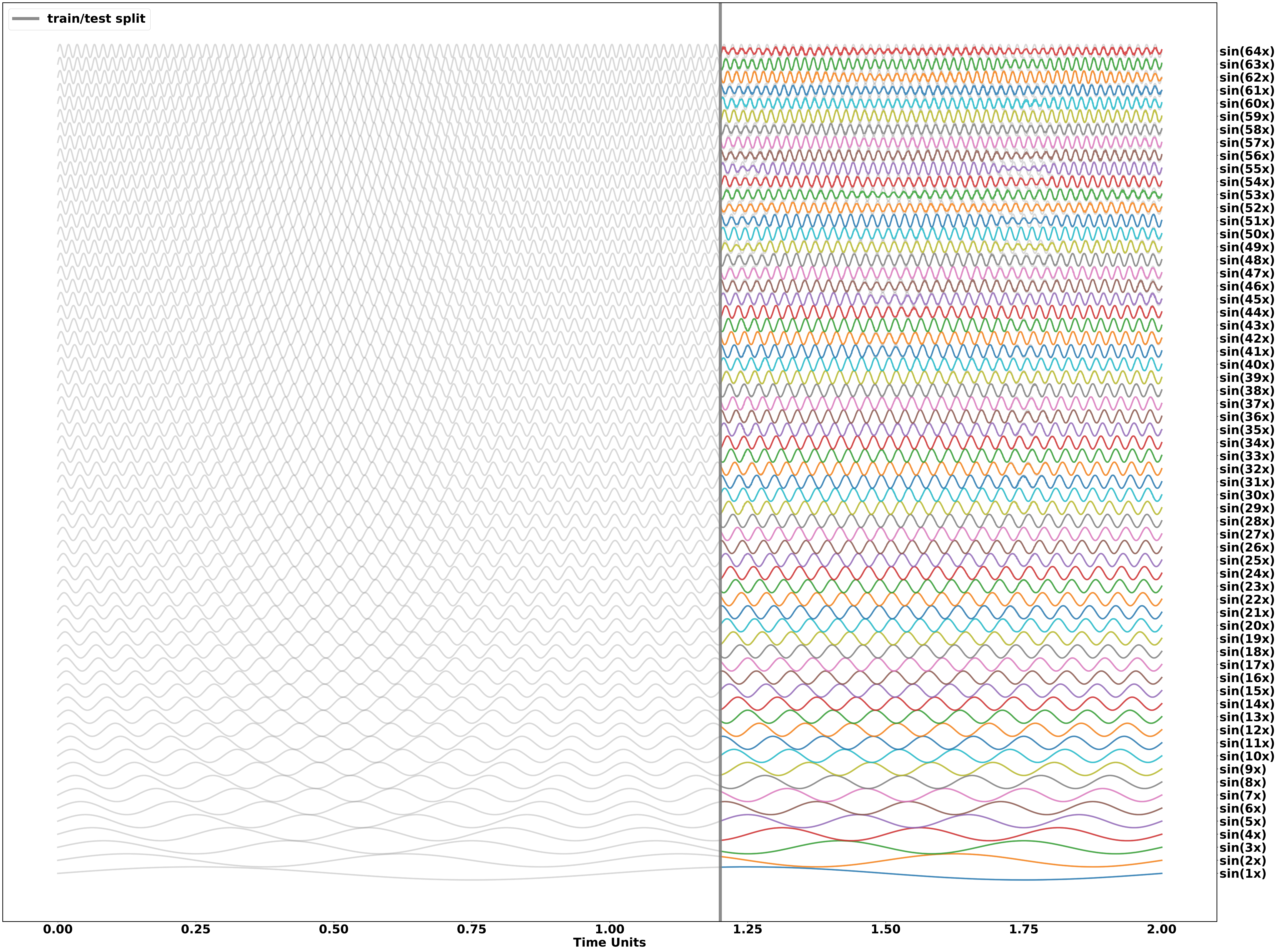}
    \caption{Reconstruction of $\sin(nt)$ under the injective temporal embedding $\{\sin(t),\cos(t)\}$.
    \tabpfn is evaluated on $\sin(nt)$ for $n = 1,\dots,64$.
    The model accurately reconstructs low and mid-range frequencies, but accuracy gradually degrades at higher n, consistent with the finite phase resolution of the sinusoidal embedding given the sampling density.}
    \label{fig:sin-cos-generalization-full}
\end{figure}

\begin{figure}[H]
    \centering
    \includegraphics[width=1.0\linewidth]{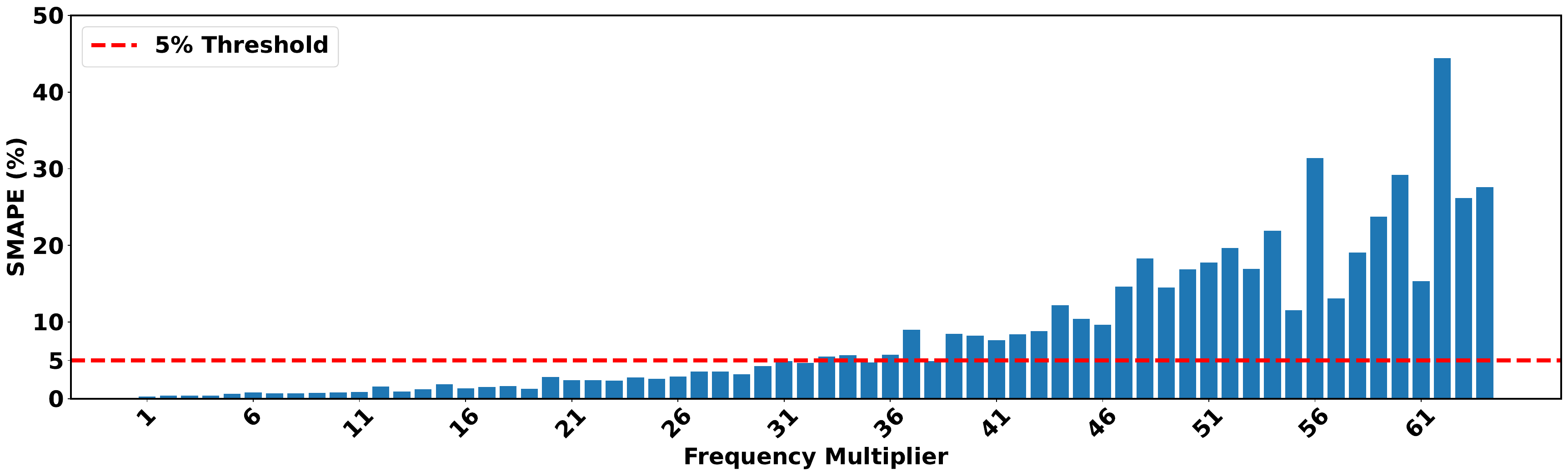}
    \caption{Symmetric mean absolute percentage error (sMAPE) for predicting $\sin(nt)$ using $\{\sin(t),\cos(t)\}$ as input features.
    Errors remain low for moderate frequencies but increase as n grows, reflecting the reduced separability of high-frequency phases in the sinusoidal coordinate system.}
    \label{fig:sin-cos-generalization-smape}
\end{figure}

\begin{figure}[H]
  \centering
  \includegraphics[width=1.0\textwidth]
  {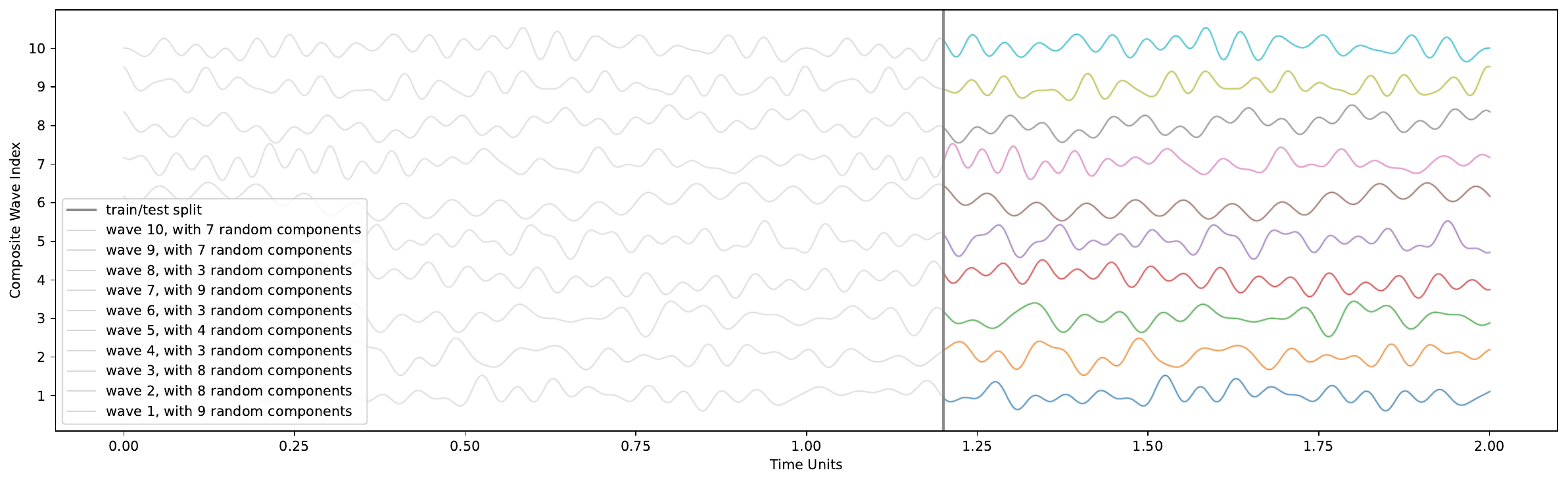}
  \caption{
      Reconstruction of composite sinusoidal signals under the injective temporal embedding $\{\sin(t),\cos(t)\}$.
    Each composite signal is the sum of 3–10 sinusoids with random frequencies, amplitudes, and phases.
    \tabpfn accurately recovers these multi-frequency patterns, consistent with interpolation in a phase-preserving temporal embedding.
    }
  \label{fig:comp-waves-generalization}
\end{figure}

\subsubsection{Phase Ambiguity and the Requirement of Injective Mapping} \label{sec:app-phase-ambiguity}

\paragraph{The Non-Injectivity of $\sin(t)$.}
When time $t$ is mapped via the feature $\phi(t) = \sin(t)$, the representation is non-injective.
Within a single period $T=2\pi$, every feature value $y \in (-1, 1)$ corresponds to two distinct phases--specifically $t$ and $\pi - t$--creating a \textit{phase ambiguity}.
To a retrieval-based model, these points are indistinguishable, resulting in a coordinate collision where the rising and falling edges of a cycle share the same address in feature space.

\paragraph{Symmetry and Harmonic Recovery.}
The capacity to recover higher-order harmonics from this ambiguous base depends on the symmetry of the target signal $y_t = \sin(nt)$ relative to these collision points:

\begin{itemize}
    \item Odd Harmonics (Consistent Mapping):
    For odd $n$, the signal is symmetric across the ambiguity: $y(t) = y(\pi - t)$.
    Consequently, both time points that share a single $\sin(t)$ feature value also share an identical target value.
    This ensures that the mapping from the feature space to the target remains consistent, allowing the model to recover the signal despite the underlying phase ambiguity.
    \item Even Harmonics (Inconsistent Mapping):
    For even $n$, however, the signal is anti-symmetric: $y(t) = -y(\pi - t)$.
    This presents the model with a mathematical contradiction--two distinct target values mapped to a single feature coordinate--hence the failure to recover even-ordered harmonics.
\end{itemize}

\paragraph{Quadrature Resolution via $\left(\sin(t), \cos(t)\right)$.}
Providing the quadrature pair $\Phi(t) = \left(\sin(t), \cos(t)\right)$ embeds time onto the unit circle $S^1 \subset \mathbb{R}^2$.
This mapping is injective: it assigns a unique coordinate to every phase within the period.
Consequently, the model can uniquely associate any target value—regardless of harmonic order—with its corresponding phase.

\newpage

\subsubsection{Discussion of Functional Synthesis vs. Phase-Space Retrieval} \label{sec:app-functional-vs-retrieval}

In this section, we investigate two competing hypotheses for how the model reconstructs higher-order harmonics ($y_t = \sin(n t)$) from base periodic features ($\sin(t), \cos(t)$).

\begin{enumerate}
    \item \textit{The Functional Approximation Hypothesis:} The model uses its internal MLP layers to compute trigonometric identities or polynomial expansions (similar to a Chebyshev expansion) to "synthesize" higher frequencies from the base inputs. An example would be, approximating
    \[
        \sin(4x) = 4\sin(x)\cos^3(x) - 4\sin^3(x)\cos(x).
    \]
    \item \textit{The Phase-Space Retrieval Hypothesis:} The model treats ($\sin(t), \cos(t)$) as a unique coordinate (an injective "key") and uses the attention mechanism to retrieve and interpolate values from the observed context window.
\end{enumerate}

\paragraph{The Half-Cycle Experiment.}

\begin{wrapfigure}{r}{0.50\textwidth}
\vspace{-2.5em}
\begin{minipage}{\linewidth}
\begin{figure}[H]
    \centering
    \includegraphics[width=1.0\linewidth]{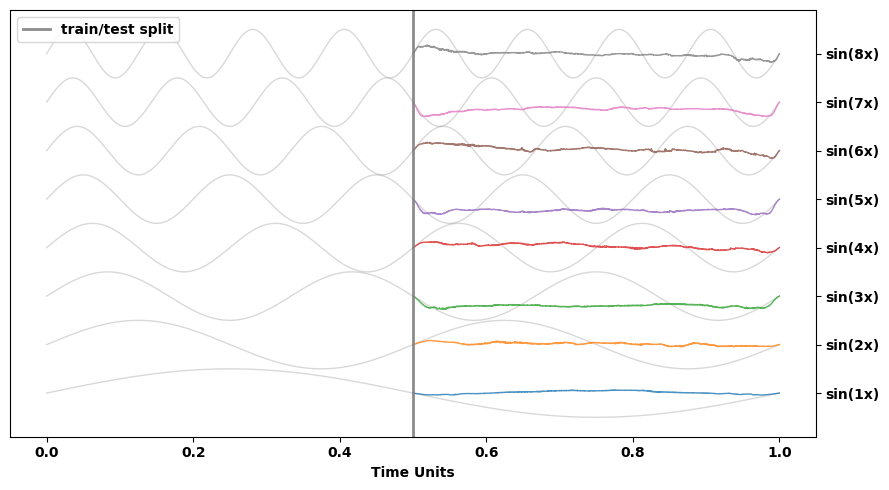}
    \caption{The Half-Cycle Experiment.
    We evaluate the model's ability to reconstruct harmonics $\sin(nt)$ when the support set features $(\sin t, \cos t)$ are restricted to a partial cycle ($t \in [0, \pi]$).
    The failure of recovering the harmonics in the unseen phase range $(\pi, 2\pi]$ suggests that the model does not generalize via global trigonometric identities.
    }
    \label{fig:half-cycle}
\end{figure}
\vspace{-2.5em}
\end{minipage}
\end{wrapfigure}

To probe these competing explanations, we test the model's ability to extrapolate beyond observed feature ranges.
If the model were performing global functional synthesis, it should theoretically be capable of generalizing the relationship $y_t = f(\sin(t), \cos(t))$ to any input value, as the underlying trigonometric identities are globally valid regardless of the observed range.

Similar to Section~\ref{sec:ablation-model-analysis}, we task the model with reconstructing higher-order harmonics ($y_t = \sin(n t)$) from base periodic features ($\sin(t), \cos(t)$).
However, we restrict the support set to $t \in \left[ 0, \pi \right]$.
In this setting, the model never observes a full cycle of the $\left(\sin(t), \cos(t)\right)$ coordinate system in context, which is a prerequisite for phase-space retrieval.

As shown in Figure~\ref{fig:half-cycle}, the model fails to reconstruct the signals in the unobserved range.
If the model were performing polynomials expansions, it could theoretically infer the relationship from the mathematically sufficient, albeit partial, view of the mapping $t \mapsto \left(\sin(t), \cos(t)\right)$.

While these results do not strictly rule out more complex forms of functional approximation, they provide empirical evidence that \tabpfn is less likely to be performing global functional synthesis in this context.
Instead, the failure to extrapolate beyond the observed coordinates point towards the model's proficiency being tied to its behavior as a context-dependent retrieval engine.

\paragraph{Conclusions.}
The evidence suggests that \tabpfnts handles seasonal signals by operating as a \textbf{meta-learned retrieval engine}.
By providing ($\sin(t), \cos(t)$), we supply a coordinate system that minimizes "collisions" (non-injectivity),
This enables \tabpfn to uniquely and accurately retrieve relevant seasonal patterns from the series history, provided those coordinates have been previously observed in the support set.

\newpage

\subsection{Synthetic Data Generation for Time Series with Covariates} \label{appendix:synthetic-cov}

\begin{figure}[H]
    \centering
    \includegraphics[width=0.95\linewidth]{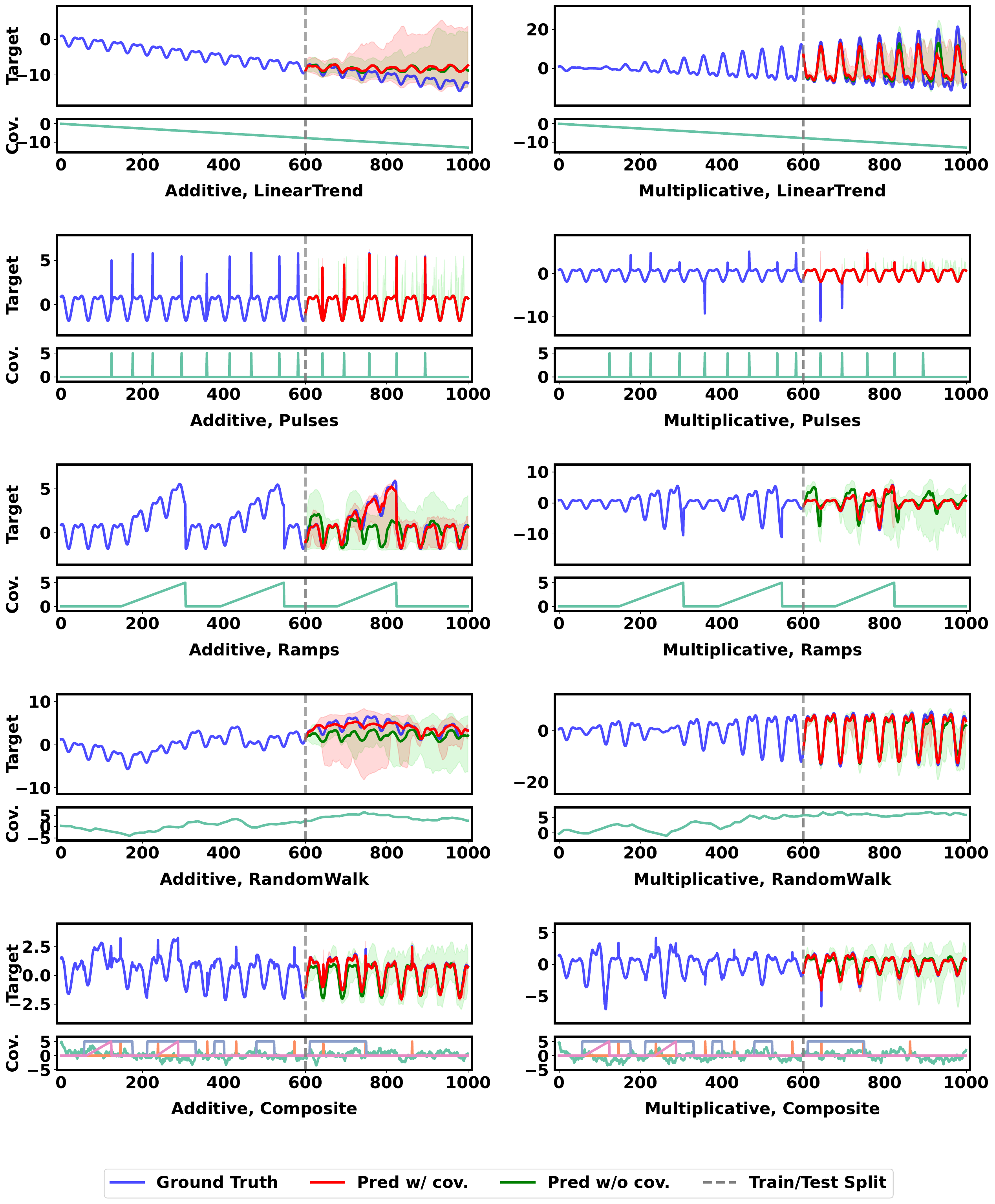}
    \caption{Forecasting visualization on synthetic covariate-augmented time series. Each subplot corresponds to one synthetic class from $\mathcal{C} = {\text{additive}, \text{multiplicative}} \times {\text{LinearTrend}, \text{Pulses}, \text{Ramps}, \text{RandomWalk}, \text{Composite}}$. Blue lines denote ground truth, red lines predictions with covariates, and green lines predictions without covariates. Shaded regions indicate predictive uncertainty, and the vertical dashed line marks the train/test split.
    }
    \label{fig:covariate_type_visualization}
\end{figure}

\pagebreak

\subsection{Ablation: Context Length vs. Accuracy}
\label{append-context-length}

In this ablation, we investigate how the amount of available context affects the performance of \tabpfnts.
We experiment with four context lengths: $1024$, $2048$, $4096$, and $10,000$.
The maximum length of $10,000$ is chosen to match the largest dataset size used during the pretraining of \tabpfn.


\vspace{1em}

\begin{figure}[H]
    \centering
    \includegraphics[width=1.0\linewidth]{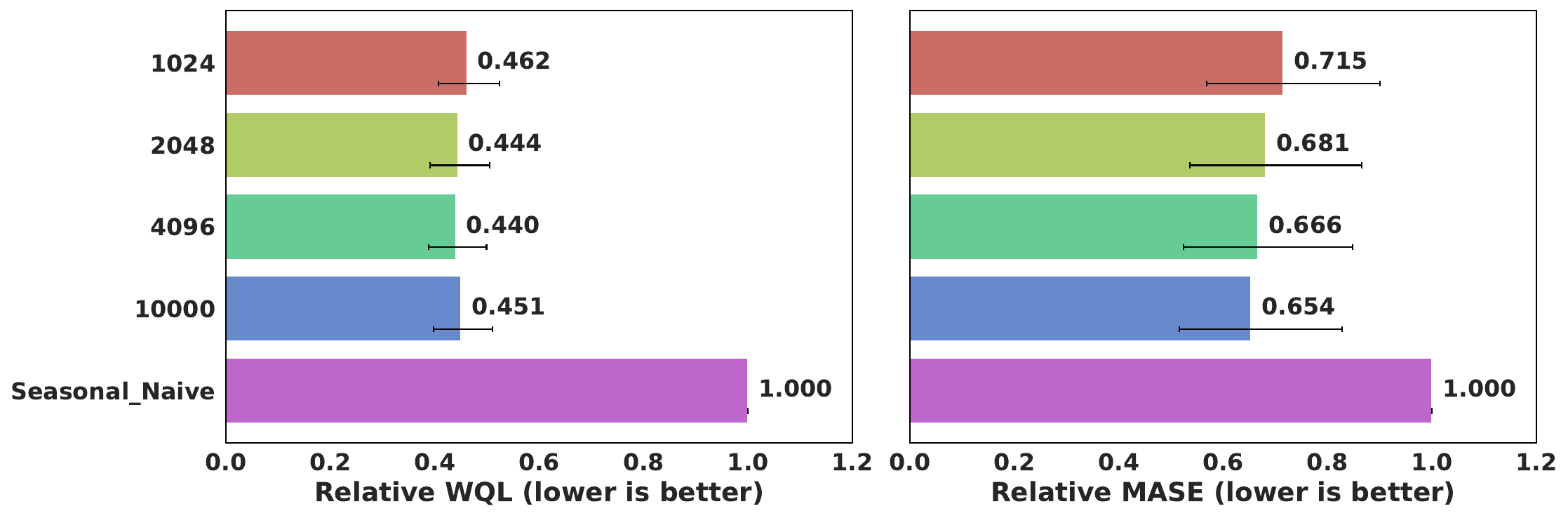}
    \caption{Effect of context length on forecasting performance of \tabpfnts.}
    \label{fig-context-length-ablation}
\end{figure}

As shown in Figure~\ref{fig-context-length-ablation}, increasing the context length leads to improved performance overall, though the gains diminish beyond 4096 points.
While MASE continues to improve with longer context, WQL shows a slight increase at the longest length.
These results suggest that moderate-length contexts are often sufficient, but the impact of longer contexts may vary depending on the forecasting objective.

\subsection{Probablistic Calibration Analysis}
\label{app:calibration}

\begin{wrapfigure}{r}{0.38\textwidth}
\vspace{-3.0em}
\begin{minipage}{\linewidth}
\begin{figure}[H]
    \centering
    \includegraphics[width=1.0\linewidth]{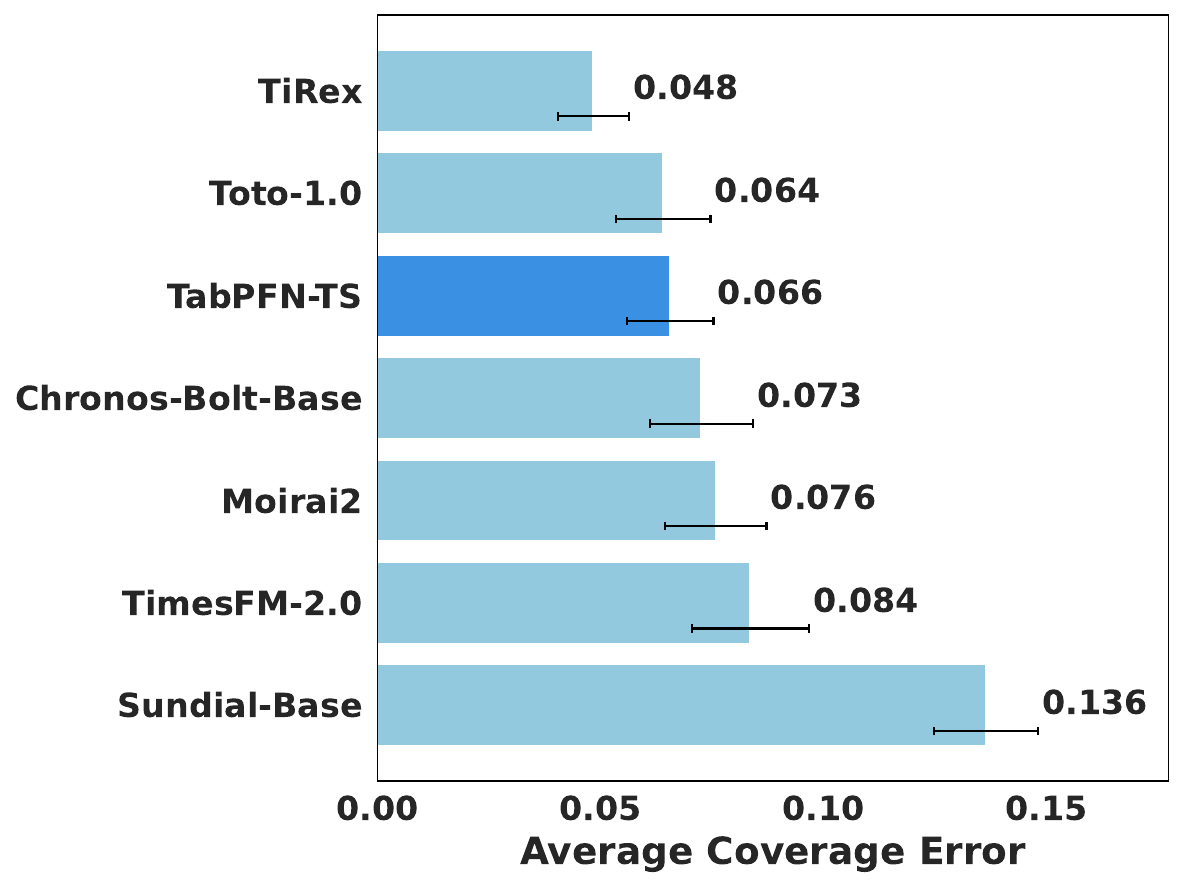}
    \vspace{-1.5em}
    \caption{Average coverage error (ACE, lower is better) across 81 tasks in GIFT-Eval. Error bars show 95\% confidence intervals.
    }
    \label{fig:calibration}
\end{figure}
\vspace{-5em}
\end{minipage}
\end{wrapfigure}

To complement Weighted Quantile Loss (WQL, see Section~\ref{sec:experiment}), we assess calibration directly, since WQL conflates
calibration with sharpness.
We report average coverage error\footnote{\url{https://ts.gluon.ai/dev/api/gluonts/gluonts.ev.metrics.html\#gluonts.ev.metrics.MAECoverage}}
\citep{gluonts_jmlr}, following the calibration framework of
\citet{gneiting2007probabilistic}:
\[
\text{ACE} = \frac{1}{|Q|} \sum_{q \in Q} \left| \text{Cov}(q) - q \right|,
\qquad
\text{Cov}(q) = \frac{1}{N}\sum_{i=1}^{N} \mathbbm{1}\!\left[y_i \le \hat{Q}_i(q)\right],
\]
where $Q = \{0.1, \ldots, 0.9\}$, $\hat{Q}_i(q)$ is the predicted $q$-quantile for test
point $i$, and $N$ ranges over all series and timesteps in a dataset. ACE is averaged
across 81 smallest tasks (out of 97) in GIFT-Eval.

Figure~\ref{fig:calibration} reports ACE for \tabpfnts and the time-series foundation
model baselines for which quantile forecasts were available.
TiRex attains the lowest
ACE (0.048), with confidence intervals separating it from the remaining models.
\tabpfnts (0.066) falls within a cluster of comparable values alongside Toto-1.0 (0.064),
Chronos-Bolt-Base (0.073), and Moirai2 (0.076), whose confidence intervals overlap
substantially with one another, while Sundial-Base (0.136) is a clear outlier on the
high end.
These results indicate that \tabpfnts's predictive uncertainty quality is comparable to most existing time-series foundation models, though not as good as TiRex.

\newpage 

\subsection{Inference Optimization Strategies} \label{sec:inference-optimization}
The inference cost of \tabpfnts is higher than competing TSFMs, as discussed in Section~\ref{sec:inference-time}.
Here we outline potential strategies that could reduce this overhead in future work.

\textbf{Patch-based sequence compression.}
The most impactful optimization would be to introduce patch-based tokenization of the input sequence, analogous to Chronos-Bolt and PatchTST. Rather than passing each time step as an individual row to \tabpfn, consecutive time steps could be aggregated into patches, reducing the effective sequence length by a factor of the patch size $p$. For a context of length $L$ and patch size $p$, this reduces the attention cost from $\mathcal{O}(L^2)$ to $\mathcal{O}((L/p)^2)$ — a reduction of $p^2$. For $p=16$ and $L=4096$, this yields a $256\times$ reduction in attention cost. Beyond reducing inference cost, patching would also enable series batching (see below) by making memory consumption per series tractable.

However, introducing patching would require either retraining a patch encoder on time series data or finding a patch aggregation strategy compatible with the existing \tabpfn architecture (e.g.\ mean pooling of temporal features within each patch).

\textbf{Series batching.}
Currently, \tabpfnts executes a separate forward pass for each individual time series. A natural optimization is to batch multiple series into a single forward pass, amortizing the fixed overhead across many series and improving throughput without any architectural changes. However, naively batching full-length series would significantly increase memory consumption due to the quadratic memory cost of attention. This makes batching most practical in conjunction with patch-based compression, which would bring per-series memory to a tractable level.

\textbf{Sparse and linear attention.}
The quadratic attention cost of \tabpfn could be addressed through sparse or linear attention mechanisms. However, since \tabpfn's attention operates over tabular rows rather than sequence tokens, standard sparse attention patterns (e.g.\ local windows, strided attention) may not directly apply. Developing attention approximations compatible with the tabular in-context learning objective of \tabpfn remains an open research direction.

\textbf{Immediate workaround: data-parallel inference.}
We note that all of the above optimizations require architectural changes and/or retraining, and therefore do not provide an immediate speed-up. In practice, the most straightforward way to reduce wall-clock inference time is data-parallel evaluation across multiple GPUs, where each GPU processes an independent subset of series. This is the approach used in our experiments, and scales linearly with the number of available GPUs. While this does not reduce the per-series computational cost, it provides a practical workaround for large-scale evaluation.

\end{document}